\documentclass[12pt]{article}

\usepackage{enumitem}
\usepackage{tabularx}
\usepackage{wrapfig}
\usepackage{titlesec}
\usepackage{mathpazo}
\usepackage{ebgaramond}
\usepackage{charter}
\usepackage{bbold}
\usepackage{booktabs}
\usepackage{array}
\usepackage{multirow}
\usepackage{pifont}
\usepackage{xcolor}
\definecolor{myblue}{RGB}{13, 71, 161}   
\definecolor{myred}{RGB}{232, 93, 117}   
\usepackage{tikz}
\usetikzlibrary{positioning, arrows.meta, shadows, shadows.blur}
\usetikzlibrary{calc} 
\usepackage{pdflscape}
\usepackage{nameref}
\usepackage{adjustbox}
\usepackage{subcaption}
\usepackage{standalone} 
\usepackage{colortbl} 
\usepackage{longtable} 

\definecolor{customgray}{gray}{10.5}  
\newcommand{\mycomment}[1]{}
\titleformat{\section}
{\normalfont\Large\bfseries\color{customgray}} 
{\thesection}{1em}{}

\titleformat{\subsection}
{\normalfont\large\bfseries\color{gray}} 
{\thesubsection}{1em}{}
\titleformat{\subsubsection}
{\color{gray}\normalsize\bfseries} 
{\thesubsubsection}{1em}{}
\titlespacing*{\section}{0pt}{1.5ex plus 1ex minus .2ex}{1.5ex plus .2ex}
\titlespacing*{\subsection}{0pt}{1.25ex plus 1ex minus .2ex}{1.25ex plus .2ex}

\usepackage[isbn=false,url=false]{biblatex}
\usepackage{times}
\usepackage{tabularray}
\usepackage{heuristica}
\usepackage{wrapfig}
\usepackage{multirow}
\usepackage{booktabs}
\usepackage{algorithm}
\usepackage{algpseudocode}
\usepackage{makecell}
\usepackage{lineno}

\usepackage[hmargin=1cm]{geometry}

\usepackage{tikz}

\usetikzlibrary{backgrounds}
\usetikzlibrary{patterns}
\usetikzlibrary{bayesnet}
\usetikzlibrary{arrows.meta, positioning, shapes.geometric,  calc, shapes, backgrounds, decorations.pathreplacing,decorations.pathmorphing, fit}
\usepackage{lmodern}
\usepackage[T1]{fontenc}

\usepackage{color}

\usepackage{array}

\usepackage{amsmath,amsfonts}
\usepackage{amssymb} 
\usepackage{mathtools}

\usepackage{listings}
\usepackage[most]{tcolorbox}

\definecolor{usercolor}{RGB}{255, 108, 108} 
\definecolor{assistantcolor}{RGB}{255, 189, 69} 

\newtcolorbox{userbox}{
    colback=usercolor!5!white,
    colframe=usercolor!75!black,
    width=\linewidth-1cm,
    left=0.5cm,
    right=0.5cm,
    boxrule=0.4mm,
    arc=2mm,
    fonttitle=\bfseries,
    title=Prompt,
    breakable
}

\newtcolorbox{assistantbox}{
    colback=assistantcolor!5!white,
    colframe=assistantcolor!75!black,
    width=\linewidth-1cm,
    right=0.5cm,
    left=0.5cm,
    boxrule=0.4mm,
    arc=2mm,
    fonttitle=\bfseries,
    title=Assistant,
    breakable
}

\lstdefinestyle{pythonstyle}{
    language=Python,
    basicstyle=\ttfamily\small\color{black},           
    keywordstyle=\color{teal}\bfseries,                 
    stringstyle=\color{brown},                          
    commentstyle=\color{gray},                          
    numberstyle=\tiny\color{gray},                      
    frame=single,                                       
    backgroundcolor=\color{gray!5},                     
    breaklines=true,                                    
    showstringspaces=false,                             
    numbers=left,                                       
    xleftmargin=1em,                                    
    rulecolor=\color{black},  
    morecomment=[s][\color{olive!80!black!80}\slshape]{"""}{"""},  
}

\lstdefinestyle{bashstyle}{
    language=bash,
    basicstyle=\ttfamily\small,
    frame=single,
    backgroundcolor=\color{gray!10},
    breaklines=true,
    showstringspaces=false
}

\usepackage{amsmath}
\usepackage{amssymb}
\usepackage{mathtools}
\usepackage{amsthm}
\theoremstyle{plain}

\theoremstyle{definition}

\theoremstyle{remark}

\definecolor{bronze}{RGB}{191, 124, 15}
\definecolor{silver}{RGB}{192, 192, 192}
\definecolor{gold}{RGB}{255, 215, 0}
\usepackage{graphicx}   
\usepackage{amsmath}    
\usepackage{amssymb}    
\usepackage{fancyhdr}   
\usepackage{geometry}   
\usepackage{titlesec}   
\usepackage{xcolor}     
\usepackage{caption}  
\titleformat{\section}{\normalfont\bfseries\Large}{\thesection}{1em}{}
\titlespacing*{\section}{0pt}{*3}{*2}

\newtoggle{final}
\togglefalse{final} 

\newcolumntype{C}{>{\centering\let\newline\\\arraybackslash\hspace{0pt}}m{2cm}}

\newcommand{\OurAgent}{\textrm{Agent K}}
\newcommand{\NumBenchmarkTasks}{81}
\newcommand{\nCompetitionsComparisonViolinReAct}{69}

\newcommand{\topPercentile}{18}
\newcommand{\quantileMastersOnKaggle}{2}
\newcommand{\numUsers}{200,000}
\newcommand{\outperformspercent}{82}

\newcommand{\nTotalMedals}{29}

\newcommand{\nGoldMedals}{9}
\newcommand{\nGoldMedalsOnMedAwardingComps}{4}
\newcommand{\nGoldMedalsNonMedAwardingComps}{5}

\newcommand{\nSilverMedals}{8}
\newcommand{\nSilverMedalsOnMedAwardingComps}{4}
\newcommand{\nSilverMedalsNonMedAwardingComps}{4}

\newcommand{\nBronzeMedals}{12}

\newcommand{\nBronzeMedalsNonMedAwardingComps}{12}
\newcommand{\AgentKEloscore}{1694}
\newcommand{\AgentKEloquantile}{83} 
 
\newcommand{\TabularTasks}{55}
\newcommand{\ComputerVision}{24}
\newcommand{\NaturalLanguage}{10}
\newcommand{\MultiModal}{11}
\newcommand{\AvgPartTab}{4000}
\newcommand{\AvgPartNlp}{1200}
\newcommand{\AvgPartMulti}{1000}
\newcommand{\TotalParti}{5000}
\newcommand{\PrizeMoney}{65000}
\newcommand{\nGoldMedalsOnMedAwardingCompstext}{four}
\newcommand{\nSilverMedalsOnMedAwardingCompstext}{four}
\newcommand{\nGoldMedalsNonMedAwardingCompstext}{five}
\newcommand{\nSilverMedalsNonMedAwardingCompstext}{four}
\newcommand{\nBronzeMedalsNonMedAwardingCompstext}{twelve}
\newcommand{\ImprovementQuantileAgentKvReAct}{5}
\newcommand{\ImprovementQuantileAgentKvReActAtTwoUnits}{19}

\newcommand{\nParticipantsEloComparison}{7,311}

\newcommand{\nGrandmasterForElo}{161}

\newcommand{\AgentKTabquantile}{70}
\newcommand{\AgentKTabGold}{3}
\newcommand{\AgentKTabSilver}{4}
\newcommand{\AgentKTabBronze}{3}
\newcommand{\TabPFNquantile}{30}
\newcommand{\TabPFNGold}{2}
\newcommand{\TabPFNSilver}{1}
\newcommand{\TabPFNBronze}{2}

\newcommand{\nMedalsTimeConstrained}{19}
\newcommand{\nMedalsTimeConstrainedCV}{7}
\newcommand{\nMedalsTimeConstrainedNLP}{6}
\newcommand{\nMedalsTimeConstrainedTAB}{6}

\usepackage{tcolorbox}

\usepackage{url}      
\usepackage{hyperref}   

\definecolor{headerblue}{RGB}{250, 245, 245}
\definecolor{titleblack}{RGB}{80, 0, 0} 
\definecolor{abstractblack}{RGB}{80, 0, 0} 

\newtcolorbox{fullbox}{
colback=headerblue,
colframe=white,
width=\textwidth,
boxrule=0pt,
arc=10pt,
outer arc=10pt,
boxsep=10pt,
left=10pt,
right=10pt,
top=10pt,
bottom=10pt
}

\usepackage{ifthen}

\newboolean{submissionMode}
\setboolean{submissionMode}{false}

\author
{Firstname Lastname,$^{1}$ 
\\
\normalsize{$^{1}$Huawei Noah’s Ark Lab, London, UK.}\\
\normalsize{$^{2}$Technical University of Darmstadt, Darmstadt, Germany.}\\
\normalsize{$^{3}$University College London, London, UK.}\\
\\
\normalsize{
Correspondence: 
firstname.lastname@huawei.com,} \\
}

\date{}

\begin{document} 

\ifthenelse{\boolean{submissionMode}}{%
\linenumbers
}{}




\begin{fullbox}
\vspace{0.1em}
\begin{center}
{\Large \textbf{Kolb-Based Experiential Learning for Generalist Agents with Human-Level Kaggle Data Science Performance}}

\vspace{1em}

\textbf{\textsf{Antoine Grosnit$^{1,3,\dagger}$, Alexandre Maraval$^{1,\dagger}$, Refinath S N$^{1}$, Zichao Zhao$^{1, 2}$, James Doran$^{1}$, Giuseppe Paolo$^{1}$, Albert Thomas$^{1}$, Jonas Gonzalez$^{1}$, Abhineet Kumar$^{1}$, Khyati Khandelwal$^{1}$, Abdelhakim Benechehab$^{1}$, Hamza Cherkaoui$^{1}$, Youssef Attia El-Hili$^{1}$, Kun Shao$^{1}$, Jianye Hao$^{1}$,  Jun Yao$^{1}$, Bal\'{a}zs K\'{e}gl$^{1,\star}$, Haitham Bou-Ammar$^{1,2,\star}$, Jun Wang$^{2,\star}$}}\\
{\small $^1$ Huawei Noah's Ark Lab} {\small $\ ^2$ AI Centre, UCL}
{\small $\ ^3$ TU Darmstadt}\\
{\small $^{\star}$ Corresponding Authors, \textdagger Equal contributions}
\vspace{0.1em}

\end{center}

\noindent
\textbf{Abstract:}

Human expertise emerges through iterative cycles of interaction, reflection, and internal model updating, which are central to cognitive theories such as Kolb’s experiential learning and Vygotsky’s zone of proximal development. In contrast, current AI systems, particularly large language models (LLMs) agents, rely on static pre-training or rigid workflows and lack mechanisms for continual adaptation. Recent studies have identified early cognitive traits in LLM agents, including reflection, revision, and self-correction, which suggest foundational elements of human-like experiential learning. This leads to a key question: Can we design LLM agents capable of structured, cognitively grounded learning similar to human processes?

To address this, we propose a computational framework of  Kolb’s learning cycle with Vygotsky’s ZPD for autonomous agents. Our architecture separates extrinsic functions (environment interaction) from intrinsic functions (internal reflection and abstraction), enabling cognitively grounded scaffolded learning, where the agent initially learns within structured, supportive environments, followed by open-ended generalisation. This approach empowers agents to master complex, many-step tasks ; domains that traditional fine-tuning or simple reflective methods could not tackle effectively.

Its potential is powerfully demonstrated through direct competition with humans in real-world Kaggle data science challenges. 
Learning fully automated, end-to-end data science code generation across \NumBenchmarkTasks{} tasks, our system, \OurAgent{}, demonstrated the ability to perform the entire workflow without human intervention, achieving an Elo-MMR score of \AgentKEloscore{}, placing it beyond median performance of the Kaggle Masters (the top \quantileMastersOnKaggle\% among over \numUsers\ users) included in our study. 
With \nGoldMedals{} gold, \nSilverMedals{} silver, and \nBronzeMedals{} bronze medals level performance -- including \nGoldMedalsOnMedAwardingComps{} gold and \nSilverMedalsOnMedAwardingComps{} silver on prize-awarding competitions -- \OurAgent{} is the first AI system to successfully integrate Kolb- and Vygotsky-inspired human cognitive learning, marking a major step toward generalist AI.

\end{fullbox}

\thispagestyle{fancy}

As early as 350 BCE, Aristotle observed that we learn not by being told, but by doing: \textit{``For the things we have to learn before we can do them, we learn by doing them'' (Nicomachean Ethics)}. From philosophy to cognitive science, this principle has remained central to our understanding of human learning.

From the earliest days of AI, the field has aspired to create systems that learn from experience rather than rely on hand-crafted rules \cite{NewellSimon1976, Turing1950}. This ambition has driven decades of progress, from developing neural networks to the rise of deep learning~\cite{LeCun2015}, where powerful models are trained on vast amounts of static data. These advances have culminated in large language models (LLMs) that exhibit remarkable generalisation and emergent reasoning abilities \cite{Brown2020,Wei2022}, achieving behaviours once thought to be uniquely human. Yet these systems learn from experience that is mostly fixed in advance: despite involving pretraining, fine-tuning, and RLHF, these models depend on static, pre-collected data, mostly scraped from the internet. As global data sources saturate \cite{Bommasani2021}, the limits of this approach become clear.

The next frontier is to build experiential agents that can learn through experience they actively generate, notably by     interacting with environments, reflecting on outcomes, and adapting their internal strategies over time. 
Reinforcement learning represents a step in this direction, enabling agents to master complex tasks through trial-and-error, with high-profile successes such as AlphaGo, AlphaZero, and MuZero \cite{Schrittwieser2020, Silver2017, Silver2016}. 
These systems achieved superhuman performance, but primarily in environments that are well-specified, simulation-friendly, and governed by clearly defined objectives \cite{silver2025welcome}. 

Aiming to extend these capabilities to more general and open-ended settings, recent advances in LLMs have enabled a new class of agents that demonstrate the capacity for basic reflection and internal reasoning across diverse tasks. Often combining LLMs with reinforcement learning or programmatic feedback loops, methods such as ReAct, Reflexion, and Voyager \cite{Reflexion, Voyager, Yao2022ReAct} allow agents to reason about past actions, revise plans, and interact more flexibly within their environments. 
While these approaches represent important progress, they typically rely on prompt-level heuristics and lack a principled architecture for structured, long-term learning or internal strategy adaptation.
Nonetheless, they reveal a critical shift: reflection, long regarded as a core mechanism in human cognition \cite{Fleming2012TheNB}, is now emerging as a viable computational capability. Thus, a natural question arises: \textit{``Can we design agents that learn the way humans do—through structured cycles of experience, reflection, abstraction, and adaptation?''}

To explore this, we draw on Kolb’s experiential learning theory \cite{Kolb1984}, a foundational model in the cognitive sciences that describes learning as an iterative cycle comprising four stages to support the development of internal models: concrete experience, reflective observation, abstract conceptualization, and active experimentation.
This framework has shaped educational theory and practice, emphasising that effective learning requires not only doing, but also structured internal reorganisation.
This was demonstrated through empirical studies~\cite{Raschick1998ImprovingFE, Burch2019AMO, Meyer2020ExperientialLC} following cohorts of students who experienced Kolb-cycle-based instructional sequences and showed significant improvements on objective learning measures. 
Complementing this, Vygotsky’s zone of proximal development (ZPD) \cite{Vygotsky1978} suggests that learners benefit most when guided through tasks just beyond their current ability, an idea that underpins modern approaches to scaffolding. Empirical studies across domains, from clinical training \cite{Yardley2012} to psychology education \cite{Knott2013}, show that embedding experiential learning within scaffolded environments enhances outcomes. 
With LLMs now exhibiting reasoning and self-reflective capabilities, these foundational theories offer a timely blueprint for computational models of agent experiential learning.

In this work, we propose a computational framework that implements Kolb’s experiential learning cycle, enabling autonomous agents to effectively learn through experience. 
To structure progression, we incorporate Vygotsky’s ZPD, guiding agents from scaffolded stages toward open-ended tasks.
 Echoing Kolb’s alternation between action and reflection, we model agent learning as a cycle between extrinsic and intrinsic functions. 
 Extrinsic functions govern outward interaction, such as executing code, selecting actions, and gathering feedback. 
 Intrinsic functions operate over the agent’s internal state, enabling it to reflect, abstract, hypothesise, and adapt its strategy. 
 These components are modular and composable, allowing nested, multi-step reasoning and ongoing self-improvement. 
 
 Unlike traditional gradient-based approaches that rely on model parameters updates, our framework enables autonomous adaptation through internal state transformations. By separating and dynamically interleaving internal cognition with external interaction, our system offers a computational analogue to human experiential learning, supporting agents that do not merely react or act, but learn and evolve from their own experience effectively.

To test the hypothesis that modelling Kolb’s experiential learning cycle enables generalist intelligence, we evaluate our framework on Kaggle \cite{kaggle}, the world’s leading platform for competitive data science. 
Kaggle challenges comprise high-stakes, real-world problems  in domains like finance, healthcare, and climate science, where success demands not only technical expertise -- such as data preprocessing, feature engineering, and model selection --  but also
iterative refinement, strategic experimentation, and adaptability, all of which align with Kolb’s learning phases.

Unlike synthetic benchmarks, Kaggle competitions are designed for human experts and evaluated via public and private leaderboards, providing a rigorous test of generalisation. 
While previous automation efforts such as AutoML have focused on specific subtasks like hyperparameter tuning \cite{CowenRivers2022HEBO}, they rely on fixed heuristics and struggle to generalise across different modalities. 
In contrast, our experiential agent autonomously manages the entire data science pipeline, from fetching Kaggle problems, building and refining solutions, to submitting its results to the platform. 
This clearly differs from earlier attempts on Kaggle that depend on offline datasets or partial automation~\cite{Jiang2024AIDE_ML_WECO,chan2024mlebenchevaluatingmachinelearning} and do not include direct comparisons against human participants on the official final leaderboard.

We argue that fully automatic Kaggle serves as a milestone environment for agents, akin to Atari \cite{mnih2013playing} in deep reinforcement learning and Go \cite{silver2017mastering} in multi-agent self-play. 
Just as those benchmarks demonstrated emergent planning and learning capabilities, Kaggle offers a rigorous testbed for measuring generality, adaptability, and human-level performance in autonomous data science.

We instantiate our framework in \OurAgent{}, a fully autonomous system that learns to construct and refine high-performance data science pipelines without human intervention. 
Across a broad range of Kaggle competitions, including tabular, computer vision, and natural language processing challenges, \OurAgent{} achieved performance at the level of experienced human data scientists. 
Its Elo-MMR places it on par with the median of Kaggle Masters, an elite group representing less than \quantileMastersOnKaggle\% of the platform’s \numUsers+ users. 
In featured and research competitions granting Kaggle medals, \OurAgent{} would have earned \nGoldMedalsOnMedAwardingComps{} gold and silver medals, and it demonstrated medal-equivalent performance (\nGoldMedalsNonMedAwardingComps{} gold, \nSilverMedalsNonMedAwardingComps{} silver and \nBronzeMedalsNonMedAwardingComps{} bronze) in many others.
To our knowledge, this is the first demonstration of a fully autonomous agent achieving consistent, human-competitive results across the full data science pipeline in real-world environments, offering empirical evidence that a computationally grounded cycle of Kolb’s experiential learning can serve as a viable foundation for generalist AI.

\section*{Computational Models of Kolb's Experiential Learning}

We now formalise our experiential learning framework by distinguishing between two core computational roles: extrinsic functions, which govern the agent’s outward interaction with the environment (e.g., selecting actions, receiving feedback), and intrinsic functions, which operate over the agent’s internal state to support reflection, abstraction, and adaptation. 
These functions are composable and can be applied iteratively, enabling structured internal reasoning processes prior to action~\cite{Christianos2023PanguAgentAF}.
A key enabler of this framework is the use of LLMs as they naturally support open-ended inputs and outputs, which is a fundamental prerequisite for experiential learning in unstructured and dynamic environments.
We present actual implementations of the extrinsic and intrinsic functions using LLM calls in Figures \ref{fig:exp-learning-in-scaffold},  and \ref{fig:exp-learning-post-scaffold} of the Method section.

\begin{wrapfigure}{r}{.5\textwidth}   
\begin{tikzpicture}[scale=.8,
  box/.style={draw=none, fill={rgb,255:red,7; green,138; blue,199}, rounded corners, minimum width=2.5cm, align=center, anchor=center, text=white, font=\bfseries},
  ->, >=Stealth,line width=2pt, rounded corners=15pt, every node/.append style={transform shape, anchor=center}
]

\def\hDist{8cm}
\def\vDist{4.2cm}

\node[] (topLeftCorner) {};
\node[right=\hDist of topLeftCorner, anchor=center] (topRightCorner) {};
\node[below=\vDist of topLeftCorner, anchor=center] (bottomLeftCorner) {};
\node[box, right=\hDist of bottomLeftCorner, below=\vDist of topRightCorner, anchor=center] (bottomRightCorner) {$\mathcal{F}_t$ \\[-0.5ex] \scriptsize Environment \\[-0.5ex] \scriptsize Feedback };

\node[box] at ($(topLeftCorner)!0.5!(topRightCorner)$) (topMiddle) {$\mathcal{I}^{k}_t \circ \dots  \circ \mathcal{I}_t^{1}$ \\ \scriptsize Reflective Observation Intrinsic Functions \\[-0.5ex] \scriptsize Abstract Conceptualisation Intrinsic Functions};

\node[box] at ($(topLeftCorner)!0.5!(bottomLeftCorner)$) (leftMiddle) {$\Sigma_t$ \\ \scriptsize Internal Agent\\[-0.5ex] \scriptsize State};
\node[anchor=west] at ($(leftMiddle.east) + (-0.23, 0)$) 
{
  \includegraphics[height=2cm]{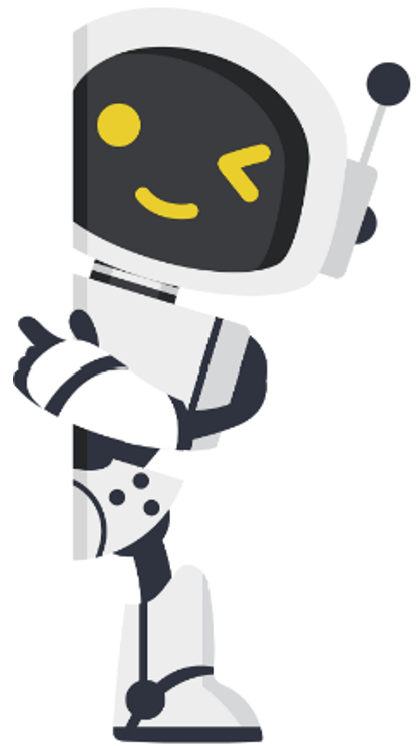}
};
\node[box, anchor=center] at ($(leftMiddle.center)$) {$\Sigma_t$ \\ \scriptsize Internal Agent\\[-0.5ex] \scriptsize State};

\node[box] at ($(topRightCorner)!0.5!(bottomRightCorner)$) (rightMiddle) {$\mathcal{E}_t\left(\Sigma^\prime_t\right)$ \\ \scriptsize Extrinsic Function\\[-0.5ex] \scriptsize Active Expert};

\node[box] at ($(bottomRightCorner)!0.5!(bottomLeftCorner)$) (bottomMiddle) {$\mathcal{U}_t\left(\Sigma^\prime_t, \mathcal{F}_t\right)$ \\ \scriptsize Internal Agent\\[-0.5ex] \scriptsize State Update};

\draw[->] (leftMiddle.north) |- (topMiddle.west);
\draw[->] (topMiddle.east) -| (rightMiddle.north);
\draw[->] (rightMiddle.south) -- (bottomRightCorner.north);
\draw[->] (bottomRightCorner.west) -- (bottomMiddle.east);
\draw[->] (bottomMiddle.west) -| (leftMiddle.south);

\end{tikzpicture}
\caption{Our computational formalisation of Kolb's experiential learning theory.}
\label{fig:exp-learning-graph}
\end{wrapfigure}

Interestingly, ReAct-like behaviour~\cite{Reflexion, Voyager, Yao2022ReAct} -- a widely used prompting strategy in which LLMs interleave reasoning (``thought'') with actions by reflecting on intermediate outcomes -- can be seen as the fundamental cognitive primitive here: a single reflect–act loop that instantiates the minimal intrinsic–extrinsic cycle. 
By chaining multiple ReAct steps, our framework naturally generalises to the full Kolb cycle of repeated reflection, abstraction and experimentation (Figure~\ref{fig:exp-learning-graph}). 
We notably show on data science problem solving that proper abstraction enables ReAct-based agents to achieve better performance with half time budget (Figure~\ref{fig:LearningCurveAll}). 
This two-phase structure mirrors the alternation in Kolb’s cycle between outward experimentation and inward conceptualisation, and provides a computational foundation for agents that learn through structured cycles of internal reorganisation and external engagement.

Specifically, the agent applies a composition of $k$ intrinsic functions, denoted by $\mathcal{I}_t^{(k)}$, to its internal state $\Sigma_t$, which corresponds to an internal memory or a summary of past experiences. 
This produces a new refined internal state $\Sigma_t^{\prime} = \mathcal{I}_t^{(k)}(\Sigma_t)$. 
Once the intrinsic phase is complete, the agent interacts with the environment via an extrinsic function $\mathcal{E}_t$, which takes $\Sigma_t^{\prime}$ as input and returns an action to apply to the environment to gather a new interaction outcome (e.g., observation, feedback, test feedback) $\mathcal{F}_t$. 
The agent then updates its internal state to $\Sigma_{t+1}$ using an update function $\mathcal{U}_t$, such that: $\Sigma_{t+1} = \mathcal{U}_t(\Sigma^{\prime}_t,\mathcal{F}_t)$. 
This formalisation preserves the alternation at the heart of Kolb’s theory, enabling the agent to reflect, abstract, and adapt before engaging with the world, thus forming a computationally grounded cycle of experiential learning as depicted in Figure~\ref{fig:exp-learning-graph}.

\begin{figure}[htbp]
\centering
\begin{tikzpicture}

\node[anchor=south] (pic3) at (0, 0)  {
    \includegraphics[width=.95\linewidth,trim={1em, 1em, 1em, 1em}, clip]{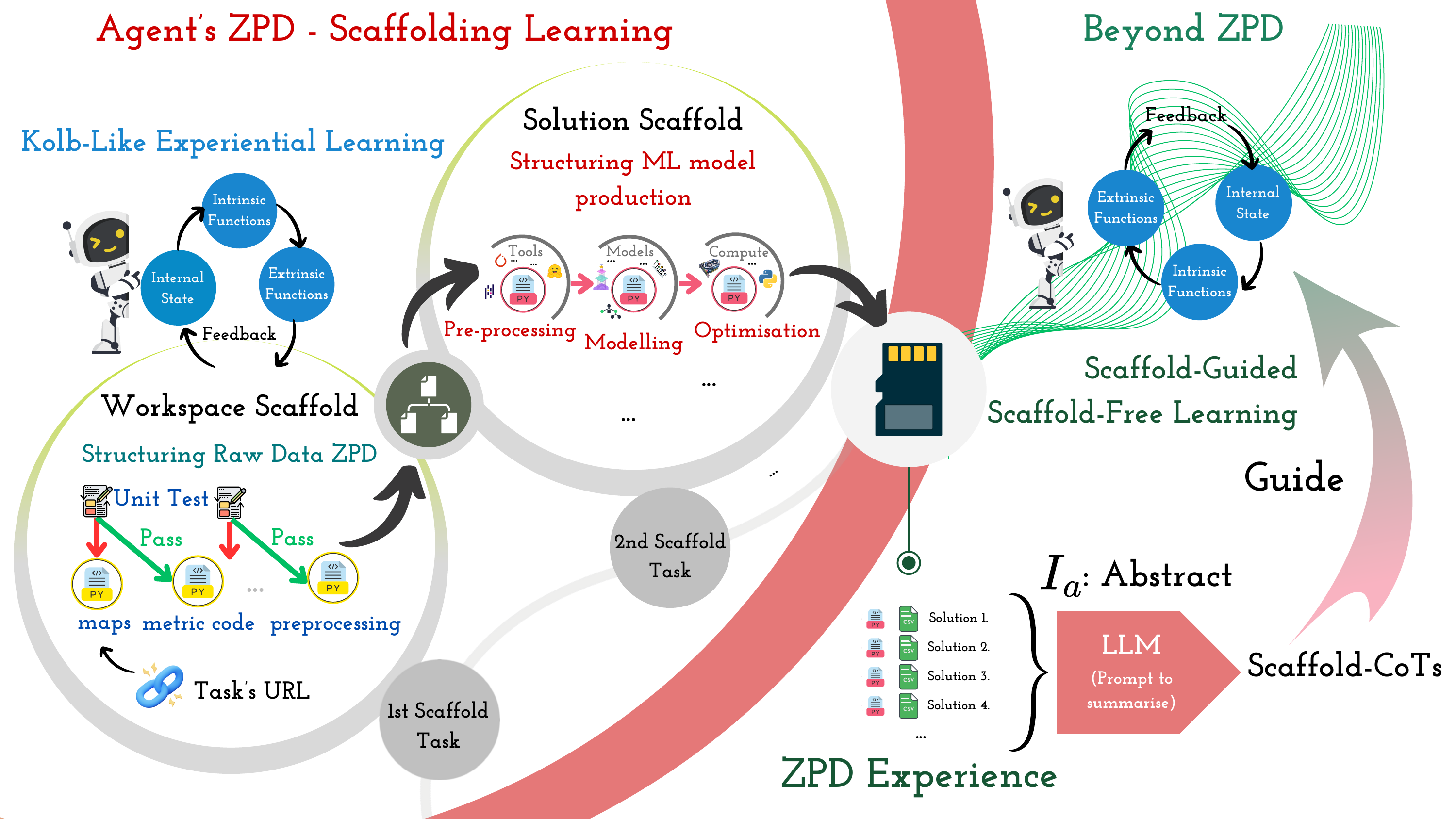}
  };
  
\node[anchor=north west] (pic1) at ($(pic3.south west) + (0cm, -.0cm)$) {
\begin{tikzpicture}

\definecolor{myforest}{RGB}{34,139,34} 

\renewcommand{\thefootnote}{\fnsymbol{footnote}}
\newcommand{\taskinfo}{\textcolor{blue}{\texttt{\textlangle task\_info\textrangle}}}
\newcommand{\tasksummary}{\textcolor{blue}{\texttt{\textlangle task\_summary\textrangle}}}
\newcommand{\dataview}{\textcolor{blue}{\texttt{\textlangle data\_view\textrangle}}}
\newcommand{\datadescr}{\textcolor{blue}{\texttt{\textlangle data\_descr\textrangle}}}
\newcommand{\errorAnalysis}{\textcolor{green!80!black}{\texttt{\textlangle error\_analysis\textrangle}}}

\newcommand{\angledsub}[2]{\langle \texttt{#1}_{#2} \rangle}
\newcommand{\planAngledsub}[1]{\textcolor{myforest!80!black}{\angledsub{plan}{#1}}}
\newcommand{\codeAngledsub}[1]{\textcolor{purple!80!black}{\angledsub{code}{#1}}}
\newcommand{\errorAngledsub}[1]{\textcolor{red!80!black}{\angledsub{error}{#1}}}
\newcommand{\errorSolveAngledsub}[1]{\textcolor{cyan!80!black}{\angledsub{error\_analysis}{#1}}}

\makeatletter
\@ifundefined{boxWidthFive}{
  \newlength{\boxWidthFive}
}{}
\@ifundefined{colOneWidthFive}{
  \newlength{\colOneWidthFive}
}{}
\@ifundefined{vertDistFormula}{
  \newlength{\vertDistFormula}
}{}
\@ifundefined{headerHeight}{
  \newlength{\headerHeight}
}{}
\@ifundefined{promptToAnswerGap}{
  \newlength{\promptToAnswerGap}
}{}
\@ifundefined{sideMargin}{
  \newlength{\sideMargin}
}{}
\@ifundefined{boxHorGap}{
  \newlength{\boxHorGap}
}{}
\@ifundefined{reflectboxWidthFive}{
  \newlength{\reflectboxWidthFive}
}{}
\makeatother

\setlength{\colOneWidthFive}{3.5cm}
\setlength{\vertDistFormula}{.1cm}
\setlength{\headerHeight}{1.em}
\setlength{\promptToAnswerGap}{.5em}
\setlength{\sideMargin}{.1cm}
\setlength{\boxHorGap}{.8cm}
\setlength{\boxWidthFive}{\dimexpr (18cm - 2\boxHorGap - 2\sideMargin - \colOneWidthFive) * 100 / 200 \relax}
\setlength{\reflectboxWidthFive}{\boxWidthFive}

\tikzset{
  prompt-node/.style={
    draw=none,
    fill=blue!10,
    rounded corners=3pt,
    text width=1.1\boxWidthFive,
    inner sep=3pt,
    font=\small,
    align=left
  },
  prompt-header/.style={
    font=\bfseries\footnotesize,
    text=white,
    fill=blue!60,
    rounded corners=0pt,
    minimum height=\headerHeight,
    text width=1.1\boxWidthFive,
    inner xsep=3pt,
    anchor=north west,
    align=left,
  },
  answer-node/.style={
    draw=none,
    fill=orange!15,
    rounded corners=3pt,
    text width=1.1\boxWidthFive,
    inner xsep=3pt,
    font=\small,
    align=left
  },
  answer-header/.style={
    font=\bfseries\footnotesize,
    text=white,
    fill=orange!60,
    rounded corners=0pt,
    minimum height=\headerHeight,
    text width=1.1\boxWidthFive,
    inner xsep=3pt,
    anchor=north east,
    align=right
  },
  console-node/.style={
    draw=none,
    fill=black,
    text=green!30,
    rounded corners=3pt,
    text width=\boxWidthFive,
    inner sep=3pt,
    font=\ttfamily\small,
    align=left
  },
  console-header/.style={
    font=\bfseries\footnotesize,
    text=white,
    fill=black!60,
    rounded corners=0pt,
    minimum height=\headerHeight,
    text width=\boxWidthFive,
    inner xsep=3pt,
    anchor=north east,
    align=left,
  },
  update-node/.style={
    draw=none,
    fill=myforest!10,
    text=black,
    rounded corners=3pt,
    text width=1.1\boxWidthFive,
    inner sep=3pt,
    align=left
  },
  update-header/.style={
    font=\bfseries\footnotesize,
    text=white,
    fill=myforest!60,
    rounded corners=0pt,
    minimum height=\headerHeight,
    text width=1.1\boxWidthFive,
    inner xsep=3pt,
    anchor=north east,
    align=left,
  },
}
  \coordinate (One) at (-.7\boxWidthFive,.5cm);
  \coordinate (Two) at (0, 1cm);
  \coordinate (Three) at (.7\boxWidthFive, .5cm);
  \coordinate (Four) at ($ (0,0) - (One) $);
  \coordinate (Five) at ($ (0,0) - (Two) $);
  \coordinate (Six) at ($ (0,0) - (Three) $);

  \draw[->, thick, black, line width=3pt]
    plot[smooth, tension=.8] coordinates {
      (One)
      (Two)
      (Three)
      (Four)
      (Five)
      (Six)
    };

\node[align=center, anchor=center] at (0, 0) { \textbf{Experiential Learning} \\ \textbf{Loop}};

\node[align=center, anchor=south east, text width=1.3\colOneWidthFive, inner sep=3pt] (sigma_t) at ($(One)+(-.5cm, .5cm)$) {
$\left\{
\begin{aligned}
  & \taskinfo, \\
  & \planAngledsub{1}, \codeAngledsub{1}, \errorAngledsub{1}, \\
  & \errorSolveAngledsub{1} \\
  & \dots,\\
  & \planAngledsub{k}, \codeAngledsub{k}, \errorAngledsub{k}
\end{aligned} 
\right\}$
};



\node[above=.2cm of sigma_t, align=center] (sigma_t_name) {\Large $\Sigma_t$ \\ Internal State};

\node[anchor=south , above=\vertDistFormula of Two, align=center, draw=none, text width=1.5\boxWidthFive] (sigma_t_1_name) {$\mathcal{I}_t^1(\Sigma_t)= \Sigma_t \cup \{ \errorSolveAngledsub{k} \}$};

\node[above=\promptToAnswerGap of sigma_t_1_name, answer-node, anchor=south] (I_r_answer) {\vspace{\headerHeight} \\  To avoid this error,  $\errorSolveAngledsub{k}$};
\node[answer-header] at ([xshift=0cm,yshift=0cm]I_r_answer.north east) {LLM Answer};

\node[prompt-node, above=\vertDistFormula of I_r_answer, anchor=south] (I_r_prompt) {
    \vspace{\headerHeight} \\
    Task info: \taskinfo \\
    This was your code: $\codeAngledsub{k}$\\ 
    Explain how to fix $\errorAngledsub{k}$, given former guidance: $\left\{\errorSolveAngledsub{i}\right\}_{i=1}^{k-1}$
};
\node[prompt-header] at ([xshift=0cm,yshift=0cm]I_r_prompt.north west) {Prompt (e.g. Error Analysis)};

\node[above=.2cm of I_r_prompt, align=center, text width=\reflectboxWidthFive, anchor=south] (I_r_name) {\Large $\mathcal{I}_t^1(\Sigma_t)$ \\ \normalsize Analyse};

\node[align=center, draw=none, text width=\boxWidthFive, anchor=south west] (sigma_prime_name)  at ($(Three)+(.0cm, .3cm)$){
    $\Sigma^\prime_t = \mathcal{I}_t^1(\Sigma_t) \cup \{\planAngledsub{k + 1}\}$
};

\node[above=\vertDistFormula of sigma_prime_name, answer-node, anchor=south] (I_a_answer) {\vspace{\headerHeight} \\  To solve this, I should $\planAngledsub{k + 1}$};
\node[answer-header] at ([xshift=0cm,yshift=0cm]I_a_answer.north east) {LLM Answer};

\node[above=\vertDistFormula of I_a_answer, prompt-node, anchor=south] (I_a_prompt) {
    \vspace{\headerHeight} \\
    Task info: \taskinfo \\
    You wrote $\codeAngledsub{k}$ and got $\errorAngledsub{k}$.\\
    Given this error and former guidance $\left\{\errorSolveAngledsub{i}\right\}_{i=1}^{k}$, write a plan...
};
\node[prompt-header] at ([xshift=0cm,yshift=0cm]I_a_prompt.north west) {Prompt (e.g. New Planning)};

\node[above=\vertDistFormula of I_a_prompt, align=center, text width=\reflectboxWidthFive, anchor=south] (I_a_name) {\Large $\mathcal{I}_t^2(\mathcal{I}_t^1(\Sigma_t))$ \\ \normalsize Plan
};

\node[align=center, text width=\reflectboxWidthFive, anchor=north west] (external_act_name)   at ($(Three)+(.0cm, .02cm)$){\Large $\mathcal{E}_t(\Sigma_t^{\prime})$ \\\normalsize External action};

\node[below=.2cm of external_act_name, prompt-node] (external_act_prompt) {
    \vspace{\headerHeight} \\
    Task info: \taskinfo \\
    You wrote $\codeAngledsub{k}$ and got $\errorAngledsub{k}$.\\
    Considering $\left\{\errorSolveAngledsub{i}\right\}_{i=1}^{k}$,
    implement a solution given $\planAngledsub{k+1}$
};
\node[prompt-header] at ([xshift=0cm,yshift=0cm]external_act_prompt.north west) {Prompt (e.g. Implement)};

\node[below=\promptToAnswerGap of external_act_prompt, answer-node] (external_act_answer) {\vspace{\headerHeight} \\  \texttt{\`{}\`{}\`{}python}\\ $\codeAngledsub{k+1}$ \\ \`{}\`{}\`{}};
\node[answer-header] at ([xshift=0cm,yshift=0cm]external_act_answer.north east) {LLM Answer};

\node[below=.7\promptToAnswerGap of external_act_answer, align=center, draw=none, text width=\reflectboxWidthFive] (z_t_name) {$z_t := \mathcal{E}_t(\Sigma_t^\prime) = \codeAngledsub{k+1}$};

\node[anchor=north , below right=\vertDistFormula and -1.25cm of Five, align=center] (env_feedback_name) {\Large $F_t(\mathcal{T}_t, z_t)$ \\Get env. feedback};

\node[console-node, below=\vertDistFormula of env_feedback_name] (console){\vspace{\headerHeight}\\
\textcolor{green!60}{user@machine:\textasciitilde\$} python script.py \\
Traceback [...] \\
\textcolor{red}{ValueError: 'NaN' encountered}
};
\node[console-header] at ([xshift=0cm,yshift=0cm]console.north east) {Env. (e.g. Run $z_t$ in a Terminal)
};

\node[below=\vertDistFormula of console, align=center, draw=none] (f_t_formula) {$\mathcal{F}_t =\{\codeAngledsub{k+1}, \errorAngledsub{k+1} \}$};

\node[align=center, anchor=north east, align=center] (update_func) at ($(Six)+(-.5cm, .5cm)$) {\Large $\mathcal{U}_t(\Sigma_t^\prime, \mathcal{F}_t)$ \\ \normalsize Update the state};

\node[update-node, align=center, below=\vertDistFormula of update_func] (update_rule){\vspace{\headerHeight}\\$\Sigma_{t+1}:=\Sigma_t^\prime \cup \{\codeAngledsub{k+1}, \errorAngledsub{k+1} \}$
};
\node[update-header] at ([xshift=0cm,yshift=0cm]update_rule.north east) {Update (e.g. Store Elements)
};


\end{tikzpicture}
};

\node[anchor=north east] (pic2) at ($(pic3.south east) + (0cm, -.0cm)$) {
  \begin{tikzpicture}[every node/.style={font=\small}]

\definecolor{myforest2}{RGB}{34,139,34} 

\renewcommand{\thefootnote}{\fnsymbol{footnote}}
\newcommand{\tasksummary}{\textcolor{blue}{\texttt{\textlangle task\_summary\textrangle}}}
\newcommand{\dataview}{\textcolor{blue}{\texttt{\textlangle data\_view\textrangle}}}
\newcommand{\datadescr}{\textcolor{blue}{\texttt{\textlangle data\_descr\textrangle}}}

\newcommand{\angledsub}[2]{\langle \texttt{#1}_{#2} \rangle}
\newcommand{\planAngledsub}[1]{\textcolor{myforest2!80!black}{\angledsub{plan}{#1}}}
\newcommand{\codeAngledsub}[1]{\textcolor{cyan!80!black}{\angledsub{code}{#1}}}
\newcommand{\logAngledsub}[1]{\textcolor{red!80!black}{\angledsub{run\_log}{#1}}}
\newcommand{\scaffoldCoT}[1]{\textcolor{magenta!80!black}{\angledsub{scaffold\_cot}{#1}}}
\newcommand{\solPool}{\textcolor{black}{\texttt{\textlangle submit\_pool}}}

\newlength{\maxWidth}
\setlength{\maxWidth}{13cm}
\newlength{\boxWidthTwo}
\newlength{\sideMarginBis}
\setlength{\sideMarginBis}{0cm}
\newlength{\boxHorGapBis}
\setlength{\boxHorGapBis}{0cm}
\newlength{\colOneWidthTwo}
\setlength{\colOneWidthTwo}{\dimexpr (\maxWidth - 2 \sideMarginBis - \boxHorGapBis) * 110 / 300 \relax}
\newlength{\vertDistFormulaBis}
\setlength{\vertDistFormulaBis}{.1cm}
\newlength{\headerHeightBis}
\setlength{\headerHeightBis}{1.em}
\newlength{\promptToAnswerGapBis}
\setlength{\promptToAnswerGapBis}{.5em}
\setlength{\boxWidthTwo}{\dimexpr (\maxWidth - \boxHorGapBis - 2\sideMarginBis - \colOneWidthTwo + .9cm) \relax}

\tikzset{
  prompt-node/.style={
    draw=none,
    fill=blue!10,
    rounded corners=3pt,
    text width=\boxWidthTwo,
    inner sep=3pt,
    font=\small,
    align=left
  },
  prompt-header/.style={
    font=\bfseries\footnotesize,
    text=white,
    fill=blue!60,
    rounded corners=0pt,
    minimum height=\headerHeightBis,
    text width=\boxWidthTwo,
    inner xsep=3pt,
    anchor=north west,
    align=left,
  },
  answer-node/.style={
    draw=none,
    fill=orange!15,
    rounded corners=3pt,
    text width=\boxWidthTwo,
    inner xsep=3pt,
    font=\small,
    align=left
  },
  answer-header/.style={
    font=\bfseries\footnotesize,
    text=white,
    fill=orange!60,
    rounded corners=0pt,
    minimum height=\headerHeightBis,
    text width=\boxWidthTwo,
    inner xsep=3pt,
    anchor=north east,
    align=right
  },
  console-node/.style={
    draw=none,
    fill=black,
    text=green!30,
    rounded corners=3pt,
    text width=1.2\colOneWidthTwo,
    inner sep=3pt,
    font=\ttfamily\small,
    align=left
  },
  console-header/.style={
    font=\bfseries\footnotesize,
    text=white,
    fill=black!60,
    rounded corners=0pt,
    minimum height=\headerHeightBis,
    text width=1.2\colOneWidthTwo,
    inner xsep=3pt,
    anchor=north east,
    align=left,
  },
  update-node/.style={
    draw=none,
    fill=myforest2!10,
    text=black,
    rounded corners=3pt,
    text width=.93\boxWidthTwo,
    inner sep=3pt,
    align=left
  },
  update-header/.style={
    font=\bfseries\footnotesize,
    text=white,
    fill=myforest2!60,
    rounded corners=0pt,
    minimum height=\headerHeightBis,
    text width=.93\boxWidthTwo,
    inner xsep=3pt,
    anchor=north east,
    align=left,
  },
}
  \coordinate (One) at (\colOneWidthTwo, .3cm);
  \coordinate (Two) at (\dimexpr \maxWidth / 2\relax, 1cm);
  \coordinate (Three) at (\dimexpr \maxWidth  - \colOneWidthTwo, 0cm);
  \coordinate (Four) at (\dimexpr \maxWidth / 2\relax, -1cm);
  \coordinate (Five) at ($ (0,-.6cm) + (One) $);

  \draw[->, thick, black, line width=3pt]
    plot[smooth, tension=1] coordinates {
      (One)
      (Two)
      (Three)
      (Four)
      (Five)
    };

\node[align=center, anchor=center] at (.5\maxWidth, 0) { \textbf{Exp. Learning} \\ \textbf{Loop} };

\node[align=center, anchor=south east, text width=\colOneWidthTwo, inner sep=3pt] (sigma_t) at ($(One)+(-.7cm, .2cm)$) {
$\left\{
\begin{aligned}
  &\tasksummary,  \\
  &\datadescr, \\
  &\hspace{2.5em}\dots \\ 
  &\left\{\scaffoldCoT{i}\right\}_{i=1}^{k},\\
  &\solPool:  \emptyset \texttt{\textrangle}
\end{aligned} 
\right\}$
};

\node[above=.2cm of sigma_t, align=center] (sigma_t_name) {\Large $\Sigma_t$ \\ Internal State};

\node[anchor=south , align=center, draw=none]  (z_t) at ([xshift=2.3cm,yshift=.2cm]Two) {$z_t:= \mathcal{E}_t(\mathcal{I}_t(\Sigma_t)) = \codeAngledsub{}$};

\node[above=\vertDistFormulaBis of z_t, answer-node, anchor=south] (I_E_answer) {
    \vspace{\headerHeightBis} \\  
    ``A first attempt to tackle this ... $\planAngledsub{}$\\ \texttt{\`{}\`{}\`{}python}\\ 
    $\codeAngledsub{}$ \\ \`{}\`{}\`{}''\\  
    
};
\node[answer-header] at ([xshift=0cm,yshift=0cm]I_E_answer.north east) {LLM Answer};

\node[prompt-node, above=\promptToAnswerGapBis of I_E_answer, anchor=south] (I_E_prompt) {
    \vspace{\headerHeightBis} \\ 
    Task info: \tasksummary, \datadescr, ...\\
    Past submissions attempts: $\left\{\scaffoldCoT{i}\right\}_{i=1}^{k}$. \\
    Generate a plan followed by a solution in python.  
};
\node[prompt-header] at ([xshift=0cm,yshift=0cm]I_E_prompt.north west) {Prompt (Abstract Scaffold CoT, Plan \& Write code)};

\node[above=.2cm of I_E_prompt, align=center, text width=\boxWidthTwo, anchor=south] (I_E_name) {\Large $\mathcal{I}_t$ \& $\mathcal{E}_t$ \\ \normalsize Abstraction \& External action};

\node[anchor=north, below right=.0cm and .75cm of Three, align=center] (env_feedback_name) {\Large $F_t(\mathcal{T}_t, z_t)$ \\Get env. feedback};

\node[console-node, below=\vertDistFormulaBis of env_feedback_name] (console){\vspace{\headerHeightBis}\\
\textcolor{green!60}{user@agk:\textasciitilde\$} python script.py \\
Training model\dots \\
$\vdots$\\
Validation MSE: 0.324
};
\node[console-header] at ([xshift=0cm,yshift=0cm]console.north east) {Env. (e.g. Execute $\codeAngledsub{}$)
};

\node[below=\vertDistFormulaBis of console, align=center, draw=none] (f_t_formula) {$\mathcal{F}_t := F_t(\mathcal{T}_t, z_t) =\{\logAngledsub{} \}$};

\node[align=center, anchor=north east, align=center] (update_func) at ($(Five)+(-.0cm, .4cm)$) {\Large $\mathcal{U}_t(\Sigma_t^\prime, \mathcal{F}_t)$ \\ \normalsize Internal State Update};

\node[update-node, align=left, below right=\vertDistFormulaBis and .5cm of update_func.south, anchor=north] (update_rule){
\vspace{\headerHeightBis}\\ 
\ \textbullet\  $\Sigma_{t+1} \leftarrow \Sigma_t^\prime$ \\
\ \textbullet\  $\Sigma_{t+1}[\solPool\texttt{\textrangle}]~\leftarrow~\{\codeAngledsub{},\logAngledsub{} \}$
};
\node[update-header] (updateHeader) at ([xshift=0cm,yshift=0cm]update_rule.north east) {Update (e.g. Add solution to the pool)
};

\end{tikzpicture}
};

  \draw[->, line  width=2pt, color={rgb,255:red,7; green,138; blue,199}] (-7.3, 6) .. controls  (-8.4, 5) and (-8.7, 1) .. (-7, -.15);
  \draw[->, line  width=2pt, color={rgb,255:red,7; green,138; blue,199}] (8, 4) .. controls (8.5, 3.5) and  (8.5, 1) .. ([xshift=-.1cm, yshift=-.3cm]pic2.north east);

\end{tikzpicture}
\caption{\textbf{From Scaffolded Experiential Learning to Autonomous Generalisation.} 
The top part of the figure shows how an autonomous agent progresses from scaffolded learning tasks within its Zone of Proximal Development (ZPD) to open-ended problem solving. 
In the scaffolded environment (on top left), 
the agent generates solutions though structured tasks gated by success and supported by feedback.
As the agent masters scaffolded tasks, it internalises strategies into Scaffold-CoTs -- realised through LLM summarisation in our setup. 
In the open-ended environment (on the top right), the scaffold is removed,
and the abstracted knowledge supports self-directed adaptation to increase the likelihood of success.
Learning in both regimes follows our computational model of Kolb’s experiential learning cycle: concrete interaction with the environment (extrinsic functions), reflective observation and internal strategy formation (intrinsic functions), and active experimentation based on revised hypotheses. 
The two bottom graphs illustrate this experiential learning process via prompt-based intrinsic and extrinsic functions.
The left graph displays an experiential learning loop for error solving during scaffold, while the right loop shows how the agent abstracts scaffold-CoTs to generate open-ended solutions.
}
\label{fig:ExpLearningEnv}
\end{figure}    

Now that we have introduced a computational framework for Kolb’s experiential learning theory, we turn our attention to formalising Vygotsky-inspired scaffolded learning, which underpins the agent’s developmental trajectory.
Together, the aforementioned Kolb-based computational framework and the guided progression through increasingly challenging environments define the core of our approach.

\section*{Scaffolded Learning and the Agent's ZPD}

Inspired by Vygotsky’s Zone of Proximal Development (ZPD)~\cite{Vygotsky1978}, we introduce the concept of the Agent's ZPD as the range of task complexity where an agent cannot yet succeed autonomously, but can succeed with appropriate scaffolded support. This region defines the agent’s learning frontier: the space where internal adaptation is still possible, provided that the environment offers the right structure, feedback, or constraints. Just as human learners grow most effectively when challenged, agents benefit from carefully structured experiences that push their boundaries without overwhelming them.

To make the above concept concrete, we define scaffolded learning as a structured progression over a set of tasks or environments $\mathcal{T}=\{\mathcal{T}_1,\mathcal{T}_2, \dots, \mathcal{T}_n\}$, which are staged to loosely reflect the natural workflow a practitioner might follow when approaching the problem domain. These correspond to the scaffolded subtasks shown on the left-hand side of Figure \ref{fig:ExpLearningEnv} labelled ``Agent's ZPD - Scaffolding Learning'' and ``Beyond ZPD''. 
Each task $\mathcal{T}_i \in \mathcal{T}$ is designed to target a specific capability or reasoning skill, and to build on knowledge acquired in earlier stages incrementally. 
This staged progression resembles curriculum learning, in that  transitions are competency-gated and heuristically adjusted~\cite{hu2020learning, willems2020masteringratebasedcurriculum}. 
The agent advances only by satisfying explicit success criteria (e.g., passing unit tests), making progress contingent on demonstrated ability. This structure aligns more closely with the ZPD, where support is withdrawn as internal competence emerges.

Additionally, we introduce a conceptual feedback function $F\left(\mathcal{T}_i, \Sigma_t\right)$ that evaluates the agent's performance on task $\mathcal{T}_i$ given its internal state $\Sigma_t$. 
This feedback may take the form of explicit performance signals, such as scalar rewards (as in reinforcement learning), binary success indicators, or richer environmental responses, and serves to guide internal adaptation throughout the learning cycle. As the agent progresses through $\mathcal{T}$, it iteratively updates its internal state through experiential interaction, enabling performance on increasingly complex environments without direct supervision or fine-tuning. This formal structure allows us to represent scaffolding as a trajectory through a task space structured by cognitive dependencies between tasks, such as needing to align data modalities before attempting predictive modelling in data science.

After completing the scaffolded stages, the agent transitions into a fully autonomous, open-ended learning phase; see the right-hand side of Figure \ref{fig:ExpLearningEnv}. To initiate this process, the agent consolidates its prior scaffolded experience by abstracting patterns from previously constructed pipelines within the scaffold. These past solutions, generated through structured experiential learning, are internally abstracted into chain-of-thought traces that capture summaries of successful reasoning steps, component structure, and validation logic. Interestingly, these traces do not merely support reasoning in similar future tasks. They function as autonomous, agent-generated cognitive scaffolds, effectively bootstrapping a chain-of-thought process in the absence of external guidance.

In the open-ended setting, the agent reuses these internalised traces as cognitive scaffolds: they guide hypothesis formation, code synthesis, and self-evaluation in tasks where no external structure is provided. From this point onward, the agent continues applying experiential learning principles independently, completing the full cycle of action, reflection, abstraction, and adaptation. This developmental path parallels that of human learners, who first acquire skills within structured environments and later apply them autonomously once scaffolds are removed \cite{VandePol2010, Wood1976}.

\section*{\OurAgent{}: Integrating Kolb's and Vygotsky's Principles to Master Kaggle Competitions}

We instantiate our framework in \OurAgent{}, the first effective AI agent to tackle autonomous data science, which is the end-to-end process of generating high-performing solutions from raw datasets and natural language problem descriptions, without human intervention.
This setting combines strategic reasoning, iterative experimentation, and abstraction across heterogeneous data types, including tabular data, computer vision, natural language, and even multi-modal domains. 
We ground our work in Kaggle competitions, which require not only technical skills but also strategic generalisation across a wide range of domains.
Kaggle presents a particularly challenging benchmark due to a combination of concrete technical difficulties: the diverse and often loosely documented file structures of competition datasets, the high risk of overfitting to small public leaderboard splits, and the need to match or outperform expert human data scientists who leverage advanced ensembling and domain-specific modelling techniques.
These factors make Kaggle a uniquely rigorous testbed for real-world, autonomous data science.

\subsection*{Scaffolded Data Science Environments}
We begin by designing the ZPD of \OurAgent{}, defining a structured learning environment where the agent can succeed with scaffolded support, but not yet independently. This scaffolded phase forms the first stage of \OurAgent{}’s developmental arc. It prepares the agent to later operate autonomously in open-ended settings (generating end-to-end data science workflows), where external guidance is removed.
Just as a teacher structures learning to progressively build students’ capabilities, our scaffolded environment guides \OurAgent{} through staged components of the data science workflow. 
This setting allows exploration, hypothesis formation, and skill acquisition, enabling the agent to learn through structured experience.

Our scaffolded environment mirrors the human data science process: first, understanding the task and structuring a workspace, then solving the problem.
Reflecting this progression, the environment is organised into two phases, as shown in Figure~\ref{fig:ExpLearningEnv}: an initial setup phase focused on data abstraction and exploration, and a solution-building phase focused on modelling and strategic optimisation. 

While our scaffolded environment offers structure, navigating it remains a non-trivial challenge. 
\OurAgent{} receives only high-level task descriptions or templates (not detailed implementations), and must independently determine how to construct each component. 
For example, it may recognise the need for a performance metric or submission interface, but must devise the logic and implementation itself. 
This mirrors how human learners are often given structured guidance while still needing to solve problems through reasoning, experimentation, and adaptation.

\subsubsection*{Stage I: Workspace Scaffold}
The first phase of our environment (denoted by ``Workspace Scaffold in Figure~\ref{fig:ExpLearningEnv}) mirrors the early, often ambiguous steps human data scientists take: transforming potentially messy, real-world inputs into structured workspaces. 
For \OurAgent{}, this requires inferring meaningful abstractions from diverse inputs  (i.e. text, images, and tables) and adapting to varied outputs, from classification to regression and ranking.
Even within a single task type, the agent must reason about output semantics (e.g., class probabilities vs. hard labels) and align input-output mappings accordingly.
This stage challenges the agent to develop a functional understanding of the task from minimal supervision, enabling it to transform raw, heterogeneous inputs into a coherent, structured workspace suitable for downstream modelling.

To generalise across diverse data science tasks, \OurAgent{} must learn to construct unified representations from inconsistent inputs, typically comprising a labelled training set and an unlabelled test set, distributed across multiple modalities.
Here, the agent autonomously generates code to align inputs with expected outputs and define task-specific evaluation criteria. 
These components include input–output mappings, transformation routines, and formatting logic for predictions; see Figure \ref{fig:env-scaffold} in the Methods section for implementation details.

\OurAgent{} advances through scaffolded learning stages only when its generated solutions satisfy general-purpose tests (e.g., pass/fail validations or execution traces), provided by the environment as feedback signals. 
These constraints validate properties such as data alignment, execution correctness, and inter-component consistency. 
When a stage fails, the agent revises its internal strategy and attempts the step again.
The environment evaluates not just isolated components but their combined behaviour across multiple stages, allowing for recursive correction and consolidation.
This mechanism supports learning through trial, failure, reflection, and abstraction, mirroring experiential learning cycles in structured problem-solving environments.

\subsubsection*{Stage II: Solution Generation Scaffold}
With the workspace constructed, the environment transitions to the second phase: solution generation (see the ``Solution Scaffold'' part in Figure~\ref{fig:ExpLearningEnv}). 
In this stage, \OurAgent{} takes autonomous steps toward building a complete task-specific solution: designing modelling strategies, engineering features, training models, and refining performance through iteration. 
This stage supports experimentation by allowing the agent to explore multiple approaches, revise underperforming solutions, and learn from feedback signals that emerge through training and evaluation.

The environment exposes \OurAgent{} to reference patterns drawn from common practices in modern data science, such as feature encoders, hyperparameter tuning strategies, and domain-specific model families, as further explained in the Methods section. 
However, the agent does not reuse these as static templates. 
Instead, it must interpret their structure, adapt them to the task context, and implement viable solutions in code. 

\subsection*{Experiential Learning in \OurAgent{}}
\OurAgent{}’s learning process is grounded in the same experiential learning loop, inspired by Kolb’s theory as detailed before. This loop alternates between extrinsic functions, where the agent interacts with its environment, and intrinsic functions, where it reflects, abstracts, and adapts its internal strategies, as summarised in Table ~\ref{tab:agentInstrinsicExtrinsic} in the Methods section.

\paragraph{Scaffolded Intrinsic Functions:}
In the scaffolded setting, intrinsic functions are triggered by feedback such as unit test failures or low validation scores. The agent uses LLM-based reasoning loops to reflect on these signals, identify the source of failure, and revise its internal plans. These cycles may involve summarising and abstracting console logs, identifying likely bugs, or proposing new solution strategies, all performed autonomously through iterative prompt completions, as we show in Figure~\ref{fig:exp-learning-in-scaffold}.

\paragraph{Open-Ended Intrinsic Functions}
In the open-ended phase, the agent operates without structural constraints, aiming to autonomously generate complete data science solutions from raw inputs. 
While it continues to follow Kolb’s experiential learning loop, still alternating between action, intrinsic processes, and adaptation, it now builds on the internal knowledge acquired during the scaffolded phase. Specifically, intrinsic functions are enhanced with LLM-based summarisation/abstraction mechanisms that distil prior experiences into high-level conceptual traces. These distilled summaries are repurposed as chain-of-thought prompts, guiding hypothesis generation, strategy formation, and iterative debugging. 

The agent begins by using these chain-of-thought prompts to propose an initial set of candidate solutions. Each candidate forms the root of a dynamically constructed tree of code, where each node represents a fully executable data science pipeline. After execution, the agent evaluates feedback, such as validation scores or runtime errors, and decides how to evolve the tree. It may refine an existing node to fix bugs or generate a new variant to improve performance. 

This approach goes beyond simple ReAct-style agents such as AIDE \cite{Jiang2024AIDE_ML_WECO}, which rely solely on LLM pretraining and generic ReAct loops.
By contrast, \OurAgent{} leverages scaffold-derived knowledge to seed its reasoning with abstractions from domain-specific insights, resulting in more focused exploration and substantially improved performance. Rather than improvising from scratch, the agent bootstraps its learning from prior conceptualisations, demonstrating how structured experiential learning can scale to autonomous generalisation in complex, unconstrained settings.

\section*{Quantitative Results} \label{Sec:BuildingBenchmark}
We evaluated \OurAgent{} on \NumBenchmarkTasks{} real-world Kaggle competitions spanning tabular (\TabularTasks{}\%), computer vision (\ComputerVision{}\%), natural language (\NaturalLanguage{}\%), and multimodal (\MultiModal{}\%) tasks. 
Unlike benchmarks that target isolated aspects of data science (e.g., tabular-only tasks \cite{erickson2025tabarena} or hyperparameter tuning \cite{Eggensperger2021HPOBench}), our benchmark tests end-to-end generalisation across the full pipeline. 
It enables cross-domain evaluation using standardised Kaggle leaderboard submissions, assessing both autonomy and predictive performance in practical, real-world settings, surpassing the scope and fidelity of prior benchmarks.

To ensure a meaningful evaluation, we selected Kaggle competitions with high human participation, averaging over \AvgPartTab\ participants in tabular tasks, \AvgPartNlp\ in NLP, and \AvgPartMulti\ in multimodal domains. The benchmark includes a balanced mix of accessible and challenging tasks, ranging from Kaggle Playground competitions to high-stakes featured and research challenges, which are widely regarded as the platform’s most competitive and demanding. 
We evaluated \OurAgent{} under the same conditions as human Kaggle participants. 
It interacts with the Kaggle API to submit predictions and is ranked on the private leaderboard, enabling direct, transparent comparison with both human data scientists and existing automated systems.

In addition to reporting performance quantiles, we evaluated whether \OurAgent{} would earn gold, silver, or bronze medals using Kaggle’s official criteria. Following standard benchmarking practice \cite{chan2024mlebenchevaluatingmachinelearning, Jiang2024AIDE_ML_WECO}, we computed medals even for competitions that did not officially award them. 
However, we clearly distinguish between official and inferred medals in our reporting and apply the same rules to human participants to ensure a fair and transparent comparison.
\subsection*{\OurAgent{}'s Medal Performance}
\begin{figure}[t!]
\centering
\includegraphics[trim={0em 1.6em 1.2cm 9em},clip, height=.75\textheight]{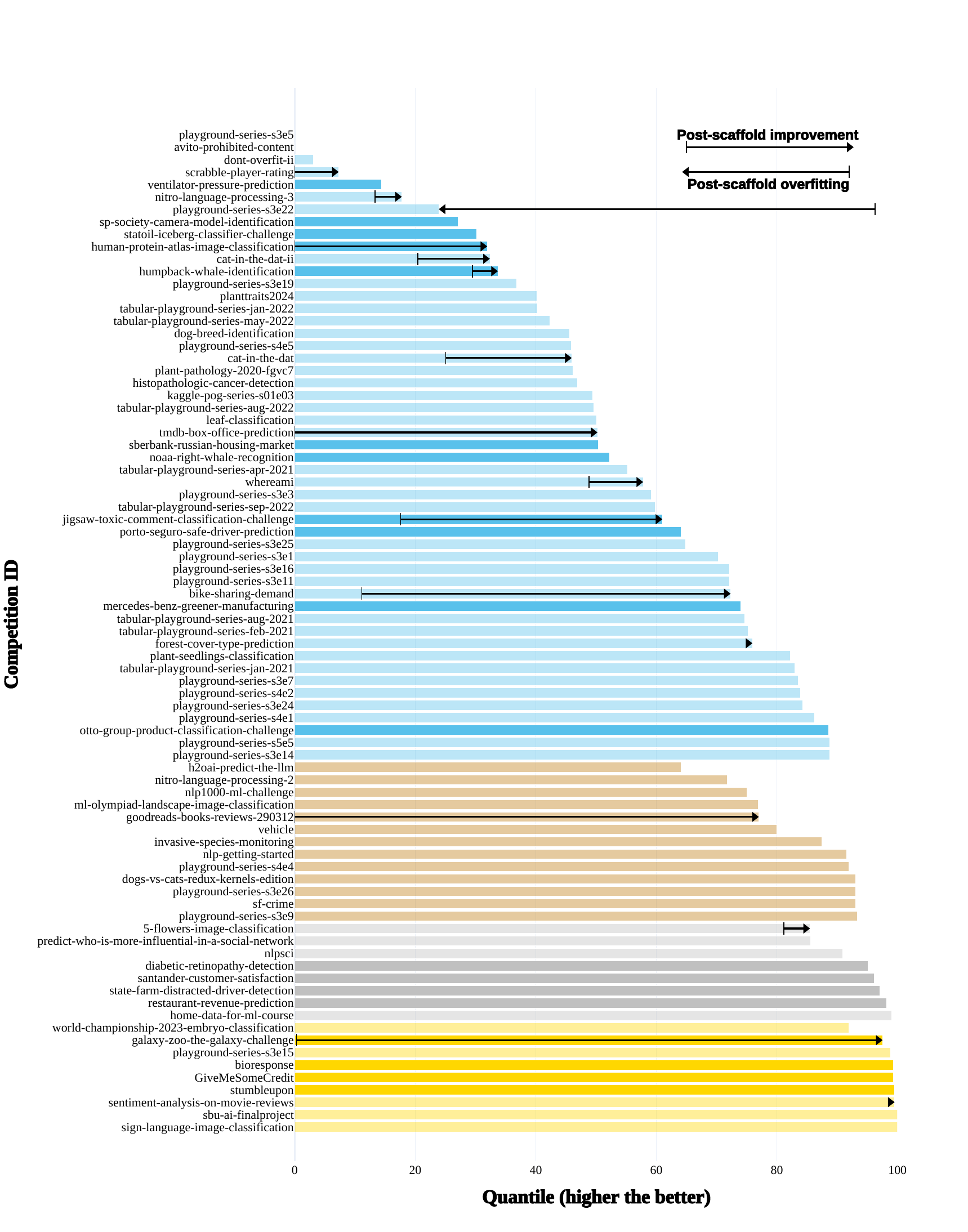}
\caption{\OurAgent{}’s performance across Kaggle competitions spanning tabular, computer vision, NLP, and multimodal tasks. The y-axis lists competition IDs; the x-axis shows quantile performance on the private leaderboard (higher is better). Bars with darker shading correspond to Kaggle competitions that granted actual medals as featured or research competitions.  On these, \OurAgent{} would earn \nGoldMedalsOnMedAwardingComps\ gold and \nGoldMedalsOnMedAwardingComps\ silver medals, and achieve the equivalent of \nGoldMedalsNonMedAwardingComps\ gold, \nSilverMedalsNonMedAwardingComps\ silver, and \nBronzeMedalsNonMedAwardingComps\ bronze medals in others.}
\label{fig:barRes}
\end{figure}

Figure~\ref{fig:barRes} summarises \OurAgent{}’s performance across Kaggle’s private leaderboards. The agent earned the equivalent of \nGoldMedalsOnMedAwardingCompstext\ gold and \nSilverMedalsOnMedAwardingCompstext\ silver in real medal-awarding competitions spanning tabular, computer vision, natural language, and multimodal tasks. 
These medals were awarded in challenges with up to \TotalParti+ participants and prize pools as high as \$\PrizeMoney\, underscoring both the competitiveness and practical difficulty of the tasks. For example, \OurAgent{} achieved gold in ``Galaxy Zoo'' (computer vision), ``Give Me Some Credit'' (tabular), and in the multimodal challenge ``Stumble Upon''. 

Beyond the featured competitions, \OurAgent{} achieved medal-equivalent rankings in a broad set of non-medal-awarding tasks, earning \nGoldMedalsNonMedAwardingCompstext\ gold, \nSilverMedalsNonMedAwardingCompstext\ silver, and \nBronzeMedalsNonMedAwardingCompstext\ bronze equivalents across tabular, computer vision, and natural language domains. 
These results further demonstrate its versatility and generalisation across diverse modalities. Notably, these competitions included large-scale challenges such as the ``Sentiment Analysis on Movie Reviews'' (NLP; 1,011 participants) and ``House Prices for ML Course'' (tabular; 6,999 participants). 
A full breakdown is provided in Extended Data in the Methods section.

Taken together, these results show that \OurAgent{} can compete at a high level, earning medals in featured Kaggle competitions with substantial prize pools and large participant pools. 
Beyond these, it achieves strong, medal-equivalent rankings across a wide range of additional challenges.
Notably, \OurAgent{} demonstrates versatility not only within individual domains but also across the full spectrum of data science tasks, including tabular data, computer vision, natural language processing, and multimodal problems. 
This breadth of performance provides compelling evidence of its general capabilities. 

But a central question remains: \textit{how does \OurAgent{} compare to human data scientists?} We turn to this next.
\subsection*{\OurAgent{} versus Human Data Scientists}

\begin{figure}[h!]
\centering
\includegraphics[width=.98\linewidth,trim={.1cm 0.cm 2cm 1cm},clip]{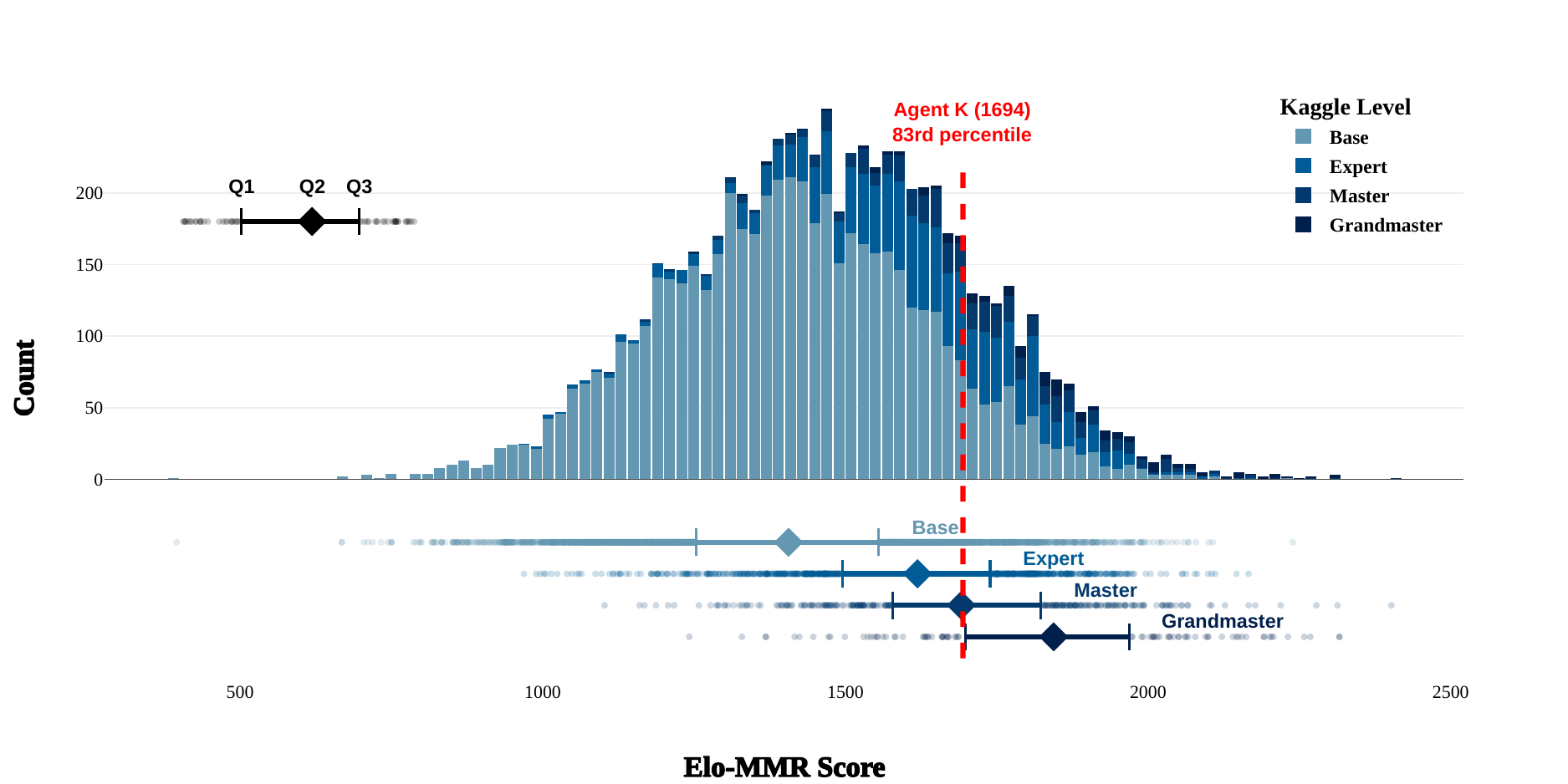}
\caption{Comparison of \OurAgent{}'s Elo-MMR score with that of human participants. The top plot shows the Elo-MMR distribution of Kaggle users who participated in at least three of the same competitions (\nParticipantsEloComparison{} in total). \OurAgent{} ranks within the top \topPercentile\% of this group. 
Bar colours reflect users’ Kaggle levels at the time of writing. The lower panel breaks down Elo-MMR scores by Kaggle level. 
}
\label{fig:elo-mmr-scoreh}
\end{figure}

To compare \OurAgent{}'s performance to human data scientists, we apply a multiplayer Elo rating system following the approach in~\cite{ebtekar2021elo}. 
Elo provides a principled way to compare agents across tasks of varying difficulty and participant pools by modelling performance as a series of head-to-head matchups. 
It is widely used for ranking in competitive settings and has been adopted by large-scale platforms such as CodeChef, reinforcing its relevance as a robust and interpretable benchmark for AI systems. 
We identify participants who competed in at least three of the same competitions as \OurAgent{}, yielding a pool of \nParticipantsEloComparison{} ``active'' human competitors. 
This group spans a diverse range of Kaggle skill levels, from base Kagglers to competition Grandmasters (\nGrandmasterForElo{}).

The histogram in Figure~\ref{fig:elo-mmr-scoreh} summarises those results. 
The x-axis represents the Elo-MMR scores, while the y-axis shows the number of participants at each score level. 
The Elo-MMR scores follow an approximately normal distribution, peaking between 1400 and 1500, where most participants are concentrated. 
A red dashed line marks our agent’s Elo-MMR score at \AgentKEloscore, placing it in the top \topPercentile th percentile.
This means our agent outperforms about \outperformspercent\% of the \nParticipantsEloComparison{} participants in the dataset.
The lower section of Figure~\ref{fig:elo-mmr-scoreh} presents a more detailed view of Elo-MMR distributions for participants grouped by Kaggle levels, from Base Kaggler to Grandmasters.
At the intersection of the red line and these distributions, we see that \OurAgent{}'s Elo-MMR score falls slightly beyond the median score achieved by Master-level participants.

\subsection*{\OurAgent{}’s Experiential Learning in Competitive Contexts}

\begin{figure}[t!]
\centering
\includegraphics[width=\linewidth,trim={1cm 0cm 1cm 3.5cm},clip]{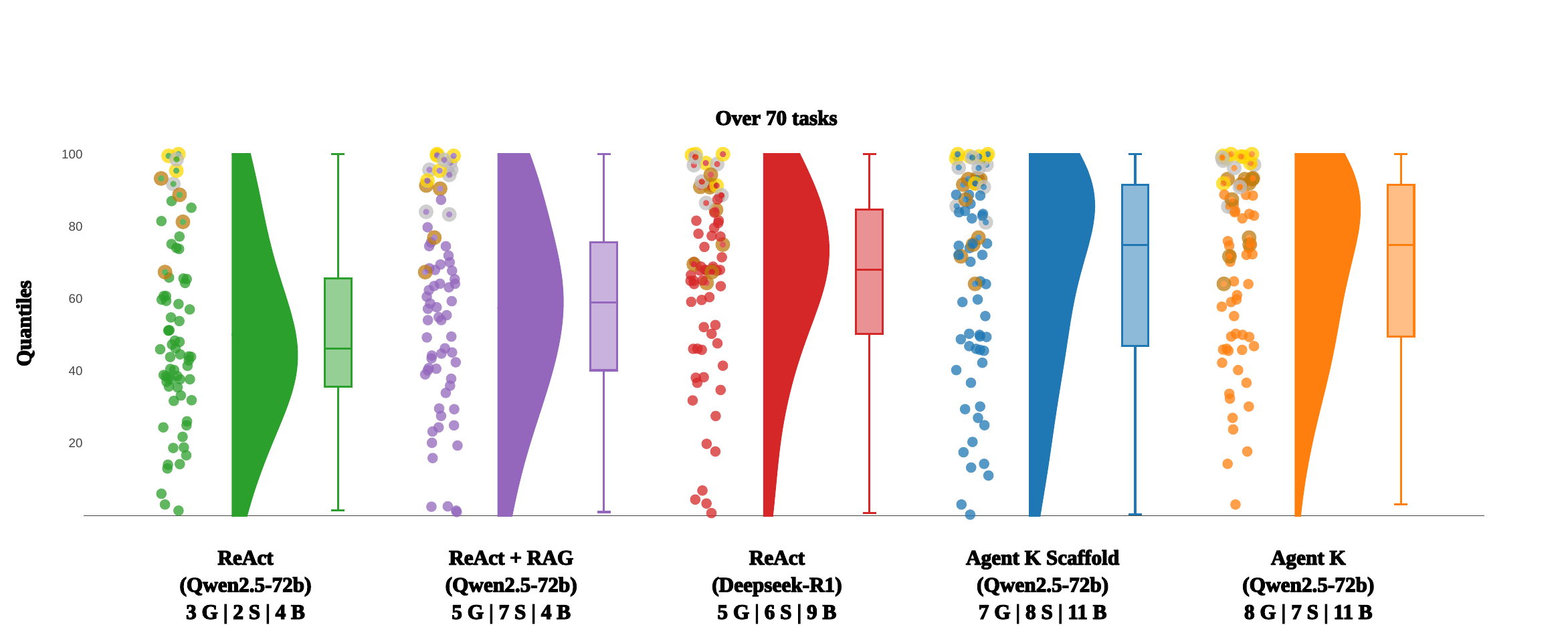}
\includegraphics[width=.9\linewidth,trim={0cm 0cm 0cm 0cm},clip]{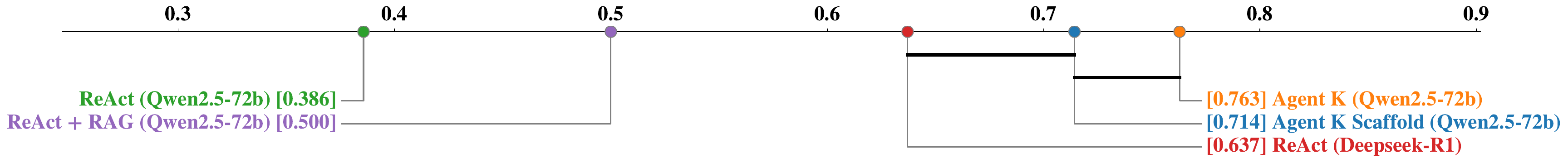}
\includegraphics[width=\linewidth,trim={1cm 0cm 1cm 3cm},clip]{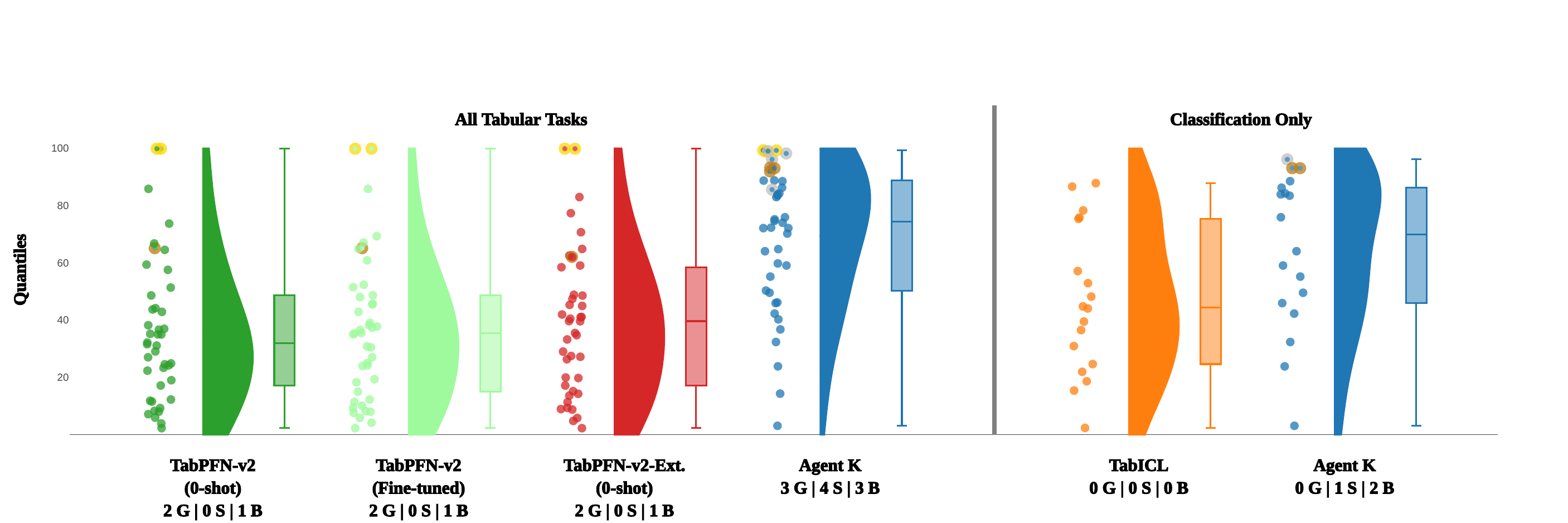}
\includegraphics[width=.9\linewidth,trim={0cm 0cm 0cm 0cm},clip]{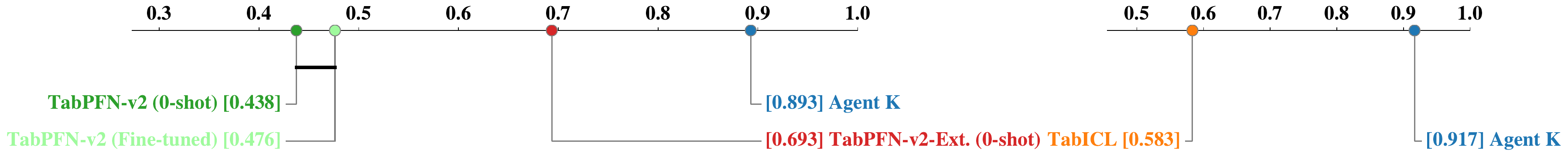}

\caption{\textbf{Performance Comparison of \OurAgent{} versus Competing Agents and Foundational Models.} (Top Row). 
We compare \OurAgent{} to three ReACT-style agents: ReACT (Qwen), ReACT (Qwen) with RAG, and ReACT (DeepSeek-R1). We also include \OurAgent{} (Scaffold Only), which is limited to scaffolded learning environments and does not support open-ended generalisation. 
We show for each method the distribution of the performance quantiles and the number of medals it achieves, as well as a critical difference diagram among each group of methods.
The full \OurAgent{} achieves the highest median performance (near the \AgentKEloquantile th percentile) and the strongest medal-equivalent record across over \nCompetitionsComparisonViolinReAct{} tasks: \AgentKTabGold\ gold (G), \AgentKTabSilver\ silver (S), and \AgentKTabBronze\ bronze (B).
(Bottom Row). We benchmark \OurAgent{} against a family of TabPFN-v2 models, including both zero-shot and fine-tuned variants. \OurAgent{} consistently outperforms all TabPFN-v2 baselines on real-world tabular tasks, where the strongest baseline achieves a ~\TabPFNquantile\% median and only \TabPFNGold\ gold, \TabPFNSilver\ silver, and \TabPFNBronze\ bronze medals. For classification tasks, we additionally compare against TabICL, a long-context in-context learning variant of TabPFN-v2, which performs notably worse than \OurAgent{}. These results demonstrate that \OurAgent{}’s structured experiential learning architecture enables broad generalisation and competitive performance across diverse data science domains.}
\label{Fig:Res1}
\end{figure}
Having established that \OurAgent{} achieves human-competitive performance across a diverse set of real-world challenges, we now evaluate how it compares to AI agents that do not rely on structured learning cycles like Kolb’s, but instead leverage the emergent reactive properties of LLM. 

In the top row of Figure \ref{Fig:Res1}, we compare \OurAgent{} against several variants: ReAct (equipped with Qwen), RAG-augmented ReAct (dubbed ReAct (Qwen) + RAG), and ReAct (equipped with DeepSeek-R1), as well as a variant of \OurAgent{} limited to scaffolded learning only. 
ReAct-based agents serve as meaningful baselines since they share our goal of leveraging LLMs for autonomous reasoning, but without a principled learning architecture grounded in cognitive theory. 
We use AIDE~\cite{chan2024mlebenchevaluatingmachinelearning, Jiang2024AIDE_ML_WECO} implementation for the ReAct-based agents, employing the same tree-based exploration strategy used by \OurAgent{}  in the post-scaffold phase. 
The  ReAct (Qwen) + RAG variant retrieves relevant Kaggle notebook or discussion elements to guide its initial solutions generation. 
Comparing to this baseline allows us to address the natural question of whether the solutions obtained through our scaffolded setting could simply be retrieved rather than discovered experientially. 
On the other hand, ReAct (DeepSeek-R1) provides a strong open-source foundation model baseline, allowing us to ask whether \OurAgent{}’s performance stems from its structured learning process or from the choice of the backbone language model. 

\OurAgent{} outperforms all competing agents, achieving the highest median performance across \nCompetitionsComparisonViolinReAct{} tasks. Its structured alternation between extrinsic functions (task execution and environment interaction) and intrinsic functions (reflection, abstraction, and strategy formation) enables better generalisation and learning efficiency. Importantly, \OurAgent{} surpasses reactive agents, suggesting that principled, cognitively inspired agent design, grounded in Kolb’s learning cycle and scaffolded within a zone of proximal development, can offer substantial advantages over purely reactive, emergent strategies.

\paragraph{\OurAgent{} vs. TabPFN-v2:} While \OurAgent{} performs strongly across a wide range of tasks, it is important to assess how it compares to domain-specialised models designed and optimised for specific data modalities. The bottom row of Figure \ref{Fig:Res1} presents this comparison.
We evaluate \OurAgent{} against multiple variants of TabPFN-v2, including zero-shot, fine-tuned, and long-context versions. 
Despite being a generalist system, \OurAgent{} outperforms all TabPFN-v2 baselines, achieving a median performance near the \AgentKTabquantile th percentile and earning \AgentKTabGold{} gold (G), \AgentKTabSilver{} silver (S), and \AgentKTabBronze{} bronze (B) medal-equivalent scores across the evaluation set. 
In contrast, the strongest variant among TabPFN-v2 baselines achieves a median of ~\TabPFNquantile\% and collects only \TabPFNGold\ gold, \TabPFNSilver\ silver, and \TabPFNBronze\ bronze medals. 
These results highlight that \OurAgent{}’s structured experiential learning not only generalises across domains but also competes with or surpasses task-specific state-of-the-art models in their domain, without any direct supervision or handcrafted features.

\begin{figure}[h!]
\centering
\begin{subfigure}[T]{0.48\textwidth}
        \centering
        \includegraphics[width=\linewidth, trim={0.2cm 0cm 1.5cm 2.5cm}, clip]{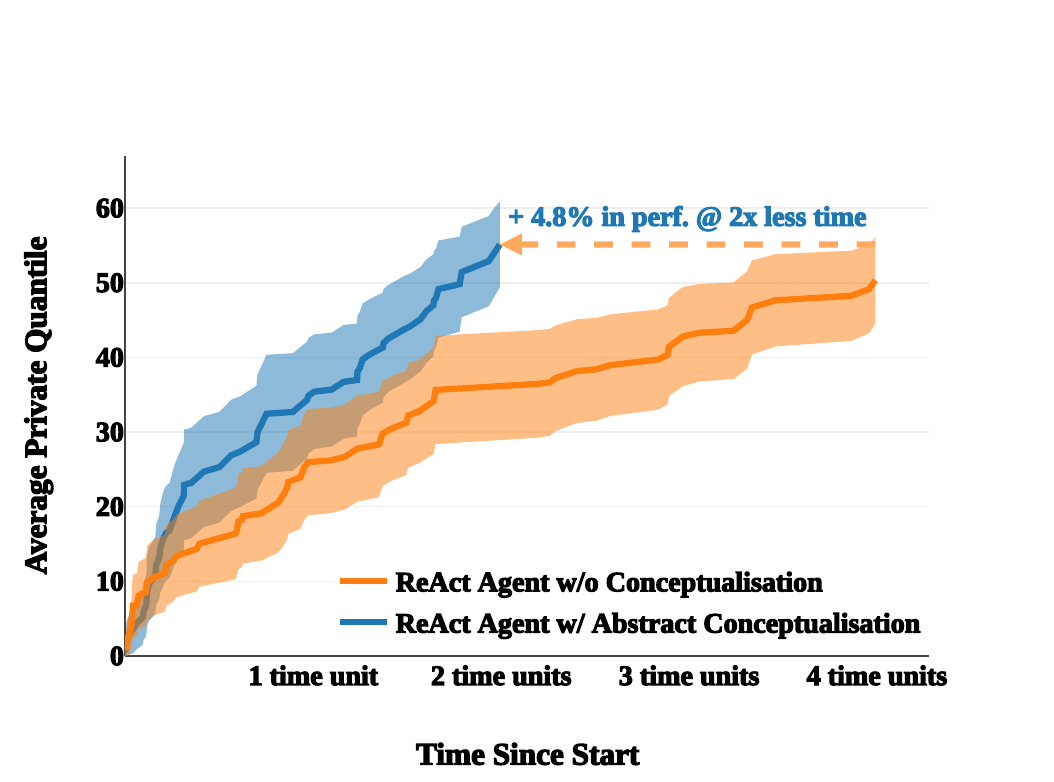}
        \caption{\textbf{Impact of abstract conceptualisation on performance.}}
        \label{fig:ablation-abstract-conceptualisation-bigpic}
    \end{subfigure}
    \hfill
\begin{subfigure}[T]{0.48\textwidth}
        \centering
       \includegraphics[width=\linewidth, trim={0cm 0cm 1.5cm 3.1cm}, clip]{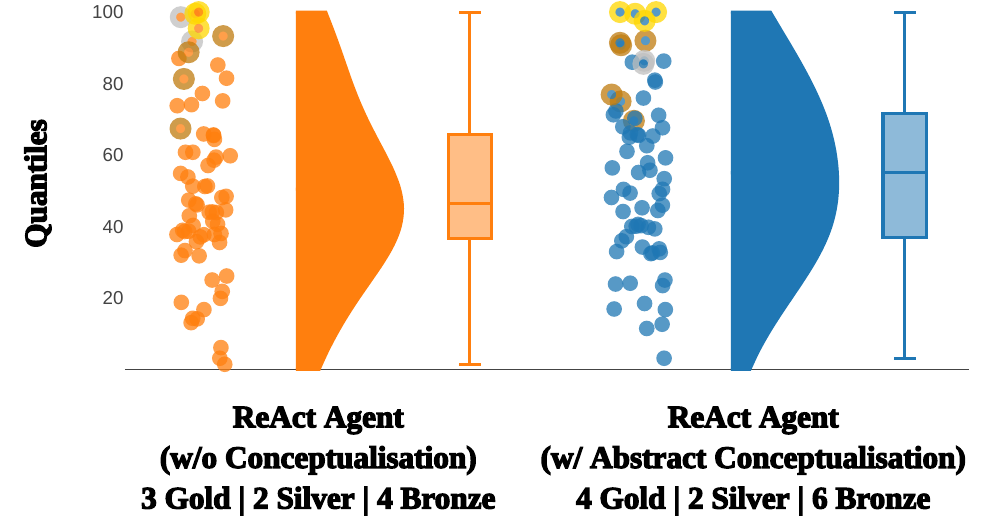}
       \includegraphics[width=\linewidth, trim={0cm 0cm 0cm 0cm}, clip]{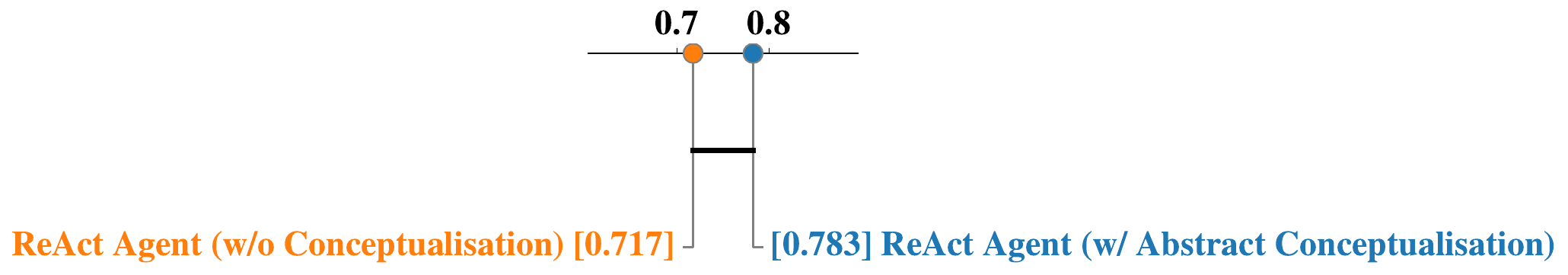}
        \caption{\textbf{Final Performance of ReAct Agents with and without abstract conceptualisation.}}
        \label{fig:ablation-abstract-conceptualisation-final-quantiles}
    \end{subfigure}
   \caption{\textbf{Ablation Study on the Impact of Abstract Conceptualisation.} A plot comparing average quantile performance in open-ended data science tasks of \OurAgent{} (with abstraction and summarisation from scaffold) versus shallow ReAct strategies. 
Since \OurAgent{} already spent half of the total runtime to  build its CoT in the scaffold, the ReAct-based agent without abstraction was provided with this total runtime  to enable a fair comparison. 
We can see that ZPD abstractions allow \OurAgent{} to achieve \ImprovementQuantileAgentKvReAct\% quantile improvements, while requiring 2x less exploration time. On the x-axis, one ``time unit'' corresponds to 12 hours for tabular competitions, and 24 hours for CV, NLP or multimodal competitions.}
\label{fig:LearningCurveAll}
\end{figure}

\paragraph{Abstract to Act, Not Just ReAct:} \OurAgent{} achieves strong results in medal count and human-level performance; we now examine how scaffolded abstraction and ZPD-driven summarisation support its success in open-ended tasks through a dedicated ablation study on the use of abstract conceptualisation. 
In these ablation experiments, agents tackle data science problems from scratch by growing a tree of solutions, where each node represents a complete code attempt, and subsequent steps involve debugging or improving past nodes. 
We compare \OurAgent{}’s strategy, which invests time in the scaffold to develop initial viable solutions and then abstracts and summarises them as CoTs for open-ended tasks, to the ReAct-based agent equipped with the same LLM as \OurAgent{} (Qwen), and that runs without the abstraction phase. 
To ensure a fair comparison, the ReAct Agent (Qwen) without abstraction was given additional exploration time to match \OurAgent{}’s total runtime, including its scaffold phase. 

In Figure~\ref{fig:ablation-abstract-conceptualisation-bigpic}, we plot the average quantile performance in open-ended data science tasks achieved by the ReAct-based agent with abstraction and summarisation from scaffold, versus the shallow ReAct strategies. 
We see that ZPD abstractions lead to \ImprovementQuantileAgentKvReAct\% quantile improvements, while requiring twice as little exploration time. 
Moreover, after 2 time units, we see that ReAct-based agent with abstract conceptualisation achieves an average of \ImprovementQuantileAgentKvReActAtTwoUnits{}\% higher quantiles than its shallow counterpart. 
Conducting a Welch t-test on the two groups of leaderboard quantiles obtained with and without abstraction, we get a p-value of $1.67e10^{-4}$ after 2 time units, meaning that average quantiles are statistically different. 
When doing another Welch t-test on the results achieved after 2 time units (resp. 4 time units) for the ReAct-based agent with (resp. without) abstract conceptualisation, the p-value is $0.26$, which does not allow to reject the hypothesis of equal means.
Moreover, we display in Figure~\ref{fig:ablation-abstract-conceptualisation-final-quantiles} the distribution of final performance quantiles and the number of medals achieved with and without starting from abstract conceptualisation. 
This shows that abstract conceptualisation notably enables the ReAct-based agent to obtain one extra gold and two extra bronze medals over the $\nCompetitionsComparisonViolinReAct{}$ competitions for which both agents made at least one valid submission.


\section*{Discussion}

This paper introduced \OurAgent{}, the first LLM-based agent to implement Kolb's theory and Vygotsky's principle to achieve high-performance results on a wide range of data science tasks, including tabular data, computer vision, natural language processing, and multimodal Kaggle challenges. 
Our agent operates fully autonomously, seamlessly handling everything from navigating a URL to building models, making submissions, and generating high-score submissions. 
While our results are successful and \OurAgent{} achieves new state-of-the-art performance of data science agents, we still want to highlight four potential limitations of our work.

\textbf{i) Use of recent technology compared to the competition release date.} Since for each Kaggle competition in our benchmark we compare against human participants who developed their solutions during the competition’s active period, we acknowledge that \OurAgent{} may benefit from using more recent technology that was not available to human competitors at the time.
For example, nothing prevents \OurAgent{} from generating a code using a Vision Transformer (ViT) architecture such as MaxViT  pretrained on ImageNet, which was released in September 2022, to solve a competition that ended months or years before.
To partially control for this factor, we report in Figure~\ref{fig:time_consistent_medal} (Methods) the number of medals that \OurAgent{} would have achieved across our competition pool after excluding submissions that relied on models whose public release dates postdate the competition deadline.
While this restriction reduces its performance, \OurAgent{}  still  secures a substantial number of medals, getting \nMedalsTimeConstrained{} out of the \nTotalMedals{} medals obtained without this constraint. 
We emphasize that this ``technology release date'' discrepancy only affects comparisons with historical human leaderboards and does not impact comparisons with baseline agents or foundation models, which are evaluated under the same technological conditions.

\textbf{ii) Risk of solution memorization.} While using newer technology can inflate performance relative to human competitors, a separate risk is that \OurAgent{} might simply reproduce solutions from top public kernels or repositories available online, which the underlying LLM would have seen during its training.
In such a scenario, high performance could result from memorization rather than genuine problem-solving ability.
We believe this risk is minimal at the \OurAgent{} scaffolding level: the agent’s solution-generation process involves significant adaptation, multi-step reasoning, and integration of diverse tools, making direct copying unlikely.
Moreover, we observe that \OurAgent{} still performs strongly in several competitions that ended after the cut-off date of the most recent LLM in our setup (e.g., end of 2023 for Qwen-2.5), where public solutions could not have influenced training. 
This is the case for competitions like ``playground-series-s4e4'' (ended in May 2024) or ``playground-series-s5e5'' (ended in June 2025)  on which \OurAgent{} outperformed more than 88\% of the participants.
This suggests that its performance is not merely attributable to regurgitating pre-existing solutions.

\textbf{iii) Competition difficulty heterogeneity.}
Kaggle competitions vary widely in difficulty.
Leaderboard-based metrics such as percentile rank or medal count are inherently influenced by the skill level and size of the participant pool.
For example, achieving a top 5\% rank in a high-profile competition that attracts many grandmasters is significantly harder than achieving the same rank in a less popular challenge. 
Furthermore, not all competitions offer medals, and these tend to attract different levels of participant commitment.
To quantify difficulty, we compute an Elo-based competition level: through all medal-awarding competitions in Kaggle’s history, we calculate the Elo-MMR ratings of all Kaggle users and we consider for each competition the average Elo of the participants who earned at least a Bronze medal.
This provides a relative measure of how competitive a given competition was.
In Figure~\ref{Fig:competition-difficulty} in the Appendix, we analyse \OurAgent{}’s medal distribution as a function of the competition level, enabling a fairer interpretation of its achievements across diverse competition types.

\textbf{iv) Absence of recent featured Kaggle competitions.}
While \OurAgent{} demonstrates strong performance on traditional Kaggle competitions, the platform’s scope has recently expanded to include more diverse and unconventional challenges. 
These range from mathematical puzzle solving~\footnote{\url{https://www.kaggle.com/competitions/santa-2024}} and agent design competitions~\footnote{\url{https://www.kaggle.com/competitions/konwinski-prize}} to ambitious tasks aimed at advancing Artificial General Intelligence (AGI), such as the ARC-AGI challenge~\footnote{\url{https://www.kaggle.com/competitions/arc-prize-2025}}.

Enabling \OurAgent{} to effectively participate in such cutting-edge competitions remains a significant challenge. It would require substantial enhancements to both the scaffolding framework and the  integration of advanced reasoning tools, domain-specific knowledge, and potentially multi-agent collaboration mechanisms.
Nonetheless, the milestones achieved by \OurAgent{} on more standard data science and machine learning tasks represent a major breakthrough. 
They demonstrate the power of embedding human-like learning paradigms into autonomous agentic systems, paving the way for future expansion into even more complex problem domains.

\printbibliography

\section*{Methods}
\begin{refsection}

\subsection*{Stage-Wise Scaffolded Learning with Feedback Control} 
We define scaffolded learning as an experiential interaction process in which an agent progresses through a sequence of staged environments: $\mathcal{T}=\{\mathcal{T}_1, \mathcal{T}_2,\dots, \mathcal{T}_n\}$, where each task $\mathcal{T}_i$ is designed to train a specific capability, and tasks reflect increasing complexity or dependency.
This structure is inspired by Vygotsky’s theory of the ZPD, where learners develop most effectively when placed in tasks slightly beyond their current ability, provided the environment offers necessary support. 
In our context, for example, LLM-based agents may struggle to generate correct, multi-step code solutions from scratch in open-ended settings, but can succeed when the task is scaffolded into modular stages with explicit validation. 
These stages are not hand-coded solutions, but reflect natural decompositions that a practitioner might follow, such as parsing data formats, defining metrics, or preparing submissions, each aligned with interpretable feedback and domain structure.

Each environment $\mathcal{T}_i$ is paired with a feedback function $F^{(i)}(\mathcal{T}_i,z_t)\rightarrow \mathcal{F}_t^{(i)}$ which evaluates the agent's action $z_t=\mathcal{E}(\Sigma_t^{\prime})$, returning a feedback signal $\mathcal{F}^{(i)}_t$, that determines whether the agent has progressed and completed the stage. 
The form of $\mathcal{F}^{(i)}_t$ can vary across the scaffold (e.g., passing a unit test or matching output structure), as we detail in the next section. 
Transitions between stages are gated by competency: the agent may only progress to $\mathcal{T}_{i+1}$ once $\mathcal{F}_{t}^{(i)}$ indicates that the agent has successfully demonstrated the necessary competency for the task at hand.
This gating mechanism ensures that scaffolded learning is not purely sequential but also adaptive, contingent on the agent's behaviour.

At each scaffolded stage, the agent receives a task description (typically in natural language), access to relevant tools (such as a Python interpreter), and feedback from the environment after attempting a solution. 
This feedback is then used to update the agent’s internal state via the update function, enabling it to revise its strategy. 

\subsection*{Implementing \OurAgent{}: From Scaffold to Autonomy}
\OurAgent{} is implemented as a modular agentic system that instantiates our experiential learning framework within a staged environment designed to reflect the structure and complexity of real-world data science tasks. This section details how the scaffolded environment supports structured capability acquisition, how intrinsic and extrinsic functions are operationalised through LLMs and code execution, and how the agent transitions to autonomous open-ended solution generation.
\subsubsection*{Scaffolded Learning Environment}
This section details the design and implementation of the workspace and solution-generation scaffolds that \OurAgent{} navigates.

\paragraph{Workspace Scaffold:} \OurAgent{}’s workspace scaffold is designed to support structured exploration of how to organise, interpret, and prepare raw task environments for downstream learning. Given only a competition URL, the agent autonomously builds an interpretable workspace by progressing through modular stages, each requiring reasoning, code generation, and validation via structured feedback.

The scaffold begins by downloading competition data using the Kaggle API and scraping the associated webpage for key elements: task descriptions, evaluation metrics, submission formats, and data specifications. These raw texts are summarised into focused prompts that compress relevant information and remove distractors (e.g., emojis, formatting artefacts). Examples of these prompts are provided in Appendix~\ref{Sec:ExtendedPrompts}.

The agent then detects the input and target modalities (e.g., tabular, image, text) and begins constructing the workspace through a series of code generation tasks. It creates mapping files that split the data by modality (e.g., input and output maps in Figure \ref{fig:env-scaffold}), as well as transformation scripts that convert targets into model-consumable formats and back. For instance, it may generate Python code to convert textual class labels into one-hot vectors and a corresponding inverse function to decode predictions into submission-ready labels.

Each scaffolded stage is paired with a task-agnostic unit test that verifies the correctness of the output, checking file existence, column structure, path validity, and basic data integrity. 
For instance, a test for an image input map might check that the file contains an ``id'' column, references to valid image paths, and has no empty or duplicated fields. 
These tests are not written or modified for individual tasks; they are general-purpose and apply across all competitions, ensuring consistency without manual intervention (see Section \ref{App:UnitTests} in the Appendix for examples). If a test fails, the agent revises its code and retries the stage. 
Meta-unit tests validate consistency across stages, for example, ensuring that all generated maps can be jointly loaded into a $\texttt{DataLoader}$ object for model training.

These scaffolded stages are defined by the environment, but the agent’s behaviour within them --  its code, strategies, and retries -- is not hand-coded. 
Instead, it dynamically constructs its solutions through reflection, planning, and feedback-driven code generation.
The process instantiates early experiential learning: the agent experiments, receives structured feedback, and adapts. 

\paragraph{Solution Scaffold:} After completing the workspace scaffold, \OurAgent{} enters the solution scaffold, where it autonomously constructs and refines predictive models using feedback from public leaderboard scores. The agent’s behaviour in this phase is conditioned on the modalities it identified. For tabular tasks, it leverages AutoML tools; for image, text, or multimodal tasks, it generates deep learning models using PyTorch. All model training, evaluation, and submission routines are, of course, automated and implemented via LLMs. 

In tabular competitions, \OurAgent{} solves the task by invoking an AutoML tool. In our implementation, this AutoML tool is RAMP~\cite{Kegl2018RAMP}, an in-house AutoML library, and writes the necessary components, such as the metric to use or the name of the target, to interface with it. However, the system is not restricted to RAMP; the agent can be extended to use any compatible AutoML system, provided it can reason about the required interfaces and generate appropriate invocation code. We also introduce the following novel tools: 

\begin{figure}[t!]
\centering
\includegraphics[width=1.0\linewidth,trim={0em, 0em, 0em, 3em}, clip]{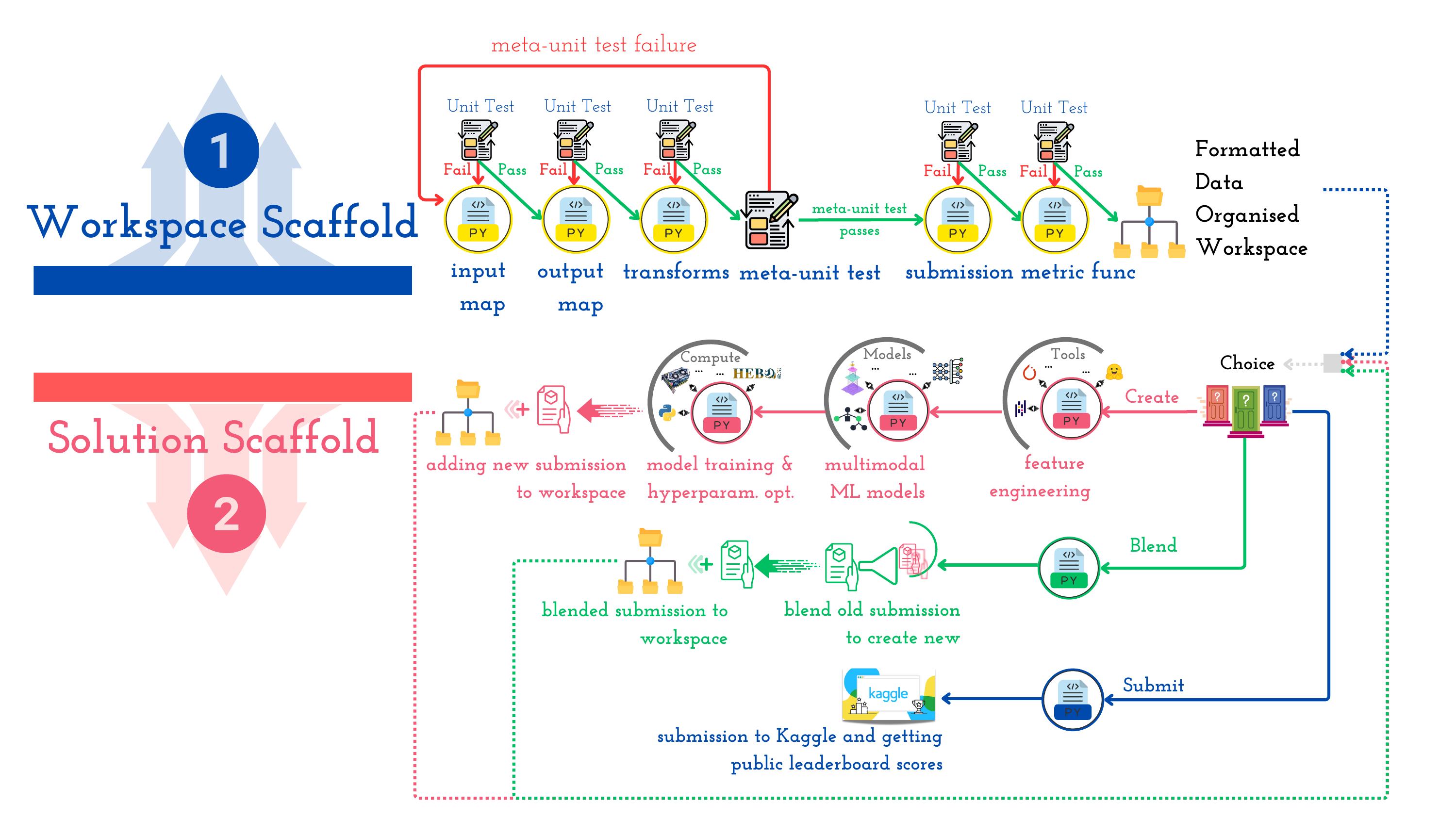}
\caption{\textbf{Two-stage scaffolded learning environment in \OurAgent{}.} In the Workspace Scaffold (top), the agent learns to build a structured workspace, mapping inputs and outputs, transforming data, and defining evaluation metrics, by passing a sequence of gated unit tests. At each stage, it must autonomously generate executable code that satisfies structural and functional constraints provided by the environment.
In the Solution Scaffold (bottom), the agent explores multiple solution strategies, including those that may access established tools for model training, feature engineering, and blending. These tools act as a learned foundation, akin to guidance from a teacher, allowing the agent to focus on reasoning. Feedback from validation scores drives iterative improvement, supporting experiential learning through reflection, abstraction, and adaptation.}
\label{fig:env-scaffold}
\end{figure}

\paragraph{LLM-Based Tool for Automated Feature Engineering:} Feature engineering is critical in enhancing machine learning performance, particularly in tabular data problems, by revealing informative patterns beyond the raw features. However, manual feature engineering is time-consuming, domain-specific, and challenging to scale. To address this, we developed an automated feature engineering tool within the environment, enabling \OurAgent{} to dynamically propose and implement feature transformations. 
The tool leverages an LLM conditioned on the problem description, feature distribution statistics, and small random data samples. Given these inputs, the LLM generates Python code to create new features, which \OurAgent{} applies to augment the dataset. 

\paragraph{Automated Blending Tools:} Model blending is a widely used ensembling technique in machine learning competitions, improving performance by aggregating the complementary strengths of diverse models \cite{tang2024autogluonMultimodal,WOLPERT1992241,bellkor2009bigchaos}.
To leverage this, we developed a dedicated blending tool within the environment.
After training multiple models, \OurAgent{} can select a subset and invoke the tool to construct a final submission based on a weighted combination of their outputs. 
To select models for blending, we leverage an LLM that reviews each candidate's architecture, loss functions, and validation performance. 
Based on its training over large code and language corpora, the LLM proposes a combination of subsets of models. 
Their predictions, such as logits in classification tasks, are aggregated and fed to a small multi-layer perceptron (MLP) trained to produce final outputs, yielding a new submission written to the workspace.

\paragraph{Class Balancing and Target Scaling Tools:} Addressing class imbalance and target scaling is critical for building effective classification and regression pipelines. 
\OurAgent{} can adjust for imbalanced class distributions in classification tasks by dynamically designing resampling strategies that rebalance the training set based on observed label frequencies and evaluation metrics.
In regression tasks, it normalises target distributions before training to improve convergence and accuracy, and reverses the transformations at inference time.
These strategies are implemented through LLM-generated code and iteratively refined based on prior solutions and task-specific feedback. 

\subsubsection*{Experiential Learning in \OurAgent{}}
To implement experiential learning in \OurAgent{}, we use different prompt strategies to enable reflection, abstraction, and integration of environment feedback, during the scaffold phase and during the open-ended generation phase. 

\paragraph{During Scaffolded Learning} 

 The implementation of intrinsic functions relies on the LLM-prompting, where the prompts integrate elements from the agent's internal state $\Sigma_t$ that contain characteristics of the competition to solve (e.g. the original task and data description from Kaggle)  and elements obtained during previous steps (e.g. previously generated codes, summaries, or queried table views).
 The details of the prompts for the intrinsic thinking and external action generation, as well as the internal state update rule, depend on each scaffold stage.
 We provide in Table~\ref{tab:agentInstrinsicExtrinsic} details on the various prompting schemes that make \OurAgent{} progress through the different stages, and we report the exact prompts in Appendix~\ref{Sec:ExtendedPrompts}.

\begin{figure}[h!]
\centering
\begin{tikzpicture}
 \node[] (pic1) {
  \begin{tikzpicture}

\definecolor{myforest}{RGB}{34,139,34} 

\renewcommand{\thefootnote}{\fnsymbol{footnote}}
\newcommand{\tasksummary}{\textcolor{blue}{\texttt{\textlangle task\_summary\textrangle}}}
\newcommand{\dataview}{\textcolor{blue}{\texttt{\textlangle data\_view\textrangle}}}
\newcommand{\datadescr}{\textcolor{blue}{\texttt{\textlangle data\_descr\textrangle}}}
\newcommand{\errorAnalysis}{\textcolor{green!80!black}{\texttt{\textlangle error\_analysis\textrangle}}}

\newcommand{\angledsub}[2]{\langle \texttt{#1}_{#2} \rangle}
\newcommand{\planAngledsub}[1]{\textcolor{myforest!80!black}{\angledsub{plan}{#1}}}
\newcommand{\codeAngledsub}[1]{\textcolor{purple!80!black}{\angledsub{code}{#1}}}
\newcommand{\errorAngledsub}[1]{\textcolor{red!80!black}{\angledsub{error\_trace}{#1}}}
\newcommand{\errorSolveAngledsub}[1]{\textcolor{cyan!80!black}{\angledsub{error\_analysis}{#1}}}

\makeatletter
\@ifundefined{boxWidthFive}{
  \newlength{\boxWidthFive}
}{}
\@ifundefined{colOneWidthFive}{
  \newlength{\colOneWidthFive}
}{}
\@ifundefined{vertDistFormula}{
  \newlength{\vertDistFormula}
}{}
\@ifundefined{headerHeight}{
  \newlength{\headerHeight}
}{}
\@ifundefined{promptToAnswerGap}{
  \newlength{\promptToAnswerGap}
}{}
\@ifundefined{sideMargin}{
  \newlength{\sideMargin}
}{}
\@ifundefined{boxHorGap}{
  \newlength{\boxHorGap}
}{}
\@ifundefined{reflectboxWidthFive}{
  \newlength{\reflectboxWidthFive}
}{}
\makeatother

\setlength{\colOneWidthFive}{3.5cm}
\setlength{\vertDistFormula}{.1cm}
\setlength{\headerHeight}{1.em}
\setlength{\promptToAnswerGap}{.5em}
\setlength{\sideMargin}{.1cm}
\setlength{\boxHorGap}{.8cm}
\setlength{\boxWidthFive}{\dimexpr (18cm - 2\boxHorGap - 2\sideMargin - \colOneWidthFive) * 100 / 200 \relax}
\setlength{\reflectboxWidthFive}{\boxWidthFive}

\tikzset{
  prompt-node/.style={
    draw=none,
    fill=blue!10,
    rounded corners=3pt,
    text width=\boxWidthFive,
    inner sep=3pt,
    font=\small,
    align=left
  },
  prompt-header/.style={
    font=\bfseries\footnotesize,
    text=white,
    fill=blue!60,
    rounded corners=0pt,
    minimum height=\headerHeight,
    text width=\boxWidthFive,
    inner xsep=3pt,
    anchor=north west,
    align=left,
  },
  answer-node/.style={
    draw=none,
    fill=orange!15,
    rounded corners=3pt,
    text width=\boxWidthFive,
    inner xsep=3pt,
    font=\small,
    align=left
  },
  answer-header/.style={
    font=\bfseries\footnotesize,
    text=white,
    fill=orange!60,
    rounded corners=0pt,
    minimum height=\headerHeight,
    text width=\boxWidthFive,
    inner xsep=3pt,
    anchor=north east,
    align=right
  },
  console-node/.style={
    draw=none,
    fill=black,
    text=green!30,
    rounded corners=3pt,
    text width=\boxWidthFive,
    inner sep=3pt,
    font=\ttfamily\small,
    align=left
  },
  console-header/.style={
    font=\bfseries\footnotesize,
    text=white,
    fill=black!60,
    rounded corners=0pt,
    minimum height=\headerHeight,
    text width=\boxWidthFive,
    inner xsep=3pt,
    anchor=north east,
    align=left,
  },
  update-node/.style={
    draw=none,
    fill=myforest!10,
    text=black,
    rounded corners=3pt,
    text width=\boxWidthFive,
    inner sep=3pt,
    align=left
  },
  update-header/.style={
    font=\bfseries\footnotesize,
    text=white,
    fill=myforest!60,
    rounded corners=0pt,
    minimum height=\headerHeight,
    text width=\boxWidthFive,
    inner xsep=3pt,
    anchor=north east,
    align=left,
  },
}
  \coordinate (One) at (-.7\boxWidthFive,.5cm);
  \coordinate (Two) at (0, 1cm);
  \coordinate (Three) at (.7\boxWidthFive, .5cm);
  \coordinate (Four) at ($ (0,0) - (One) $);
  \coordinate (Five) at ($ (0,0) - (Two) $);
  \coordinate (Six) at ($ (0,0) - (Three) $);

  \draw[->, thick, black, line width=3pt]
    plot[smooth, tension=.8] coordinates {
      (One)
      (Two)
      (Three)
      (Four)
      (Five)
      (Six)
    };

\node[align=center, anchor=center] at (0, 0) { \textbf{Experiential Learning} \\ \textbf{Loop}};

\node[align=center, anchor=south east, text width=\colOneWidthFive, inner sep=3pt] (sigma_t) at ($(One)+(-.5cm, .5cm)$) {
  \tasksummary, \datadescr, \dataview, $\dots$,
 \\ $\planAngledsub{1}$, $\codeAngledsub{1}$, $\errorAngledsub{1}$, $\errorSolveAngledsub{1}$\\
  $\dots$,\\ $\planAngledsub{k}$, $\codeAngledsub{k}$, $\errorAngledsub{k}$
};

\draw [decorate,decoration={brace,amplitude=5pt},xshift=-5pt]
  ([yshift=-5pt]sigma_t.south west) -- ([yshift=5pt]sigma_t.north west)
  node [midway,xshift=-6pt] {};

\draw [decorate,decoration={brace,amplitude=5pt,mirror},xshift=5pt]
  ([yshift=-5pt]sigma_t.south east) -- ([yshift=5pt]sigma_t.north east)
  node [midway,xshift=6pt] {};

\node[above=.2cm of sigma_t, align=center] (sigma_t_name) {\Large $\Sigma_t$ \\ Internal State};

\node[anchor=south , above=\vertDistFormula of Two, align=center, draw=none, text width=1.5\boxWidthFive] (sigma_t_1_name) {$\mathcal{I}_t^1(\Sigma_t)= \Sigma_t \cup \{ \errorSolveAngledsub{k} \}$};

\node[above=\promptToAnswerGap of sigma_t_1_name, answer-node, anchor=south] (I_r_answer) {\vspace{\headerHeight} \\  ``Given error message and ... \\ $\errorSolveAngledsub{k}$ ''};
\node[answer-header] at ([xshift=0cm,yshift=0cm]I_r_answer.north east) {LLM Answer};

\node[prompt-node, above=\vertDistFormula of I_r_answer, anchor=south] (I_r_prompt) {
    \vspace{\headerHeight} \\
    You are a Kaggle GM... \\ 
    Task info: \tasksummary, \datadescr, \dataview,...\\
    This was your code: $\codeAngledsub{k}$\\ 
    Explain how to fix: $\errorAngledsub{k}$\\
    Also consider previous guidance: $\left\{\errorSolveAngledsub{i}\right\}_{i=1}^{k-1}$
};
\node[prompt-header] at ([xshift=0cm,yshift=0cm]I_r_prompt.north west) {Prompt (e.g. Error Analysis)};

\node[above=.2cm of I_r_prompt, align=center, text width=\reflectboxWidthFive, anchor=south] (I_r_name) {\Large $\mathcal{I}_t^1(\Sigma_t)$ \\ \normalsize Analyse};

\node[align=center, draw=none, text width=\boxWidthFive, anchor=south west] (sigma_prime_name)  at ($(Three)+(.0cm, .3cm)$){
    $\Sigma^\prime_t = \mathcal{I}_t^1(\Sigma_t) \cup \{\planAngledsub{k + 1}\}$
};

\node[above=\vertDistFormula of sigma_prime_name, answer-node, anchor=south] (I_a_answer) {\vspace{\headerHeight} \\  ``To solve this, I should... \\ $\planAngledsub{k + 1}$ ''};
\node[answer-header] at ([xshift=0cm,yshift=0cm]I_a_answer.north east) {LLM Answer};

\node[above=\vertDistFormula of I_a_answer, prompt-node, anchor=south] (I_a_prompt) {
    \vspace{\headerHeight} \\
    You are a Kaggle GM... \\ 
    Task info: \tasksummary, \datadescr, \dataview,...\\
    This was your code: $\codeAngledsub{k}$ and you got $\errorAngledsub{k}$\\
    Design a plan taking into account: $\left\{\errorSolveAngledsub{i}\right\}_{i=1}^{k}$
};
\node[prompt-header] at ([xshift=0cm,yshift=0cm]I_a_prompt.north west) {Prompt (e.g. New Planning)};

\node[above=\vertDistFormula of I_a_prompt, align=center, text width=\reflectboxWidthFive, anchor=south] (I_a_name) {\Large $\mathcal{I}_t^2(\mathcal{I}_t^1(\Sigma_t))$ \\ \normalsize Plan
};

\node[align=center, text width=\reflectboxWidthFive, anchor=north west] (external_act_name)   at ($(Three)+(.0cm, -.0cm)$){\Large $\mathcal{E}_t(\Sigma_t^{\prime})$ \\\normalsize External action};

\node[below=.2cm of external_act_name, prompt-node] (external_act_prompt) {
    \vspace{\headerHeight} \\
    You are a Kaggle GM... \\ 
    Task info: \tasksummary, \datadescr, \dataview,...\\
    This was your code: $\codeAngledsub{k}$ and you got $\errorAngledsub{k}$\\
    Taking into account $\left\{\errorSolveAngledsub{i}\right\}_{i=1}^{k}$,\\
    implement a solution following $\planAngledsub{k+1}$...
};
\node[prompt-header] at ([xshift=0cm,yshift=0cm]external_act_prompt.north west) {Prompt (e.g. Implement)};

\node[below=\promptToAnswerGap of external_act_prompt, answer-node] (external_act_answer) {\vspace{\headerHeight} \\  \texttt{\`{}\`{}\`{}python}\\ $\codeAngledsub{k+1}$ \\ \`{}\`{}\`{}};
\node[answer-header] at ([xshift=0cm,yshift=0cm]external_act_answer.north east) {LLM Answer};

\node[below=\promptToAnswerGap of external_act_answer, align=center, draw=none, text width=\reflectboxWidthFive] (z_t_name) {$z_t := \mathcal{E}_t(\Sigma_t^\prime) = \codeAngledsub{k+1}$};

\node[anchor=north , below=\vertDistFormula of Five, align=center] (env_feedback_name) {\Large $F_t(\mathcal{T}_t, z_t)$ \\Get env. feedback};

\node[console-node, below=\vertDistFormula of env_feedback_name] (console){\vspace{\headerHeight}\\
\textcolor{green!60}{user@machine:\textasciitilde\$} python script.py \\
Traceback (most recent call last): [...] \\
\textcolor{red}{ValueError: invalid literal for int() with base 10: 'NaN'}
};
\node[console-header] at ([xshift=0cm,yshift=0cm]console.north east) {Env. (e.g. Run $z_t$ in a Terminal)
};

\node[below=\vertDistFormula of console, align=center, draw=none] (f_t_formula) {$\mathcal{F}_t =\{\codeAngledsub{k+1}, \errorAngledsub{k+1} \}$};

\node[align=center, anchor=north east, align=center] (update_func) at ($(Six)+(-.5cm, .5cm)$) {\Large $\mathcal{U}_t(\Sigma_t^\prime, \mathcal{F}_t)$ \\ \normalsize Update the state};

\node[update-node, align=center, below=\vertDistFormula of update_func] (update_rule){\vspace{\headerHeight}\\$\Sigma_t^\prime \cup \{\codeAngledsub{k+1}, \errorAngledsub{k+1} \}$
};
\node[update-header] at ([xshift=0cm,yshift=0cm]update_rule.north east) {Update (e.g. Store Elements)
};

\node[below=\vertDistFormula of update_rule, align=center, draw=none] (sigma_t_plus_one) {$\Sigma_{t+1}:= \mathcal{U}_t(\Sigma^\prime_t , \mathcal{F}_t)$};

\end{tikzpicture}
};
\end{tikzpicture}
\caption{\textbf{Experiential Learning in \OurAgent{} Scaffold - Error Handling.} 
Given internal state $\Sigma_t$ containing competition summarised information and past implementation attempts for a given stage $\mathcal{T}_t$ along with error logs, the experiential learning loop for error handling starts with ($\mathcal{I}_{t}^1$) the generation of analysis of the error trace obtained at the previous step.
It is followed by another intrinsic step ($\mathcal{I}_{t}^2$) to generate a new plan to pass this stage, taking into account the past error analyses. The external action function ($\mathcal{E}_{t}$) prompt the Agent's LLM to generate a piece of code that follows the plan stored in the provisory internal state $\Sigma^\prime$. 
The code is then tested in the environment, and the console output is added to the internal state to enable further debugging (if there is again an error). If the unit test passes, the environment switch to the next pipeline stage and feedback new stage-based information to the agent.
}
\label{fig:exp-learning-in-scaffold}
\end{figure}

 \begin{table}[h]
    \centering
    \resizebox{\textwidth}{!}{
    \begin{tabular}{ccccl}\toprule
         Stage&  Intrinsic Functions $\mathcal{I}$ &Extrinsic Functions $\mathcal{E}$ &Environment Feedback $F$ &Internal State Update $\mathcal{U}$\\ \midrule
         Competition understanding &  Summarise & Output summaries& Keep the summaries & Add Summary\\
         \makecell{Modality identification \\ (Tab., Image, Text)}&  Think \& Summarise & JSON with modalities& Store modalities & Add modality \\
         Create Input / Target maps &  Plan & Code & Run Unit Test & CSV map files \\
         Select Metric &  Plan & Code & Run Unit Test& Python code \\
         Create submission formats &  Plan & Code & Run Unit Test& Python code \\
         \midrule
         Feature Engineering &  Plan & Code & Run Unit Test& Python code\\ 
         Create Embedders &  Plan & Code & Run Unit Test& Python code\\ 
         Class Imbalance &  Identify \& Plan & Code & Run Unit Test& Python code\\ \
         Create Target Transform &  Identify \& Plan & Code & Run Unit Test& Python code\\ 
         Create Model Head &  Plan & Code & \makecell{Run Unit Test \\ \& Trigger training} & Add validation scores\\ 
         Create Solution summary &  Summarise & Output the summary & Keep summary & Summary \\ 
         Ensemble &  Think & Select solutions to ensemble & Run Blending & Add validation scores \\ 
         \midrule
         Error Code &  Think & Summarise the error & 
         Keep summary & Add error summary\\ \bottomrule
    \end{tabular}
    }
    \caption{Overview of Intrinsic and Extrinsic Functions, Environment Role, and Internal State updates. The bottom rows correspond to the solution generation scaffold, which applies for all types of competitions. The group of seven rows below correspond to the solution design scaffold, which applies to CV, NLP, and Multimodal problems, as for tabular-only competitions, the agent uses RAMP to create the solutions. The ``Error Code'' stage corresponds to the retrial of a stage requiring a code output and that failed during a previous attempt. For these stages, we indicate in the column  ``Internal State Update'' what the update is when the unit test passes, as when an error happens, the code and error logs are added to the internal state.}
    \label{tab:agentInstrinsicExtrinsic}
\end{table}
 
 We illustrate in Figure~\ref{fig:exp-learning-in-scaffold} a typical loop of  experiential learning loop when \OurAgent{} tries to generate a code to pass some setup stage $\mathcal{T}_t$. 
 In the example, we assume that the agent already generated some code at the previous steps, which led to some error, whose traces were stored in the internal state.
 We prompt the agent to analyse the encountered error and to design a new plan that serves to generate an external action consisting of  a piece of code which should solve the current stage. The code is parsed and executed within a unit test in the environment, and the console output is stored to be appended to the internal state, which acts as the update function.

\paragraph{During Open-Ended Solution Generation}

\begin{figure}[h!]
\centering
\begin{tikzpicture}[scale=0.8, every node/.style={scale=0.8}]
    \definecolor{myforest2}{RGB}{34,139,34}
    \newcommand{\angledsubTer}[2]{\langle \texttt{#1}_{#2} \rangle}
    \newcommand{\planAngledsubTer}[1]{\textcolor{myforest2!80!black}{\angledsubTer{plan}{#1}}}
    \newcommand{\codeAngledsubTer}[1]{\textcolor{cyan!80!black}{\angledsubTer{code}{#1}}}
    \newcommand{\logAngledsubTer}[1]{\textcolor{red!80!black}{\angledsubTer{run\_trace}{#1}}}
    \newcommand{\scaffoldCoTTer}[1]{\textcolor{magenta!80!black}{\angledsubTer{scaffold\_cot}{#1}}}
    \newcommand{\solPoolTer}{\textcolor{black}{\texttt{\textlangle submit\_pool}}}
    
    \renewcommand{\thefootnote}{\fnsymbol{footnote}}
    \newcommand{\tasksummaryTer}{\textcolor{blue}{\texttt{\textlangle task\_summary\textrangle}}}
    \newcommand{\dataviewTer}{\textcolor{blue}{\texttt{\textlangle data\_view\textrangle}}}
    \newcommand{\datadescrTer}{\textcolor{blue}{\texttt{\textlangle data\_descr\textrangle}}}
    \newcommand{\errorAnalysisTer}{\textcolor{blue}{\texttt{\textlangle error\_analysis\textrangle}}}
    \newcommand{\computerResourceTer}{\textcolor{blue}{\texttt{\textlangle compute\_resources\textrangle}}}
    \newcommand{\remainingTimeTer}{\textcolor{blue}{\texttt{\textlangle remaining\_time\textrangle}}}

    \newcommand{\angledsub}[2]{\langle \texttt{#1}_{#2} \rangle}
    \newcommand{\planAngledsub}[1]{\textcolor{myforest2!80!black}{\angledsub{plan}{#1}}}
    \newcommand{\codeAngledsub}[1]{\textcolor{cyan!80!black}{\angledsub{code}{#1}}}
    \newcommand{\logAngledsub}[1]{\textcolor{red!80!black}{\angledsub{run\_log}{#1}}}
    \newcommand{\scaffoldCoT}[1]{\textcolor{magenta!80!black}{\angledsub{scaffold\_cot}{#1}}}
    
    \def\specialInsightHeaderTextA{Scaffold CoTs}
    \def\specialInsightHeaderTextB{Prev. Best Node}
    \def\specialInsightHeaderTextC{Prev. Error}
    
    \def\specialInsightBodyTextA{- Tried \texttt{\textlangle solution A\textrangle}, got validation score \texttt{\textlangle val\_A \textrangle}\par - Tried \texttt{\textlangle solution B\textrangle}, got valid. score \texttt{\textlangle val\_B \textrangle}\par - ...}
    \def\specialInsightBodyTextB{Here is the code of the best attempt so far: \texttt{\textlangle code\textrangle}}
    \def\specialInsightBodyTextC{Previous (buggy) implementation: \texttt{\textlangle code\textrangle}\par
    Execution output: \texttt{\textlangle error log\textrangle}}
    
    \def\actionTypeTextA{Draft New Node}
    \def\actionTypeTextB{Improve Best Node}
    \def\actionTypeTextC{Debug Node}
    
    \def\specialInstructBodyTextA{In the light of the above insights, implement an initial solution...} 
    \def\specialInstructBodyTextB{Given the above code, implement a new solution that could improve the score...}
    \def\specialInstructBodyTextC{Given the error message, find}

    \def\llmResponseTitleTextA{Draft Node}
    \def\llmResponseTitleTextB{Improve Node}
    \def\llmResponseTitleTextC{Debug Node}

    \def\llmResponseBodyTextA{A first attempt to tackle this would be ... } 
    \def\llmResponseBodyTextB{To further improve the model, I will ... }
    \def\llmResponseBodyTextC{To fix the issue, we need to  ... }

    \def\pythonCode{ \vspace{\headerHeightBis} \\  
    ``$\planAngledsub{}$\\ \texttt{\`{}\`{}\`{}python}\\ 
    $\codeAngledsub{}$ \\ \`{}\`{}\`{}''\\  }
    
    \newlength{\solGenWidth}
    \newlength{\solGenHeight}
    \newlength{\headerheight}
    
    \newlength{\innerymargin}
    \newlength{\innerwidth}
    
    \newlength{\sharedMetaDataHeight}

    \newlength{\smallboxwidth}
    \newlength{\smallboxheight}
    \newlength{\smallboxgap}
    
    \newlength{\commonInstrBoxWidth}
    \newlength{\commonInstrBoxHeight}
    \newlength{\commonInstrBoxStartY}
       
    \setlength{\headerheight}{0.5cm}
    
    \setlength{\solGenWidth}{16cm}
    
    \setlength{\smallboxwidth}{\dimexpr \solGenWidth * 90 / 300 \relax}
    \setlength{\smallboxheight}{3cm}
    \setlength{\smallboxgap}{\dimexpr (\solGenWidth - 3 \smallboxwidth) / 6 \relax}

    \setlength{\innerymargin}{0.2cm}
    \setlength{\innerwidth}{\dimexpr \solGenWidth - 2 \smallboxgap \relax}
    
    \setlength{\sharedMetaDataHeight}{3.cm}

    \setlength{\commonInstrBoxWidth}{\innerwidth}
    \setlength{\commonInstrBoxHeight}{2.5cm}
    
    \setlength{\solGenHeight}{\dimexpr \headerheight + \innerymargin + \sharedMetaDataHeight  + \innerymargin + \smallboxheight + \innerymargin + \commonInstrBoxHeight + \innerymargin \relax}

    \tikzset{
  prompt-node/.style={
    draw=none,
    fill=blue!10,
    rounded corners=3pt,
    text width=\boxWidthTwo,
    inner sep=3pt,
    font=\small,
    align=left
  },
  prompt-header/.style={
    font=\bfseries\footnotesize,
    text=white,
    fill=blue!60,
    rounded corners=0pt,
    minimum height=\headerHeightBis,
    text width=\boxWidthTwo,
    inner xsep=3pt,
    anchor=north west,
    align=left,
  },
  answer-node/.style={
    draw=none,
    fill=orange!15,
    rounded corners=3pt,
    text width=\boxWidthTwo,
    inner xsep=3pt,
    font=\small,
    align=left
  },
  answer-header/.style={
    font=\bfseries\footnotesize,
    text=white,
    fill=orange!60,
    rounded corners=0pt,
    minimum height=\headerHeightBis,
    text width=\boxWidthTwo,
    inner xsep=3pt,
    anchor=north east,
    align=right
  },
  console-node/.style={
    draw=none,
    fill=black,
    text=green!30,
    rounded corners=3pt,
    text width=1.14\colOneWidthTwo,
    inner sep=3pt,
    font=\ttfamily\small,
    align=left
  },
  console-header/.style={
    font=\bfseries\footnotesize,
    text=white,
    fill=black!60,
    rounded corners=0pt,
    minimum height=\headerHeightBis,
    text width=1.14\colOneWidthTwo,
    inner xsep=3pt,
    anchor=north east,
    align=left,
  },
  update-node/.style={
    draw=none,
    anchor=north west,
    fill=myforest2!10,
    text=black,
    rounded corners=3pt,
    text width=\solGenWidth,
    inner sep=3pt,
    align=center
  },
  update-header/.style={
    font=\bfseries\footnotesize,
    text=white,
    fill=myforest2!60,
    rounded corners=0pt,
    minimum height=\headerHeightBis,
    text width=.99\boxWidthTwo,
    inner xsep=3pt,
    anchor=north east,
    align=left,
  },
}

    \newlength{\xpos}  
    \newlength{\ypos} 
    
    \setlength{\ypos}{0cm}
    \setlength{\xpos}{0cm}
    
    \node[draw=none] at(-.6, 1) (sigma_t_name) {$\Sigma_t$};
    \node[draw=none, text width=\solGenWidth, anchor=west, align=center] at (0, 1) (sigma_t) {
    $\left\{
\begin{aligned}
    &\tasksummaryTer, \datadescrTer, \computerResourceTer, \remainingTimeTer, \dots, \\
    &\left\{ \scaffoldCoTTer{i} \right\}_{i=1}^k, \solPoolTer \texttt{\textrangle}
\end{aligned}
\right\}$
        };

    \draw[draw=none, fill=gray!10] (0,\ypos) rectangle (\solGenWidth, -0.96\solGenHeight)  node[pos=0.5] (sol_gen_box) {};;
    \draw[draw=none, fill=blue!60] (0, - \headerheight) rectangle (\solGenWidth, \ypos);
    \node[draw=none, anchor=center, font=\bfseries\color{white}] at (.5\solGenWidth, - 0.5*\headerheight) {Solution Generation Prompt};

    \setlength{\ypos}{\dimexpr \ypos - \headerheight\relax}
     \foreach \i/\label in {0/A, 1/B, 2/C} {    
        \setlength{\xpos}{\dimexpr \smallboxgap +\i\smallboxwidth + \numexpr 2 * \i\relax  \smallboxgap \relax}
        \coordinate (smallOrigin) at (\xpos, \ypos);
        \node[anchor=center, font=\bfseries\color{black}, scale=1, minimum height=\headerheight] at ($(smallOrigin) + (0.5*\smallboxwidth, \ypos)$)  {\csname actionTypeText\label\endcsname};
        \ifnum \i<2
            \draw[dashed] (\xpos + \smallboxwidth + \smallboxgap, \ypos - \headerheight) -- ++(0, -\solGenHeight + 2\innerymargin + 2\headerheight);
        \fi
    }

    \setlength{\ypos}{\dimexpr \ypos - \headerheight - 2\innerymargin\relax}
    
    \coordinate (innerOrigin) at (\smallboxgap, \ypos - \headerheight);
    \draw[fill=teal!60, draw=none]
        (innerOrigin) ++(0, 0) rectangle ++(\innerwidth, \headerheight);
    \node[anchor=center, font=\bfseries\color{white}]
        at ($(innerOrigin) + (0.5\innerwidth, 0.5*\headerheight)$) {Shared Instructions};

    \setlength{\ypos}{\dimexpr \ypos - \headerheight - \innerymargin\relax}
    
    \coordinate (innerOrigin) at (\smallboxgap, \ypos - \sharedMetaDataHeight);
    \draw[fill=white, draw=none] (innerOrigin) rectangle ++(\innerwidth, \sharedMetaDataHeight);
    \node[anchor=north west, text width=.95\innerwidth, align=justify]
        at ($(innerOrigin) + (0, \sharedMetaDataHeight)$) {
            You are a Kaggle grandmaster attending a competition... Given a task you need to create a plan and implement a solution in Python...
            \par- \tasksummaryTer, \datadescrTer, \dots
            \par- \computerResourceTer  \& \remainingTimeTer 
            \par- \dots
        };
        
    \setlength{\ypos}{\dimexpr \ypos - 2\innerymargin - \sharedMetaDataHeight \relax}
    
    \coordinate (innerOrigin) at (\smallboxgap, \ypos);
    \draw[fill=teal!60, draw=none]
        (innerOrigin) rectangle ++(\innerwidth, -\headerheight);
    \node[anchor=center, font=\bfseries\color{white}]
        at ($(innerOrigin) + (0.5\innerwidth, -0.5\headerheight)$) {Reflection \& Abstraction elements};

    \setlength{\ypos}{\dimexpr \ypos - \headerheight - \innerymargin \relax}

    \foreach \i/\label in {0/A, 1/B, 2/C} {
        \setlength{\xpos}{\dimexpr \smallboxgap +\i\smallboxwidth + \numexpr 2 * \i\relax  \smallboxgap \relax}
        \coordinate (smallOrigin) at (\xpos, \ypos - \smallboxheight);

        \draw[fill=white, draw=none] (smallOrigin) rectangle ++(\smallboxwidth, \smallboxheight);
        \draw[fill=teal!70, draw=none]
            (smallOrigin) ++(0, \smallboxheight - \headerheight) rectangle ++(\smallboxwidth,\headerheight);

        \node[anchor=west, font=\bfseries\color{white}, scale=0.9]
            at ($(smallOrigin) + (0.15,\smallboxheight - 0.5*\headerheight)$)
            {\csname specialInsightHeaderText\label\endcsname};

        \node[anchor=north west, text width=\smallboxwidth, align=justify, scale=0.85]
            at ($(smallOrigin) + (0,\smallboxheight - \headerheight - 0.15)$)
            {\csname specialInsightBodyText\label\endcsname};
    }
    \node[anchor=center] 
      at ($(sigma_t_name.south |- sol_gen_box.east)$) 
      {\shortstack{$\mathcal{I}_t$ \\ \\ {\scriptsize\&} \\  \\ $\mathcal{E}_t$}};


    \setlength{\ypos}{\dimexpr \ypos - \headerheight - \smallboxheight - 2\innerymargin  +0.3cm\relax}
    
      \draw[fill=gray!15, draw=none]
      (0, \ypos - \headerheight - \smallboxheight - \innerymargin +0.3cm)
      rectangle ++(\solGenWidth  , \smallboxheight + \headerheight + 2\innerymargin -0.3cm) node[pos=0.5] (response_box) {};

    \draw[fill=orange!60, draw=none]
      (0, \ypos - \headerheight + \innerymargin) rectangle ++(\solGenWidth, \headerheight);
    \node[anchor=center, font=\bfseries\color{white}] 
      at (\smallboxgap + 0.5\innerwidth, \ypos - 0.5*\headerheight + \innerymargin) {LLM Response};

    \newlength{\dashlen}
    \setlength{\dashlen}{\dimexpr \smallboxheight + \innerymargin\relax}

    \foreach \i/\label in {0/A, 1/B, 2/C} {
        \setlength{\xpos}{\dimexpr \smallboxgap +\i\smallboxwidth + \numexpr 2 * \i\relax  \smallboxgap \relax}
        \coordinate (smallOrigin) at (\xpos, \ypos - \headerheight - 0.9\smallboxheight);
    
        \draw[fill=orange!15, draw=none] (smallOrigin) rectangle ++(\smallboxwidth, 0.9\smallboxheight );
        \draw[fill=orange!60, draw=none]
            (smallOrigin) ++(0, 0.9\smallboxheight - \headerheight) rectangle ++(\smallboxwidth,\headerheight);
    
        \node[anchor=west, font=\bfseries\color{white}, scale=0.9]
            at ($(smallOrigin) + (0.15,0.9\smallboxheight - 0.5*\headerheight)$)
            {\csname llmResponseTitleText\label\endcsname};
    
        \node[anchor=north west, text width=\smallboxwidth, align=justify, scale=0.85] (header\label)
            at ($(smallOrigin) + (0,0.9\smallboxheight - \headerheight - 0.15)$)
            {\csname llmResponseBodyText\label\endcsname $\planAngledsub{}$};
            \vspace{0.3em}

        \node[anchor=north west, fill=gray!15, inner sep=4pt, font=\ttfamily\scriptsize,
          text width=\smallboxwidth - 0.8cm, rounded corners=1pt]
          at ($(smallOrigin) + (0.25, 0.9\smallboxheight - \headerheight - 28)$) {
            \texttt{\`{}\`{}\`{}python}\\ 
            $\codeAngledsub{}$ \\ \`{}\`{}\`{}\\  
        };

        \ifnum \i<2
            \draw[dashed] 
              (\xpos + \smallboxwidth + \smallboxgap, \ypos - \headerheight + 0.2cm)  -- ++(0, -0.9\dashlen);
        \fi
    }
    \node[anchor=center] 
      at ($(sigma_t_name.south |- response_box.east)$) 
      {$z_t$};

    \setlength{\ypos}{\dimexpr \ypos - \smallboxheight - \headerheight - 2\innerymargin + 0.05cm\relax}
    
    \draw[fill=black]
      (0, \dimexpr \ypos - \headerheight - \smallboxheight - \innerymargin + 0.8cm\relax)
      rectangle ++(\solGenWidth, \dimexpr \smallboxheight + \headerheight + 2\innerymargin - 0.8cm\relax) node[pos=.5] (env_box) {};
    
    \draw[fill=black!30, draw=black!30]
      (0, \ypos - \headerheight + \innerymargin) rectangle ++(\solGenWidth, \headerheight);
    \node[anchor=center, font=\bfseries\color{white}] 
      at (\smallboxgap + 0.5\innerwidth, \ypos - 0.5*\headerheight + \innerymargin) {Environment (Run $\codeAngledsub{}$ and collect $\logAngledsub{}$)};
    
    \newlength{\termwidth}
    \setlength{\termwidth}{\dimexpr \solGenWidth - 1cm\relax} 
    
    \newlength{\termx}
    \setlength{\termx}{\dimexpr 0.5\solGenWidth - 0.5\termwidth\relax}
    
    \coordinate (termOrigin) at (\termx, \ypos - \headerheight - \smallboxheight + 0.5cm);

    \newlength{\consoleHeight}
    \setlength{\consoleHeight}{\dimexpr \smallboxheight - 0.5cm\relax}
    
    \node[anchor=north west, fill=black, text=white, font=\ttfamily\small, inner sep=3pt,
          text width=\termwidth, align=left ,text=green!30,]
      at ($(termOrigin) + (0, \consoleHeight)$) {
        \textcolor{green!60}{user@agk:\textasciitilde\$} python script.py \\
        Training model\dots \\
        $\vdots$ \\
        Validation MSE: 0.324
    };

    \node[anchor=center] 
      at ($(sigma_t_name.south |- env_box.east)$) 
      {$\mathcal{F}_t$};

    \setlength{\ypos}{\dimexpr \ypos - \smallboxheight - \headerheight - 2\innerymargin + 0.5cm\relax}
    
    \draw[fill=myforest2!10, draw=none]
      (0, \dimexpr \ypos - \headerheight - \smallboxheight - \innerymargin - 0.7cm\relax)
      rectangle ++(\solGenWidth, \dimexpr \smallboxheight + \headerheight + 2\innerymargin + 0.7cm\relax) 
      node[pos=.5] (update_box) {};

    \draw[fill=myforest2!60, draw=none]
      (0, \ypos - \headerheight + \innerymargin) rectangle ++(\solGenWidth, \headerheight);
    \node[anchor=center, font=\bfseries\color{white}] 
      at (\smallboxgap + 0.5\innerwidth, \ypos - 0.5*\headerheight + \innerymargin) {Internal State Update};
    
    \newlength{\subboxwidth}
    \setlength{\subboxwidth}{\dimexpr 0.5\solGenWidth - 1cm\relax}
    \newlength{\subboxheight}
    \setlength{\subboxheight}{3cm}
    
    \coordinate (contentOrigin) at (0, \ypos - \headerheight - \innerymargin + 0.1cm);
    
    \draw[fill=blue!10, draw=none, rounded corners=2pt]
      ($(contentOrigin) + (0.5cm, 0)$) rectangle ++(\subboxwidth, -\subboxheight);
    
    \draw[fill=blue!60, draw=none]
      ($(contentOrigin) + (0.5cm, 0)$) rectangle ++(\subboxwidth, -0.5cm);
    
    \node[anchor=west, font=\bfseries\small\color{white}, draw=none]
      at ($(contentOrigin) + (0.7cm, -0.25cm)$) {Prompt (Logs entity recognition)};
    
    \node[anchor=north west, font=\small, text width=\subboxwidth - 0.4cm]
      at ($(contentOrigin) + (0.7cm, -0.7cm)$)
      {From $\logAngledsub{}$ identify:};
    
    \draw[fill=gray!10, draw=none, rounded corners=1pt]
      ($(contentOrigin) + (0.7cm, -2.8cm)$) rectangle ++(\subboxwidth - 1.1cm, +1.5cm);
    
    \node[anchor=north west, font=\ttfamily\scriptsize, text width=\subboxwidth - 1.3cm, align=left, draw=none]
      at ($(contentOrigin) + (0.8cm, -1.3cm)$) {
    \{\\
    \ \ \texttt{"is\_successful"}: \texttt{"?"},\\
    \ \ \texttt{"score"}: \texttt{"?"},  ...\\
    \}
    };

    \draw[fill=orange!15, draw=none]
      ($(contentOrigin) + (\subboxwidth + 1.5cm, 0)$) rectangle ++(\subboxwidth, -\subboxheight);
    
    \draw[fill=orange!60, draw=none]
      ($(contentOrigin) + (\subboxwidth + 1.5cm, 0)$) rectangle ++(\subboxwidth, -0.5cm);
    
    \node[anchor=west, font=\bfseries\small\color{white}]
      at ($(contentOrigin) + (\subboxwidth + 1.7cm, -0.25cm)$) {LLM Answer};
    
    \node[anchor=north west, font=\small, text width=\subboxwidth - 0.4cm]
      at ($(contentOrigin) + (\subboxwidth + 1.7cm, -0.7cm)$)
      {
      \color{green!60!black}\texttt{<log\_elements>}: 
      };
    
    \draw[fill=gray!10, draw=none, rounded corners=1pt]
      ($(contentOrigin) + (\subboxwidth + 1.7cm, -2.8cm)$) rectangle ++(\subboxwidth - 1.1cm, +1.5cm);
    
    \node[anchor=north west, font=\ttfamily\scriptsize, text width=\subboxwidth - 1.3cm, align=left]
      at ($(contentOrigin) + (\subboxwidth + 1.8cm, -1.3cm)$) {
    \{\\
    \ \ \texttt{"is\_successful"}: \texttt{"true"},\\
    \ \ \texttt{"score"}: \texttt{"0.324"} ,...\\
    \}
    };

    \draw[->, thick,black]
      ($(contentOrigin) + (0.5cm + \subboxwidth, -0.5*\subboxheight)$)
      -- ($(contentOrigin) + (\subboxwidth + 1.5cm, -0.5*\subboxheight)$);

    \node[anchor=north, font=\small, align=center] 
      at ($(contentOrigin) + (0.5\solGenWidth, -\subboxheight)$) 
      {$\Sigma_{t+1}[\solPoolTer\texttt{\textrangle}]~\leftarrow~\{\codeAngledsub{},  \color{green!60!black}\texttt{<log\_elements>} \color{black}\}$};

    \node[anchor=center] 
      at ($(sigma_t_name.south |- update_box.east)$) 
      {$\mathcal{U}_t$};

\end{tikzpicture}
\caption{\textbf{Experiential Learning in Open-Ended Generation of \OurAgent{}.}}
\label{fig:exp-learning-post-scaffold}
\end{figure}

In the open-ended solution generation phase, the agent generates a diverse set of solution nodes, sequentially building a tree of submissions. 
The experiential learning loop is depicted in Figure~\ref{fig:exp-learning-post-scaffold}, showing the content of the internal state, the intrinsic and extrinsic functions, as well as the environment feedback and state update.

\textbf{Internal Agent State $\Sigma_t$:} The internal state consists of the task and data descriptions, the available device, the remaining runtime, and the list of the previously generated open-ended solutions along with their respective outputs. Moreover, $\Sigma_t$ contains the specialized CoT obtained during the scaffolded phase to guide the initial node generation to enable abstraction and guide the exploration of the solution space.  

\textbf{Solution generation through reflection and abstraction:} From internal state $\Sigma_t$, a new solution is obtained by prompting the LLM with the relevant elements of the state.
The prompting scheme defining the intrinsic and extrinsic functions $\mathcal{I \& \mathcal{E}}$ depends on whether the agent should \textbf{i)} generate an initial solution draft, \textbf{ii)} try to improve the  current best solution stored in $\Sigma_t$, or \textbf{iii)} debug a previously deficient code. 
The three cases are illustrated in Figure~\ref{fig:exp-learning-post-scaffold}, where \textbf{i)} in the first column, the scaffold CoTs are given in the prompt to guide the first solution generations, \textbf{ii)} in the middle column, the code of the best solution so far is provided, and \textbf{iii)} in the right-most column, the buggy implementation along with the error message are added to the prompt.

\textbf{Environment feedback $\mathcal{F}_t$:} the code generated by the agent is executed in the environment, which produces logs that are recorded to provide a feedback $\mathcal{F}_t$ to the agent.

\textbf{State update $\Sigma_{t+1} = \mathcal{U}_t(\Sigma^\prime_t, \mathcal{F}_t)$: } to update  its internal state, the update function involves a prompting mechanism to let the Agent analyse whether the feedback corresponds to a successful run or not, identify the validation metric, and whether this metric should be maximised or minimised, in order to be able to identify the best solution.

This experiential learning loop is repeated until we reach a specified total runtime or when a maximum number of solutions is produced.

\subsection*{Experimental Setup, Baselines, Resources}
We consider two families of baselines: 1) ReAct-based agents \cite{Jiang2024AIDE_ML_WECO}, and 2) foundational tabular prediction models \cite{hollmann2025accurate}.
Before detailing these two categories, we first provide insights into the composition of the Kaggle competitions we used as an evaluation set and on how we compute the performance quantiles.

\subsubsection*{Kaggle Competitions Set}

\begin{figure}[h!]
    \centering
    \begin{subfigure}[t]{0.5\textwidth}
        \centering
        \includegraphics[width=\linewidth,trim={.4cm .9cm 2.cm 2.8cm},clip]{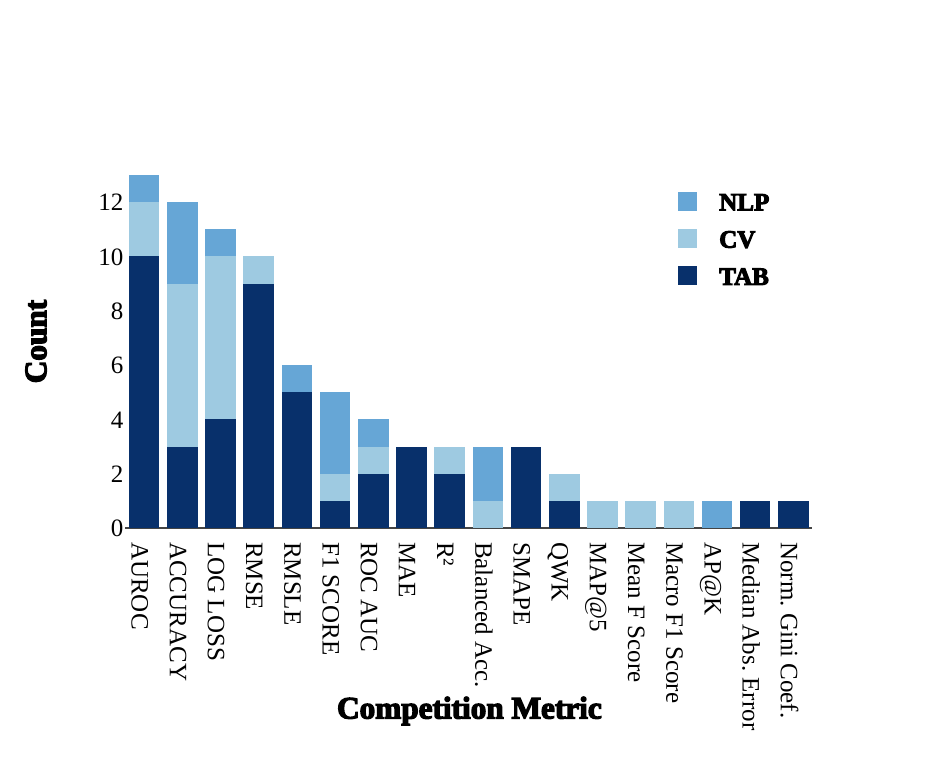}
        \caption{\textbf{Different Metric functions for all input modalities.}}
        \label{fig:benchmark_details:metrics}
    \end{subfigure}
    \hfill
    \begin{subfigure}[t]{0.44\textwidth}
        \centering
        \includegraphics[width=0.96\linewidth,trim={.0cm 1.2cm 1.7cm 2.3cm},clip]{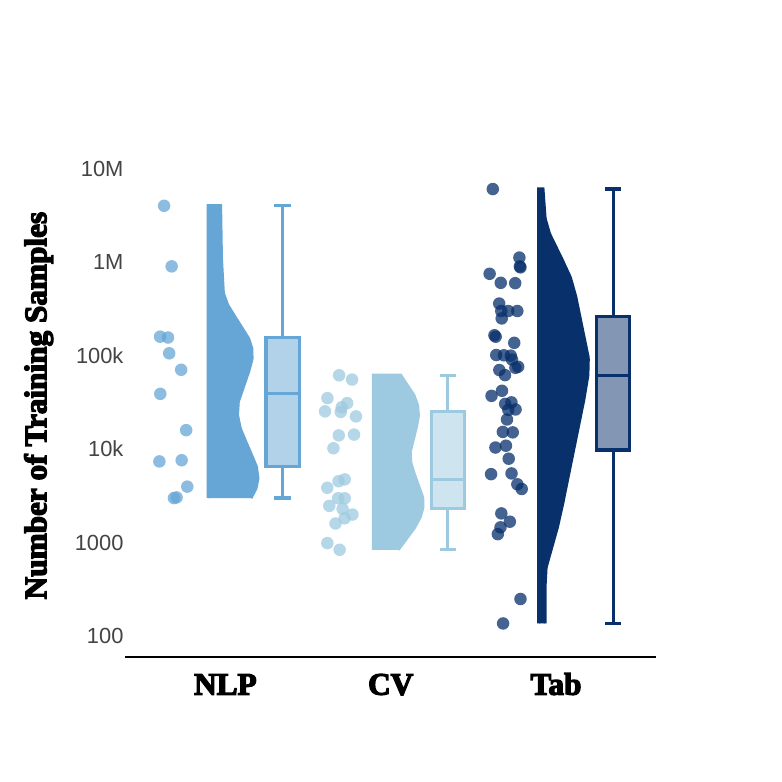}
        \caption{\textbf{Training samples per modality.}}
        \label{fig:benchmark_details:train_points}
    \end{subfigure}
    \begin{subfigure}{0.42\textwidth}
        \centering
        \includegraphics[width=0.96\linewidth,trim={.2cm 0.cm 2cm 2.5cm},clip]{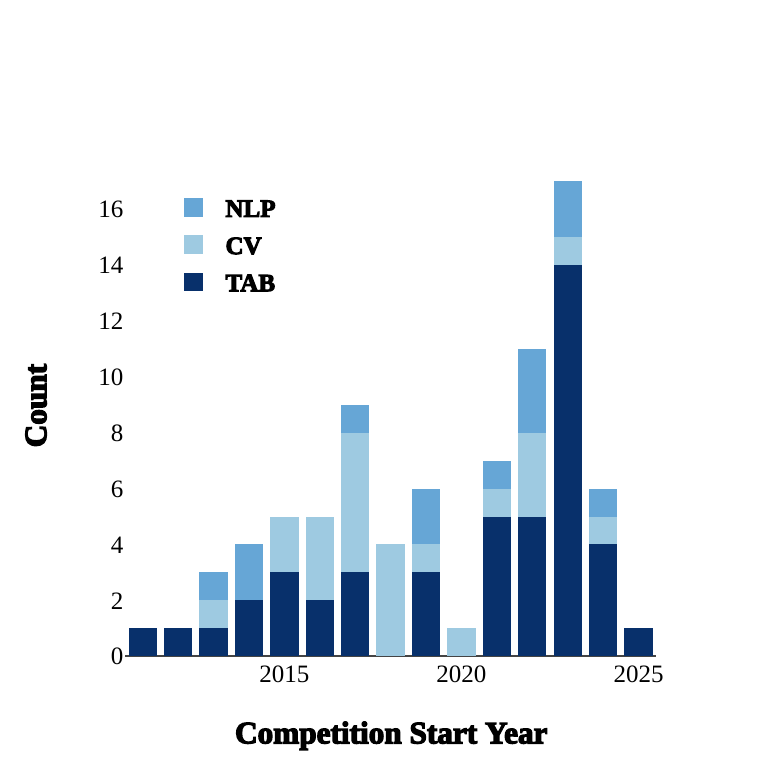}
        \caption{\textbf{Competition Start Dates.}}
        \label{fig:benchmark_details:start_date}
    \end{subfigure}
    \hfill
    \begin{subfigure}{0.56\textwidth}
        \centering
        \includegraphics[width=0.96\linewidth,trim={.3cm 1.2cm 1.7cm 2.3cm},clip]{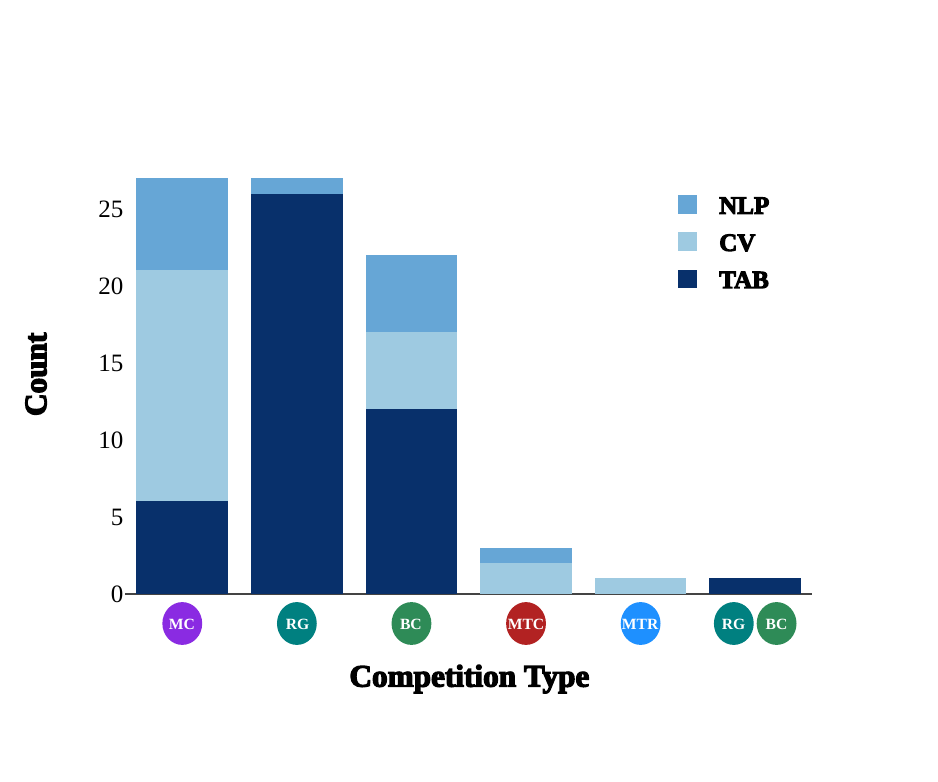}
        \caption{\textbf{Distribution of competition types.}}
        \label{fig:benchmark_details:competition_types}
    \end{subfigure}
    \caption{Overview of competition metrics, start dates, and types across different input modalities. Those are computed computed over the \NumBenchmarkTasks{} benchmark tasks we consider.}
    \label{fig:benchmark_details}
\end{figure}

All competitions included in our benchmark are listed on \url{kaggle.com} and are accessible through their API. 
Table~\ref{tab:benchmark_competitions_list} in the Appendix lists those competitions with their respective URL and Figure~\ref{fig:benchmark_details} shows different statistics of the selected tasks.
Figure~\ref{fig:benchmark_details:metrics} shows the varied sets of metrics we considered, where some are standard, such as RMSE, others are less common, e.g., median absolute errors, or quadratic weighted kappas.
Metrics relate to the nature of the competitions, whose distribution is presented in Figure~\ref{fig:benchmark_details:competition_types}.
Most competitions are regressions or classifications, with a few being more complex multi-target tasks.
Figure~\ref{fig:benchmark_details:start_date} presents the starting year of the competitions included in our benchmark spanning 2011 to 2025.
Finally, Figure~\ref{fig:benchmark_details:train_points} illustrates the scale of our competitions by showing the distribution of available labelled inputs. 
The dataset sizes range from a few hundred samples for competitions designed to emphasize overfitting risks, to several million examples, demonstrating that our benchmark aims to address data science challenges at real-world scale.

\subsubsection*{Performance Quantile Computation}
Given a competition $C$, let  $k_C$ be the number of submissions that any participant can decide to select as their final submissions, and assume that this competition uses a metric that should be minimised. 
If a method $A$ generated $n$ distinct valid submissions with public scores $p_1^\text{pub} \leq \dots \leq p_n^\text{pub}$ (potentially across different attempts), we need to assess its final performance based on at most $k_C$ of them.
To do so, we use a greedy selection process, i.e., we select the top-$\min(k_C, n)$ submissions and observe their private scores $p_1^\text{priv}, \dots, p_{\min(k_C, n)}^\text{priv}$ -- which are not necessarily in increasing order.
Finally, we consider $p_\text{final}^A = \max_{i\in\{1,\dots,\min(k_C, n\}} p_i$ as the final score of $A$, and compute the associated quantile by determining the fraction of participants who obtained a score better than $p_\text{final}^A$. If the final competition leaderboard contains $N$ entries with scores $s_1, \dots, s_N$:
\begin{equation}
    q^A = 100 - 100 \cdot \frac{|\{ i \in {1, \dots, N} \ |\  s_i < p_\text{final}^A \} |}{N}
\label{eq:calquantile}
\end{equation}
such that if $p_\text{final}^A$ matches or outperforms the best score in this competition, $q_A = 100$ and if it is worse than any score, $q_A = 0$.

\begin{table}[h]
\centering
\caption{\textbf{Kaggle Medals Attribution}. The thresholds follow \href{https://www.kaggle.com/progression}{Kaggle's guidelines}, and  the ``$\star$  (Top 10 + 0.2 \%) means that an extra gold medal will be awarded for every 500 additional teams in the competition.
For example, a competition with 500 teams will award gold medals to the top 11 teams, and a competition with 5000 teams will award gold medals to the top 20 teams.''}
\label{tab:KaggleProgression}
\begin{tabular}{>{\raggedright\arraybackslash}p{2.5cm} c c c c}
\toprule
\rowcolor{white} 
\textcolor{black}{\textbf{Medal}} & 
\textcolor{black}{\textbf{0-99 Teams}} & 
\textcolor{black}{\textbf{100-249 Teams}} & 
\textcolor{black}{\textbf{250-999 Teams}} & 
\textcolor{black}{\textbf{1000+ Teams}} \\
\hline
\hline
\rowcolor{bronze} 
\textcolor{black}{Bronze} & Top 40\% & Top 40\% & Top 100 & Top 10\% \\
\rowcolor{silver} 
\textcolor{black}{Silver} & Top 20\% & Top 20\% & Top 50 & Top 5\% \\
\rowcolor{gold} 
\textcolor{black}{Gold} & Top 10\% & Top 10 & Top 10 + 0.2\%* & Top 10 + 0.2\%* \\
\end{tabular}%
\end{table}

The medals we report are based on the Kaggle medals attribution system, which takes into account the final leaderboard quantile achieved and the number of participants.  
We provide the precise medal attribution rules in Table~\ref{tab:KaggleProgression}. Note that we apply this system even for competitions that did not award actual Kaggle medals.

\subsubsection*{ReAct-based agents}\label{Sec:ReActExpl}
ReAct-based agent corresponds to the second stage of \OurAgent{}, initialised without the scaffold chain-of-thought.
To ensure a fair comparison with our method, we assigned it a runtime budget equal to the combined budgets of both phases of \OurAgent{}, as shown in Table~\ref{tab:runtimes}.
Each version was executed for two attempts, as was \OurAgent{}.  
We take the implementation from~\cite{chan2024mlebenchevaluatingmachinelearning} and adapt the main hyperparameters to the different runtimes, as summarised in Table~\ref{tab:reactExplHyp}. 

\begin{table}[h]
    \centering
    \begin{tabular}{ccl}\toprule
         Hyperparameter&  Value &Role\\ \midrule
         $N_\text{max}$ &  5000 & Max number of iterations \\
         $\tau_\text{node}$ &  $3/16 \times \text{Total\_Runtime}$ &Max runtime per node \\
         $N_\text{draft}$&  5 &Max initial nodes allowed\\
 Max debug depth&3 &Max node debug iterations\\
 Probability of debug &  50 \% &Choose to debug node\\ \bottomrule
    \end{tabular}
    \caption{ReAct-based agent hyperparameters. We report $\tau_\text{node}$ as a function of the total runtime, which depends on the nature of the competition, as specified in Table~\ref{tab:runtimes}. These hyperparameters are also used for the post-scaffold stage of \OurAgent{}.}
    \label{tab:reactExplHyp}
\end{table}

\paragraph{Shallow ReAct-based agents}
We evaluate ReAct augmented with a tree-exploration strategy as defined in \cite{Jiang2024AIDE_ML_WECO}.
We run this method with Qwen2.5-72B, the same model used in \OurAgent{} experiments. 
To further assess the merit of our scaffolded approach compared to a stronger reasoning model, we ran ReAct-based agent equipped with Deepseek-R1 using the same hyperparameters.

\paragraph{ReAct (Qwen) + RAG}
To address the concern that any Data-science-specific chain-of-thought could independently lead to performance as strong as the ones we achieve with \OurAgent{}, we evaluated another variant of ReAct-based agent, where the initial solution generation prompts are guided using examples drawn from a Kaggle-based RAG database.  
We construct this database of cases similar to \cite{Guo2024DS-Agent}, selecting high-quality discussions and notebooks from past Kaggle competitions. These documents are pre-processed into structured entries containing notebook summaries and technical discussions , and are indexed for semantic retrieval. 
The agent is allowed to generate up to $N_\text{draft}$ new solution nodes per problem and is allowed to retrieve the top N most relevant cases from the database based on semantic similarity to the current problem description. 
Each new node incorporates one of these retrieved examples into the prompt, applied sequentially, allowing the agent to leverage concrete prior cases while formulating a new solution. 
When the agent revisits or improves existing nodes, it does so without additional retrieval, ensuring that external knowledge is only introduced during the generation of new branches. 
This protocol ensures a disciplined and competition-based prompting strategy that facilitates the reuse of relevant knowledge while preserving internal consistency in solution development. 

\subsubsection*{Tabular Foundation Models}\label{Sec:TabPFN}

Our agent was evaluated on several tabular benchmarks, with its performance compared to TabPFN v2 \cite{hollmann2025accurate} and other variants of this model. 
TabPFN is a state-of-the-art tabular foundation model trained on a large set of synthetic tasks. 
It predicts by processing a sequence of labelled examples without requiring additional parameter updates during inference. 
As this model by itself does not handle the full data science pipeline, we run it from the automated setup conducted by \OurAgent{}, to focus the comparison on the predictions generation quality.

Since TabPFN v2 supports a maximum of 10,000 input samples, we used a K-Means-based sampling strategy to adapt larger datasets to this constraint. For each batch of 10,000 test samples, we applied K-Means clustering on the test set and used the resulting cluster centres to select a representative subset of 10,000 samples from the training data. 
These selected training points, along with the corresponding test batch, were then provided as input to TabPFN v2. 
The model’s predictions were aggregated across all batches to generate the final submission files. 
While our own K-means strategy proved effective, we also compared \OurAgent{}'s to other TabPFN extensions (TabPFN-Ext.) that were released in the literature to address its limitations.

\paragraph{TabICL:} TabICL \cite{qu2025tabicl} is another foundation model for tabular data, designed to handle datasets with up to 100,000 samples on affordable hardware. Currently, it supports only classification tasks. We included TabICL in our benchmarking experiments to provide an additional point of comparison for \OurAgent{}.

\paragraph{TabPFN Fine-Tuning:} To further enhance TabPFN's performance, we also conducted a full fine-tuning of the vanilla TabPFN V2 models for both classification and regression tasks. Each vanilla model was fine-tuned on the training set of the specific competition, and the resulting model was used to make predictions on the test set. Due to the 10,000-sample input constraint, we again used K-Means-based sampling to divide the training set into N subsets, each containing S samples. 
Even though we used a different sample size $S$ for training due to GPU memory limitations while maintaining model fidelity, we kept 10,000 samples while performing the inference.

\subsubsection*{Computational Resources}
\OurAgent{} was evaluated on modest hardware to emphasise its accessibility and potential for broad democratisation. Each experiment was executed in an isolated container running Ubuntu 22.04.3 LTS. For each job, compute resources were limited to a single NVIDIA V100 GPU (32 GB memory) and 9 Intel Xeon CPU cores. Furthermore, Table~\ref{tab:runtimes} reports the runtime limits assigned to \OurAgent{} and each baseline across different competition types, ensuring that all methods operated under equivalent time constraints for a fair comparison.

\definecolor{agentkColor}{RGB}{200,220,255}  
\definecolor{reactColor}{RGB}{255,220,220}   
\definecolor{pfnColor}{RGB}{220,255,220}     

\begin{table}
    \centering
    \begin{tabular}{clcc}\toprule
         Method & Model & \makecell{Runtime Limit\\(Tabular)} & \makecell{Runtime Limit\\(CV/NLP)} \\ \midrule
         
         \rowcolor{agentkColor}
         \OurAgent{} (Scaffold \& Beyond ZPD) & Qwen2.5-72B & 1 day \& 1 day & 2 days \& 2 days \\
         
         \rowcolor{reactColor}
         ReAct (Qwen) & Qwen2.5-72B & 2 days & 4 days \\
         \rowcolor{reactColor}
         ReAct (Deepseek R1) & Deepseek-R1 & 2 days & 4 days \\
         \rowcolor{reactColor}
         ReAct (Qwen) + RAG & Qwen2.5-72B & 2 days & 4 days \\
         
         \rowcolor{pfnColor}
         TabPFN-v2 Fine-Tuned & TabPFN-v2 & 2 days & - \\
         
         \bottomrule
    \end{tabular}
    \caption{Runtimes per method for Tabular and CV/NLP competitions, grouped by method type. For the Tab-PFN baselines we only report the time limit for TabPFN-v2 Fine-Tuned as for the other versions we let it run without specifying a time limit (in practice they run within at most a few hours).}
    \label{tab:runtimes}
\end{table}

\subsubsection*{Statistical Tests}\label{Sec:Stats}
In order to compare the results obtained by different agents, we conducted several statistical tests.
We display in Figure~\ref{Fig:Res1}  the commonly used critical difference (CD) diagrams to visualise the statistical difference of the results from different methods.
The CD diagram summarises the result of multiple pairwise comparisons.
Its construction starts with a Friedman test which is used to detect if there are statistically significant differences in the performance ranks of the compared methods. 
If this test fails, it means that there is a possibility that the hypothesis that all methods have the same average performance cannot be rejected.
On the other hand, if the test succeeds, we move towards  a post-hoc analysis via pairwise comparisons with a Wilcoxon signed-ranked test, which allows us to  determine which pairs of methods are significantly different.

Finally, the data used to perform the tests are the percentiles achieved on the competitions leaderboards (as detailed in the Performance Quantile Computation section), considering all the competitions where each method  made at least one valid submission.
They are transcribed into relative ranks and averaged over each method.
The CD diagram therefore shows at the same time the relative rank of each method (x-axis) and the pairwise statistical difference of the results. 
The latter is denoted by a horizontal bar that denotes that two methods joined by that bar are not statistically different.

\subsection*{Discussion on technology timeline}

\begin{figure}[h!]
\centering
\includegraphics[width=.6\linewidth,trim={.2cm .7cm 2cm 2.3cm},clip]{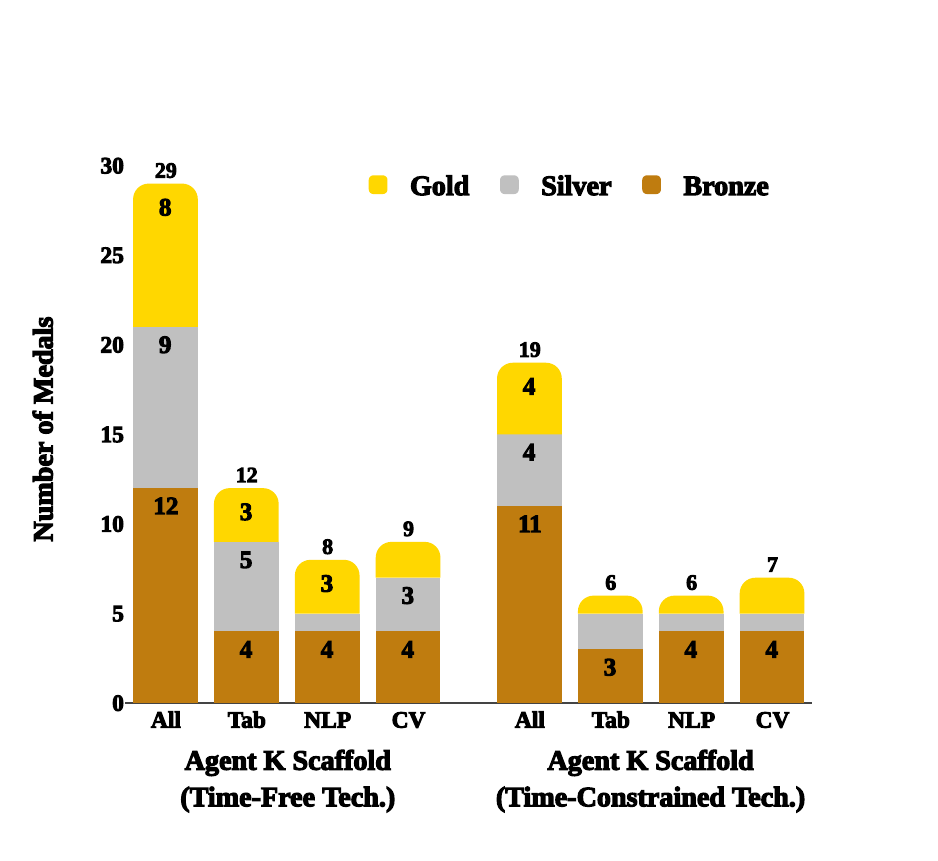}
\caption{\textbf{Medal Distribution by Modality}. On the left we show the number of medals achieved on all competitions without time-restricting the models used by \OurAgent{} in its solutions. On the right we show the same count when restricting the models and the version of their weights if needed, to be consistent with the competition timeline.}
\label{fig:time_consistent_medal}
\end{figure}

During the scaffold, the LLM generates fusion models using a modality-dependent embedder. 
One limitation when comparing to human participants is that, depending on the competition date, not all models and pre-trained weights were available at the time. Figure~\ref{fig:time_consistent_medal} shows that \OurAgent{} manages to get medals even when it uses technology only available at the time of the competition. Even when technology is constrained the agent is still capable of achieving \nMedalsTimeConstrained{} medals.

\paragraph{Time-controlled performance on CV and NLP competitions}
At scaffold phase, \OurAgent{} often relies on existing architectures and pre-training weights which were not released at the time of some Kaggle competitions. 
To assess the impact of using more advanced technology, we identify for each scaffold submission which architectures, as well as the version of the pre-trained weights that are used, and check their release dates to compare them with the competition end date. 
If we remove all submissions made using a model not available during the competition period, we see that \OurAgent{} still secures \nMedalsTimeConstrainedCV{} medals on CV competitions, and \nMedalsTimeConstrainedNLP{} medals on NLP competitions, showing that the good performance of our agent is not only attributable to the use of recent technology.

\paragraph{Time-controlled performance on Tabular tasks.}
On tabular tasks, the agent generates submissions at scaffold phase using RAMP tool, which is based on optimising and blending several base models from \texttt{lightgbm}~\cite{Ke2017LightGBM}, \texttt{xgboost}~\cite{Chen2016xgboost} and \texttt{catboost}~\cite{Prokhorenkova2018CatBoost}. 
Considering submissions made on competitions posterior to these base models still yields a total of \nMedalsTimeConstrainedTAB{} medals.

\subsection*{Data Availability}
In our experiments, we rely on Kaggle competitions (listed in Table~\ref{tab:benchmark_competitions_list} in the Appendix) whose data and leaderboards are publicly available on the dedicated Kaggle page of each competition.
Moreover, we make the RAG database available, uploading a zip file in the  repository where we host  our agent implementation.

\subsection*{Code Availability}
We open sourced our code on GitHub under MIT license\footnote{\url{https://github.com/huawei-noah/HEBO/tree/dev-agent/Agent_K}}.
Our repository includes the complete implementation of \OurAgent{} (for both scaffold and post-scaffold stages), the scripts to run the ReAct-based agent baselines, and the pre-processing and post-processing code we designed to run the various TabPFN experiments.
Additionally, we curated a documentation with the setup instructions for each type of experiments, and also detailing the configurations and hyperparameters we used, in order to ease replicating our results.

\subsection*{Author contributions}
Jun Wang and Haitham Bou-Ammar conceived the research direction and made key scientific decisions on human-inspired learning theory for agentic systems with applications to data science.

Antoine Grosnit and Alexandre Maraval designed the agent architecture and scaffold mechanisms and implemented its core structure.

Refinath S N, James Doran, Jonas Gonzalez, Abhineet Kumar, and Khyati Khandelwal implemented the different elements involved in the data-science agent, and extensively tested by Zichao Zhao.  

\OurAgent{} experiments on CV and NLP competitions were handled by Alexandre Maraval, and the ones on Tabular competitions were run by Abhineet Kumar. 
ReAct-based agent baselines were evaluated by Alexandre Maraval (with Qwen) and Refinath S N (with Deepseek-R1 and with RAG). TabPFN baselines were integrated and run by Refinath S N.

Bal\'{a}zs K\'{e}gl, Albert Thomas, Giuseppe Paolo, Abdelhakim Benechehab, Hamza Cherkaoui, and Youssef Attia El-Hili  developed the RAMP-based AutoML tool used in the scaffold for tabular tasks.

Jianye Hao, Jun Yao, and Shao Kun provided strategic oversight and high-level guidance throughout the project.

\printbibliography
\end{refsection}

\newpage

\appendix

\ifthenelse{\boolean{submissionMode}}{%
  \setcounter{page}{1}%
  \resetlinenumber[1]%
}{
}

\begin{refsection}

\section{Related Work}\label{Sec:RelatedWork}

\subsection{Data-science benchmarking}

Assessing agents' performance on data science tasks has been approached from various perspectives. 
While many benchmarks, such as those for coding proficiency \cite{jimenez2024swebench, zhuo2024bigcodebench}, focus on general code generation capabilities, they do not necessarily capture the full range of skills required for end-to-end data science.
Assessing the performance of agents on data-science tasks has been considered from different scopes.

Some benchmarks are tailored to specific aspects of data science. For example, \cite{Lai2022DS1000} focuses on code infilling based on data from Stack Overflow and common libraries, targeting localised code generation rather than comprehensive pipeline development.

Unlike traditional AutoML benchmarks \cite{Gijsbers2019OpenSourceAutomMLBench}, which typically provide a standardised input format to facilitate solver design, data science agent benchmarks evaluate broader capabilities. These include understanding the task requirements and adapting solutions accordingly, offering flexibility in problem formulation and solution strategies.

Authors in \cite{Huang2023MLAgentBench} introduce a benchmark using six Kaggle tasks, where success is defined as achieving a 10\% improvement over a baseline provided by starter code. 
However, this baseline can sometimes represent minimal predictive performance, such as random predictions, rather than a robust comparison against established leaderboards. 
This benchmark also offers predefined scripts and environments for each task, guiding the problem-solving approach.
Other benchmarks, such as \cite{Cao2024spider2v}, emphasise the orchestration of workflows and interactions with professional software tools, while \cite{Hu2024InfiAgentDABenchEA} provides a carefully curated set of questions requiring language models to generate code that answers specific dataset-related queries.
This differs from end-to-end problem-solving in competitions like Kaggle, whose questions are primarily focused on tabular data.
Additionally, \cite{tang2024mlbench} centres on understanding machine learning repositories rather than the broader data science process, which limits its applicability to end-to-end DS agent evaluation.
\cite{zhang2024benchmarkDSAgent} uses Kaggle datasets to derive specific questions about dataset characteristics, such as imbalance, missing data, and correlations. 
Although this approach provides a focused evaluation, it does not capture the full scope of capabilities required for complete data science workflows, which is a broader objective we aim to achieve.

Closer to our experimental study, \cite{Jing2024DSBenchHF} curates a list of 74 Kaggle competitions to assess the data modelling capacity of LLM agents. 
Nevertheless, the authors pick the benchmark competitions based on the simplicity of their raw structures, featuring only a train, test and sample submission file.
Moreover, they do not support comparison with the Kaggle leaderboard but split the training set to create a custom test set instead.
Likewise, MLE-bench~\cite{chan2024mlebenchevaluatingmachinelearning} includes 75 machine learning competitions from Kaggle to assess agent performance by comparing it against user submissions. However, because this assessment relies on a custom split of the training data rather than the competition's actual private test set, it remains uncertain whether an agent’s high ranking in this context would align with results on the original Kaggle leaderboard.

Our experiments aim to truly assess the end-to-end capacity of agents to achieve high performance in data-science tasks autonomously. 

\subsection{Traditional automated data science (Auto-DS)}
Achieving performance and automation in data science tasks has been a long-standing challenge. 
Numerous techniques have been developed to automate the human-intensive steps of the DS pipeline using machine learning.
Given that each step in the DS pipeline poses challenges, many approaches have focused on specific aspects of the problem, often resulting in narrow application ranges or use-case limitations.
For example, probabilistic inference rules have been used to automate data cleaning~\cite{Rekatsinas2017HoloCleanDataRepairs}, while various deep learning methods have been developed to filter, select, or transform features to enhance the accuracy of downstream models~\cite{Kanter2015DeepFS, Kaul2017AutoLearnA, Khurana2016Cognito}.
In the modelling domain, a significant research area known as neural architecture search (NAS) optimises the structure of neural networks for predictions~\cite{liu2018darts, White2023NeuralAS}, shifting the manual effort from model design to the design of a model search space.
Beyond neural architecture, many hyperparameters can be automatically tuned to achieve better performance using black-box optimisation methods.
Techniques like Bayesian optimisation address black-box optimisation in a sample efficient way and are supported by various libraries~\cite{akiba2019optuna, Balandat2020Botorch, CowenRivers2022HEBO, dreczkowski2023framework}, which users can utilise by correctly identifying and providing the hyperparameters of their pipelines. 
Additionally, meta-learning has been introduced to further improve data efficiency by leveraging past experiments to predict performance on new tasks based on related tasks solved with similar configurations~\cite{Hospedales2022SurveyMetaLearning, maraval2023endtoend, Moharil2024TowardsEffAutoML}.

Contrasting with these targeted methods, several libraries have been developed to cover broader parts of the data
science pipeline, such as H2O-AutoML~\cite{LeDell2020H2OAutoML}, AutoKeras~\cite{Jin2023AutoKeras}, Auto-sklearn~
\cite{Feurer2020Auto-sklearn}, and AutoGluon~\cite{tang2024autogluonMultimodal}. 
These tools aim to automate and optimise by making rule-based decisions based on input data modality and combining a fixed collection of models that can be refined using preset optimiser tools. 
However, these approaches have limitations, including restricted scope, flexibility, and usability barriers. 
Their performance is limited by the set of predefined models and hyperparameters they support, and they require expert knowledge to properly format the problem~\cite{Chen2023SEED} (e.g., manual specification of column types for AutoKeras). 
Tools like H2O-AutoML are also limited to tabular data tasks, requiring data to be structured in specific ways.

\subsection{Data-science Agents}
The emergence of data-science agents powered by large language models and multi-agent systems represents a
shift towards more autonomous and adaptable solutions. Unlike traditional Auto-DS approaches, these agents are designed
to perform end-to-end tasks by leveraging advanced reasoning and decision-making capabilities.

\subsubsection{Coding Agents}
Recent advancements in LLMs~\cite{Achiam2023GPT4TR, Jiang2023Mistral7, meta2024llama3,
Ren2023PanGuTT}, combined with techniques like chain of thought prompting~\cite{tutunov2023can,
Zheng2023CodeGeex} and ReAct~\cite{Yao2022ReAct} have significantly expanded the scope of AI applications beyond natural language processing.
These models are now capable of generating code and interacting with APIs~\cite{Deng2023FuzzTesting, Shen2023PanGuCoder2BL, Qin2023ToolLLMFL}, laying the groundwork for their integration into multi-agent systems~\cite{Guo2024MultiAgentLLM,
hong2024metagpt, OpenInterpreter2023, xagent2023, wu2023autogen} that can execute complex tasks by thinking, planning, calling tools~\cite{Qu2024ToolLW}, and making decisions~\cite{Huang2022TowardsRI, Li2022InteractiveDecisisonMaking, Sha2023LanguageMPCLL, Yao2023TreeOfThoughts} across various domains, including  robotics~\cite{mandi2023roco, Mower2024ROSLLMAR, Singh2023ProgPromt}, medicine~\cite{tang2024medagents, Thirunavukarasu2023LargeLM}, and also data science~\cite{Guo2024DS-Agent, Hong2024DataInterpreter, zhang2024mlcopilot}.
DS agents exploit the intrinsic capacities of LLM-Agents. They are augmented with specific tools for writing, executing and debugging code,  data visualisation tools~\cite{yang2024matplotagent}, or access to standard ML libraries APIs~\cite{Shen2023HuggingGPT}, offering a more integrated approach to data science
automation.  

\subsubsection{Intelligent DS assistants}

Several agents function as intelligent assistants, enhancing human decision-making by providing automated solutions while keeping users in the loop.
For instance, \texttt{CleanAgent}~\cite{Qi2024CleanAgent} handles data cleaning based on user specifications, and \texttt{LAMBDA}~\cite{sun2024lambda} generates data analysis reports based on technical queries.
\texttt{MatPlotAgent}~\cite{yang2024matplotagent} pairs a coding agent with a visual agent to meet the expert's needs in terms of data visualisation, while \texttt{WaitGPT}~\cite{xie2024waitgpt} offers a smoother monitoring of the code generation by creating graph representations of the generated code and supporting on the fly manual edition or questions.
Although these agents enhance user efficiency, they require substantial user input. They are not designed to autonomously achieve high performance in complex, real-world scenarios, such as winning a Kaggle competition.

\subsection{End-to-end DS agents}

End-to-end data-science agents aim to manage the entire data science pipeline autonomously, from data preprocessing to model evaluation and hyperparameter optimisation, without requiring significant human oversight.
While some existing approaches address specific aspects of this workflow, such as \texttt{CAAFE}~\cite{hollmann2023CAAFE} which allows agents to iteratively generate codes to modify tabular features and reflect based on accuracy feedback, or \texttt{EVAPORATE-CODE+}~\cite{Arora2023DataMiningAgent}, which focuses on data mining by generating code to extract structured data from semi-structured documents, other methods have tackled the entire data science pipeline more holistically.

Recent multi-agent systems like  \texttt{SEED}~\cite{Chen2023SEED}, \texttt{AutoM3L}~\cite{luo2024AutoM3L}, \texttt{Data Interpreter}~\cite{Hong2024DataInterpreter}, \texttt{AutoML-Agent}~\cite{Trirat2024AutoMLAgent}, \texttt{HuggingGPT}~\cite{Shen2023HuggingGPT}, \texttt{MLCopilot}~\cite{zhang2024mlcopilot}, \texttt{DS-Agent}~\cite{Guo2024DS-Agent}, \texttt{SELA}~\cite{Chi2024SELA}, \texttt{MLZero}~\cite{Fang2025MLZeroAM},  \texttt{AutoKaggle}~\cite{Li2024AutoKaggle} and \texttt{AIRA-dojo}~\cite{toledo2025ai} achieve capacities to deal with data-science by empowering their system with numerous tools, using retrieval from expert demonstrations, and supporting complex multiple steps reasoning. 


These systems employ various strategies for generating automated pipelines for different stages of the data science process such as preprocessing, modelling, and feature selection.

While \texttt{DS-Agent}, \texttt{MLCopilot} , \texttt{AIDE} and \texttt{AIRA-dojo} generate and refine the entire solution script as a single block, \texttt{SEED} performs a skyline-based optimisation to select the modules to apply to solve the task.
Alternatively, \texttt{HuggingGPT} delegates the planning to a high-level agent, and  \texttt{Data-Interpreter} lets this agent decide on a hierarchical and dynamic plan.
Meanwhile, \texttt{AutoM3L} follows a pre-defined modality-specific pipeline to produce final predictions. 
Our data-science agent integrates elements from these various approaches by imposing a modality-dependent structure during the  setup scaffold phase while allowing agents greater flexibility in selecting the optimal sequence of steps to achieve the best possible outcome.

Moreover, several existing systems also incorporate mechanisms for continual learning, enabling them to leverage past experiences to improve performance on new tasks.
For instance, \texttt{DS-Agent} achieves that by adopting a case-based reasoning approach combined with an offline-online setup, retrieving insights during the online phase from a pool of scripts that performed well on offline tasks.
Similarly, \texttt{MLCopilot} divides its framework into offline and online phases, using an LLM to infer verifiable machine learning strategies from diverse offline experiments, such as determining valid learning rate ranges for specific models.
Beyond building a knowledge database to retrieve from, \texttt{Data-Interpreter} uses prior experiences to create new tools that the agents can later call.

However, these systems theoretically or practically suffer from partial coverage or limited capacities.
Indeed, \texttt{AutoM3L} cannot generate solutions involving model ensembling and requires specifically structured input data, while \texttt{HuggingGPT} lacks support for model training and tabular models.
Moreover, \texttt{SEED} and \texttt{DS-Agent} do not demonstrate capabilities for handling computer vision tasks, and \texttt{AIDE}, \texttt{AutoKaggle}, and \texttt{Data Interpreter} have only shown strong performance on more straightforward tabular datasets or Kaggle competitions.
\texttt{AutoML-Agent} requires comprehensive task-specific prompts which can be advantageous for experts customising its performance but poses a challenge for less-experienced users.
Additionally, while some efforts have been made to utilise past experiments for new task resolution, the strategic selection of task order to enable progressive learning remains an underexplored area.
We aim to address these gaps by developing a system capable of handling various input modalities and achieving consistently high performance in real-world data science competitions. 


Beyond these practical continual learning strategies, there is a rich literature on online learning and online convex optimization~\cite{Hazan2016OnlineLearning} that provides strong theoretical guarantees for regret minimization in sequential decision-making. 
These methods assume a well-defined and typically convex loss structure, with each step involving an update based on newly observed feedback.
While valuable for formal analysis, such frameworks are generally restricted to lower-dimensional, well-specified optimisation problems and do not directly address the high-dimensional, multi-modal, and tool-rich environment faced by end-to-end DS agents in real-world competitions. 
Our approach instead integrates flexible decision-making across heterogeneous subtasks, where the feedback signal may be sparse, delayed, or only indirectly comparable (e.g., via leaderboard or validation scores).



\section{\OurAgent{} - Pipeline Details} \label{sec:technical-details}
According to our formulation, the design of the agent framework is broken down into two main parts: the \textit{setup} pipeline and the \textit{solution generation} pipeline. 
The setup pipeline's goal is to preprocess the raw information and raw data to prepare a workspace and a \texttt{DataLoader} object directly usable by the solution generation pipeline to solve the task. 
All the steps related to downloading, summarising, organising and cleaning the data are done in the setup pipeline, whereas steps related to feature engineering, model training, hyperparameter optimisation and submission belong to the prediction pipeline.
In this section we give more detail about the pipeline steps, their objective and important aspects of how they work.

\subsection{Setting up Competitions} \label{subsec:setup}
The goal is to automate the process of setting up a task. 
The pipeline, presented in Figure \ref{fig:env-scaffold} depicts the sequence of stages that \OurAgent{} tackles in order to setup a new Kaggle competition, given its URL. 

\subsubsection{Expected Result - Usable \& Interpretable Workspace}\label{subsec:setup_expected_results}
At the end of a successful setup, we expect to have a \textit{workspace} containing useful files, including raw information and raw data from the competition website, as well as new files created by \OurAgent{}, such as metadata, different CSV files as well as runnable code files. 
Our agent creates additional CSV files only if it deems it necessary from the context of the competition:
\begin{itemize}
    \item \texttt{train\_tab\_input\_map.csv} - tabular inputs from the train set
    \item \texttt{train\_txt\_input\_map.csv} - text inputs from the train set
    \item \texttt{train\_img\_input\_map.csv} - paths to the image inputs from the train set
    \item \texttt{train\_tab\_target\_map.csv} - tabular targets from the train set
    \item \texttt{train\_txt\_target\_map.csv} - text targets from the train set
    \item \texttt{train\_img\_target\_map.csv} - paths to the image targets from the train set
    \item \texttt{test\_tab\_input\_map.csv} - tabular inputs from the test set
    \item \texttt{test\_txt\_input\_map.csv} - text inputs from the test set
    \item \texttt{test\_img\_input\_map.csv} - paths to the image inputs from the test set
\end{itemize}

When \OurAgent{} produces code to generate the CSV files, this code is  kept in the workspace, which contributes to the interpretability of the result.
Furthermore, our agent also need to implement the useful functions:
\begin{itemize}
    \item \texttt{code\_transform\_tab\_target\_train.py} - code to transform the targets into a format usable for training models and back to the original format
    \item \texttt{code\_metric.py} - the metric function implementation
    \item \texttt{code\_submission\_format.py} - code to transform from the predicted targets format to the exact format accepted by the competition (e.g. from a probability of a class to the label of that class)
    \item \texttt{code\_submission\_format\_alt.py} - an alternative to the previous function, in case it fails. It can also be used as a backup for the solution generation pipeline when we format the predictions of the model for submission.
\end{itemize}

\subsubsection{Setup Pipeline Steps}
\paragraph{Fetch \& Scrape Raw Data} Given the URL of the competition, \OurAgent{} can use the Kaggle API to download its associated data. 
The agent is also given a scraping tool which automatically retrieves information from the competition's web page, such as the raw description of the competition, its data description, metric description, etc.

\paragraph{Summarise} From the scraped raw descriptions, \OurAgent{}  creates digestible summaries that it can later reuse for  LLM prompting. 
The raw text from the competition pages can contain elements that are irrelevant to the task (e.g. emojis, images, urls, \dots) or worse, information that is misleading.
Summarising the raw descriptions can deal with those issues and enables the agent to compress the important information from the task into a more useful and shorter text that takes less tokens in the future prompts.

\paragraph{Detect Modalities} We consider three main modalities: \textit{vision} data, \textit{language} data and \textit{tabular} data, which can be purely tabular or contain time-series.
The agent categorises all competitions' inputs and targets with respect to these three modalities, based on the summarised descriptions.
For instance, a competition where participants are asked to create an image classifier would have image inputs and tabular targets (classification).
In another case, if a competition consists in attributing a toxicity score to movie reviews and includes as input the reviews as well as the categories of the movies, then the input modalities would be a mixture of text and tabular, while the modality of the target would be tabular (regression). 

\paragraph{Create Plan \& Metadata} After the raw data is fetched, the summaries are written and the modalities are detected, \OurAgent{} creates its plan, i.e. its workflow.
This plan determines the sequence of steps that should be done in order to complete the setup. 
This plan is also \textit{dynamic} in the sense that our agent can decide to prioritise stages over others provided it is permitted by the general sequence order of the environment.
\OurAgent{} can come back to previous stages when executing its plan, and it can attempt some stages multiple times if they haven't been done properly the first time, as we detail in the paragraph dedicated to unit tests.

\paragraph{Create Maps \& Transforms} The main task in the setup pipeline is for the agent to create \textit{maps} and \textit{transforms}, as outlined in Section \ref{subsec:setup_expected_results}. The idea is to split all raw training data into input and target maps, as well as to split all raw data (train and test) into different modalities. 
A task could have multiple types of inputs and targets so we require a map for each of them. 
These maps are then collated together in a custom \texttt{DataLoader} object that makes it easy to consume by a model in the prediction pipeline.
Importantly, the agent is also tasked with creating transform functions in order to treat targets as consumable data for a model. 
For instance if the targets in the training set are text-encoded labels (in a classification task) the \textit{transform} function could map these to a one-hot encoding and the \textit{inverse transform} function would map them back to the text labels. 
These functions are crucial to enable the prediction pipeline.

\paragraph{Create Submission Format \& Metric Function} Finally, \OurAgent{} generates code to create the submission format and the metric function.
Both are generally described in the raw competition description. 
The submission format is the exact format expected by the competition to read the predictions. It can sometimes be different from the format of the raw train targets so in addition to the transforms, we also need a function that maps the predicted targets on the test inputs, in the exact format expected.

\subsubsection{Unit Tests}\label{App:UnitTests}
An important part of the setup pipeline is the set of unit tests used to assert if a stage or a group of stages has been successfully done by the agent.
It is worth noting that these unit tests are manually designed and not generated by the agent itself.
However, \textbf{this does not mean that they are tailored to each task}.
In fact they are designed to be agnostic to the competition to setup and are rather specific to the framework itself.
Knowing the properties of the expected result of a particular stage, we can test if the code or the data our agent generate follows an expect format.
This way our agent receives insightful feedback through error messages when its code is faulty, which it integrates into its reasoning process to correct itself.

For example, suppose our agent generates a runnable code to create the training input map of the image modality.
During the state transition of the environment, the unit test associated to this stage is ran. 
The unit test checks that a CSV file with a specific name exists and is loads it.
It then assesses that the table has a column named ``\texttt{id}'' and that it has at least one row.
It also checks the existence of a column apart from the ``\texttt{id}'' column and that these other columns contain paths to images with known extensions.
Moreover, the unit test  throws an error if one of the image paths listed in that map is invalid.
Finally it can check various trivial things such as non emptiness of columns or absence of duplicated columns. 
This shows that the unit tests are specific to the setup environment and not specific to the task. 

If an error is raised in the unit test, the error is saved in memory and the current stage is attempted again. 
In this new trial, our agent will be able modify and fix its previous code based on the unit test error message that is accessible in its memory.
This is repeated until the unit test passes or a budget limit is reached (in which case the setup would fail there).

Finally we have \textit{meta}-unit tests which are unit tests that depend on multiple stages.
They verify that the result of multiple single stages coordinate well together.
If a \textit{meta}-unit test fails, the agent has the opportunity to go back multiple stages and re-attempt the relevant stages.
For instance, after all maps and transforms are done for the training data, we test if the \texttt{DataLoader} can be constructed properly and if we can successfully load a batch. 
This step is crucial as it validates that all the maps generated by the agent up to this point share a common ``\texttt{id}'' column so the \texttt{DataLoader} can collate them properly.
If this test fails, the agent has to go back to the start and figure out which codes to correct given the error raised in the \textit{meta}-unit test to eventually correct its mistakes.

This paragraph highlights the usefulness of the unit tests, showing that they  do not only serve as checks for the agent, but that they also provide useful error messages to fuel effective reflection.
The \textit{meta}-unit tests also provide additional verifications that can only be done once a group of stages have passed their own individual unit tests.
However, there is always a limit to what we can test.
It is possible that the agent finds a way to pass all unit tests but ends up setting up the competition incorrectly.
Because we want our unit tests to remain task-agnostic we cannot extend our coverage indefinitely.
In the future we would like to extend this work to also use the agent to generate additional unit tests that would be specific to the current competition being setup.
In that way we can extend the coverage at inference-time.

\subsection{Generating Solutions}




After the agent has generated code that constitutes a fully-functioning solution, and while training that solution, we allow for some optimisations and support such as creating additional submission files to give \OurAgent{} more choice.
These are part of the scaffold and are not directly coded by the agent itself but are there rather as a design choice.

First, we do Bayesian Optimisation (BO) over some parameters, i.e. the learning rate and the optimiser choice. 
The training is run for a maximum of \texttt{n\_trials} trials, each trial being limited to \texttt{max\_epochs} epochs or  \texttt{max\_time}, whichever is attained first.

We then generate Cross-Validated submissions by training the model with the best learning rate and optimiser found during BO, on \texttt{k\_folds} folds of the training set. 
We then average the predictions to create a new submission file.
For the competitions involving images, we also add a submission file created via simple Test-Time-Adaptation (TTA).
This method augments the data at test time with the same data augmentation pipeline used for the training dataloader.
Using the best checkpoint saved from the training, we reproduce this for \texttt{tta\_rounds} rounds and average the predictions to create a new submission file.

Finally, blending is used after at least \texttt{blend\_after} full working and different solutions have been generated.
Values are specified in Table \ref{tab:hyperparameters}.

\begin{table}
    \caption{Hyperparmeters used for the training of the solutions generated during the scaffold stage.}
    \label{tab:hyperparameters}
    \begin{center}
    \begin{tabular}{c|c|l}
        \hline 
        Name & Value & Role \\
        \hline
        \texttt{max\_epochs} & 30 & Max training epochs allowed \\
        \texttt{max\_time} & 10h & Max training time allowed \\
        \texttt{batch\_size} & 32 & Training \& testing batch size \\
        \texttt{lr} & $[1e-6, 1e-2]$ & BO search space for learning rate \\
        \texttt{optimizer} & \{Adam, SGD, AdamW\} & BO search space for optimiser \\
        \texttt{n\_trials} & 20 & Max BO rounds allowed \\
        \texttt{k\_folds} & 5 & Cross-Validation folds \\
        \texttt{tta\_rounds} & 4 & Number of TTA rounds averaged \\
        \texttt{blend\_after} & 3 & max nb. of solutions before blending \\
        \hline
    \end{tabular}
    \end{center}
\end{table}

\subsection{Pipeline Pseudocodes}
\begin{algorithm}[h!]
\caption{Pseudocode for the Setup Pipeline}\label{pseudocode:setup}
\begin{algorithmic}
    \State \textbf{Inputs}: URL
    \Statex
    \State $\mathcal{D} \gets$ FetchAndScrapeData(URL)
    \State $\mathcal{S} \gets$ SummariseDataAndTaskDetails($\mathcal{D}$)
    \State $\{\mathcal{X}, \mathcal{Y}\} \gets$ DetectModalities($\mathcal{S}$)
    \State $\mathcal{P} \gets$ CreatePlan($\{\mathcal{X}, \mathcal{Y}\}, \mathcal{S}$)
    \Statex\Comment{\textit{Every} \textbf{while} loop in the pipeline is subject to a retry limit}
    \While{{\textit{meta}-unit test fails}}
        \While{unit test fails}
            \State $\mathcal{M}^{\text{train}}_{\text{input}} \gets$ CreateTrainInputMaps($\mathcal{D}, \mathcal{S}, \{\mathcal{X}, \mathcal{Y}\}, \mathcal{P}$)
        \EndWhile
        \While{unit test fails}
            \State $\mathcal{M}^{\text{train}}_{\text{target}} \gets$ CreateTrainTargetMaps($\mathcal{D}, \mathcal{S}, \{\mathcal{X}, \mathcal{Y}\}, \mathcal{P}$)
        \EndWhile
        \While{unit test fails}
            \State $\mathcal{T}^{\text{train}}_{\text{target}} \gets$ CreateTrainTargetTransforms($\mathcal{D}, \mathcal{S}, \{\mathcal{X}, \mathcal{Y}\}, \mathcal{P}$)
        \EndWhile
    \EndWhile
    \Statex
    \While{\textit{meta}-unit test fails}
        \While{unit test fails}
            \State $\mathcal{M}^{\text{test}}_{\text{input}} \gets$ CreateTestInputMaps($\mathcal{D}, \mathcal{S}, \{\mathcal{X}, \mathcal{Y}\}, \mathcal{P}$)
        \EndWhile
    \EndWhile
    \Statex
    \While{unit test fails}
        \State $\mathcal{F}_{\text{submission}} \gets$ CreateSubmissionFormat($\mathcal{D}, \mathcal{S}, \{\mathcal{X}, \mathcal{Y}\}, \mathcal{P}$)
    \EndWhile
    \Statex
    \While{unit test fails}
        \State $\mathcal{F}_{\text{metric}} \gets$ CreateMetric($\mathcal{D}, \mathcal{S}, \{\mathcal{X}, \mathcal{Y}\}, \mathcal{P}$)
    \EndWhile
    \Statex
    \State \textbf{Outputs}: Train and test dataloaders $\texttt{DL}_{\text{train}}$ and $\texttt{DL}_\text{{test}}$
\end{algorithmic}
\end{algorithm}

We give in Algorithms \ref{pseudocode:setup} and \ref{pseudocode:solution_generation} the pseudocode for the setup pipeline and the pseudocode for the solution generation pipeline, respectively.
Note that the input of the setup pipeline is the competition URL and the outputs are the setup files and the dataloaders.
The solution generation pipeline takes as inputs the outputs of the setup pipeline and carries on generating code to create one or more submission files.
The output of the solution generation pipeline is a set of solutions along with their performance on the leaderboard.

\paragraph{Retry Budget} Note that in the pseudocode, there are steps such as ``\textbf{while} \dots~\textbf{end while}''.
In practice, we do not retry each step indefinitely, there is an upper bound on the number of retries we allow \OurAgent{} to do for each step.
As soon as this number is reached we consider the agent to be stuck and we end the episode by raising an error.

\begin{algorithm}[h!]
\caption{Pseudocode for the Solution Generation Pipeline}\label{pseudocode:solution_generation}
\begin{algorithmic}
    \State \textbf{Inputs}: $\mathcal{S}_{\text{setup}} \gets \{\mathcal{D}, \mathcal{S}, \{\mathcal{X}, \mathcal{Y}\}, \mathcal{P}, \mathcal{M}^{\text{train}}_{\text{input}}, \mathcal{M}^{\text{train}}_{\text{target}}, \mathcal{M}^{\text{test}}_{\text{input}}, \mathcal{T}^{\text{train}}_{\text{target}}, \mathcal{F}_{\text{submission}}, \mathcal{F}_{\text{metric}}\}$, $\texttt{DL}_{\text{train}}$ and $\texttt{DL}_\text{{test}}$, Embedders $\mathcal{E} \gets \{\}$, Heads $\mathcal{H} \gets \{\}$, Solutions $Sol \gets \{\}$
    \Statex
    \If{\textit{tabular-only} task}
        \State AdoptClassicalMLFramework()
        \State $Sol \gets$ UseRampTool($\mathcal{S}_{\text{setup}}, \texttt{DL}_{\text{train}}, \texttt{DL}_{\text{test}}$)
    \Else
        \State AdoptDNNFramework()
        \While{Agent decides to improve performance}
            \Statex\Comment{Create Input Encoder(s)}
            \If{$\mathcal{S}_{\text{setup}}$ has tabular inputs}
                \State $\mathcal{F}_{\text{f.e.}} \gets$ TabularFeatureEngineering($\mathcal{S}_{\text{setup}}, \texttt{DL}_{\text{train}}$)
                \State $\mathcal{E}_{\text{tab}} \gets$ TabularEmbedding($\mathcal{S}_{\text{setup}}, \texttt{DL}_{\text{train}}$)
                \State $\mathcal{E} \gets \mathcal{E}~\cup~\{\mathcal{F}_{\text{f.e.}}, \mathcal{E}_{\text{tab}}\}$
            \EndIf
            \If{$\mathcal{S}_{\text{setup}}$ has image inputs}
                \State $\mathcal{F}_{\text{img}} \gets$ ImageTransforms($\mathcal{S}_{\text{setup}}, \texttt{DL}_{\text{train}}$)
                \State $\mathcal{E}_{\text{img}} \gets$ ImageEmbedding($\mathcal{S}_{\text{setup}}, \texttt{DL}_{\text{train}}$)
                \State $\mathcal{E} \gets \mathcal{E}~\cup~\{\mathcal{F}_{\text{img}}, \mathcal{E}_{\text{img}}\}$
            \EndIf
            \If{$\mathcal{S}_{\text{setup}}$ has text inputs}
                \State $\mathcal{F}_{\text{txt}} \gets$ TextPreprocessing($\mathcal{S}_{\text{setup}}, \texttt{DL}_{\text{train}}$)
                \State $\mathcal{E}_{\text{txt}} \gets$ TextEmbedding($\mathcal{S}_{\text{setup}}, \texttt{DL}_{\text{train}}$)
                \State $\mathcal{E} \gets \mathcal{E}~\cup~\{\mathcal{F}_{\text{txt}}, \mathcal{E}_{\text{txt}}\}$
            \EndIf
            \Statex\Comment{Create Target Head(s)}
            \State $\mathcal{H}_{\text{tab}} \gets$ TabularHead($\mathcal{S}_{\text{setup}}$)
            \State $\mathcal{H} \gets \mathcal{H}~\cup~\{\mathcal{H}_{\text{tab}}\}$
            \Statex\Comment{Generate \texttt{submission.csv} and get validation score}
            \State $\texttt{sub}, val \gets \text{TrainSolutionAndPredict}(\mathcal{S}_{\text{setup}}, \texttt{DL}_{\text{train}}, \texttt{DL}_{\text{test}}, \mathcal{E}, \mathcal{H})$
            \State $Sol \gets Sol~\cup~\{\texttt{sub}\}$
            \If{enough solutions created}\Comment{Blend existing solutions}
                \State $\texttt{sub}, val \gets \text{BlendSolutionsAndPredict}(\mathcal{S}_{\text{setup}}, \texttt{DL}_{\text{train}}, \texttt{DL}_{\text{test}}, \mathcal{E}, \mathcal{H}, Sol)$
                \State $Sol \gets Sol~\cup~\{\texttt{sub}\}$
            \EndIf
        \EndWhile
        \Statex\Comment{Send submissions and get private score and rank}
        \State $r \gets \text{SendSubmissionandGetLeaderboardRank}(\texttt{sub})$
    \EndIf
\end{algorithmic}
\end{algorithm}

\section{Competition difficulty analysis}\label{sec:competition-difficulty}

As detailed in the ``Quantitative Results'' section, \OurAgent{}'s achievements include several gold medals across different types of competitions — community, playground, research, and featured — however, some of these competitions are less complex than others, and not all competitions award medals according to the Kaggle system. In this section, we provide further insights on competitions' difficulties by measuring the levels of their participants.
To do so, we analyse the Elo scores of all Kaggle participants, going through all the medal-awarding competitions on Kaggle, and updating the Elo-MMR of the competitors based on their private ranks (the Elo score of a new participant being initialised at 1,500).
Doing so, we can access for each competition (medal-awarding or not), the Elo scores of the participants at the time this competition ended.

\begin{figure}[h!]
\centering
\includegraphics[trim={0.1cm 0.cm 0.3cm 2cm},clip,width=0.95\linewidth]{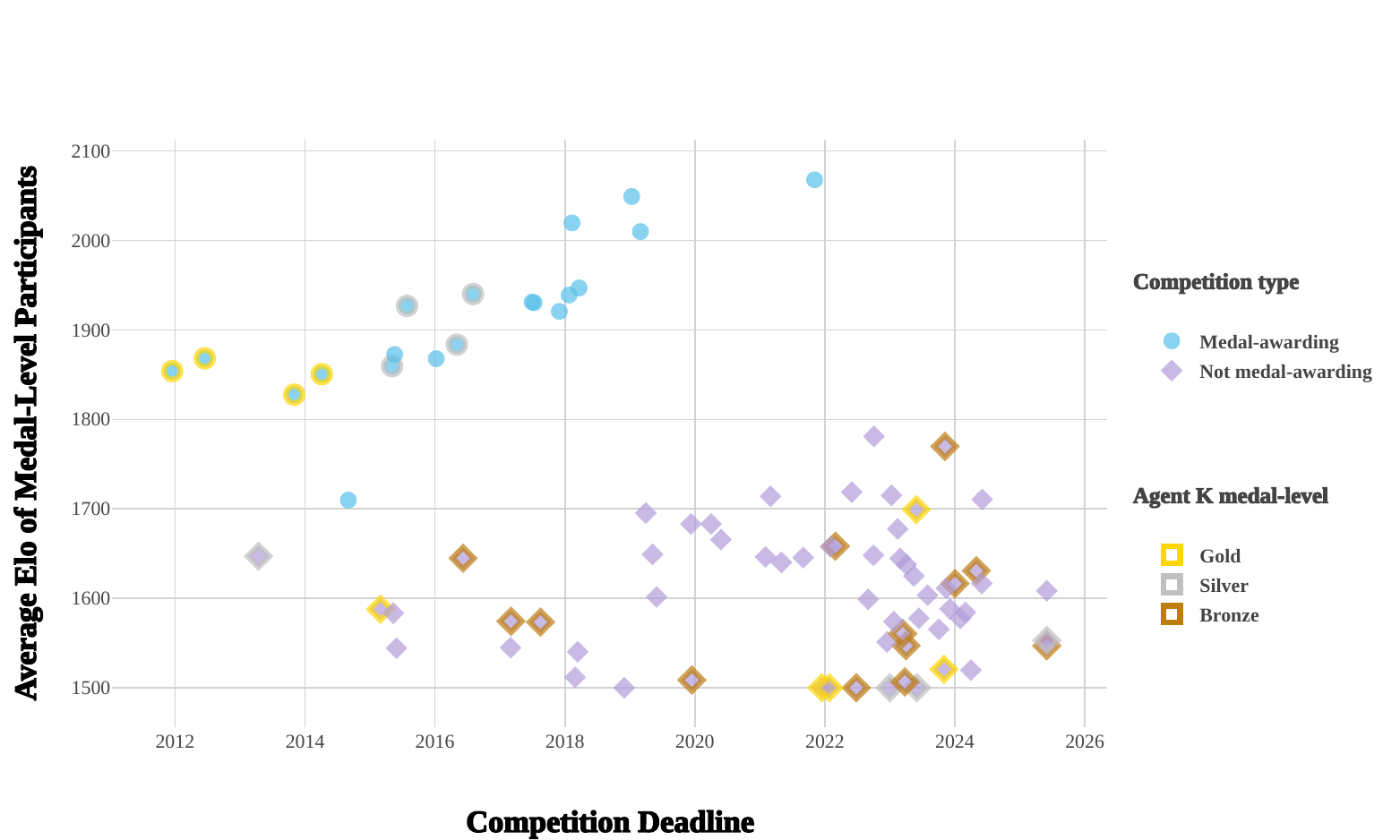}
\caption{\textbf{Competition difficulty analysis.} Each marker in the plot represents a Kaggle competition included in our benchmark. The x-axis indicates the competition's closing date, while the y-axis shows the average Elo score of participants who achieved at least a bronze medal (or a bronze medal-level score for non-medal awarding competitions).
For competitions in which \OurAgent{} attained a medal-level score, the marker is outlined in the corresponding medal color.
Notably, since \OurAgent{} achieved \nGoldMedalsOnMedAwardingComps{} gold medals across medal-awarding competitions, the figure includes \nGoldMedalsOnMedAwardingComps{} blue circular markers with gold outlines.
}
\label{Fig:competition-difficulty}
\end{figure}
We present the results of our analysis on Figure~\ref{Fig:competition-difficulty} where we display for each competition included in our benchmark, the average Elo scores of the participants that have reached at least a bronze medal-level score.
We observe a gap between the average Elo score on the non-medal-awarding competitions (purple diamond markers) and on the medal-awarding ones (blue circle markers), which confirms the idea that stronger data scientists tend to participate more in medal-awarding competitions.  
Moreover, we highlight the medal-level scores achieved by \OurAgent{} by outlining the competition markers in the colors of the medals.
For instance, the \nGoldMedalsOnMedAwardingComps{} gold (resp. \nSilverMedalsOnMedAwardingComps{} silver) medals obtained by \OurAgent{} on medal-awarding competitions appear as golden (resp. silver) disks on the figure, with their y-position matching the average Elo of the participants who won at least a bronze medals on these competitions.

This analysis pushes us to further improve our performance on the competitions involving stronger data-scientists.

\section{Benchmark details}
We provide the list of competitions included in our benchmark in Table~\ref{tab:benchmark_competitions_list}.

\newcommand{\gettingstarted}{%
  \begin{tikzpicture}[baseline=-0.5ex]
    \node[circle, fill={rgb:red,1.0000; green,0.5490; blue,0.0000}, text=white, inner sep=0pt, minimum size=1.2em, text width=1.3em, align=center] {\tiny {\textbf{GS}}};
  \end{tikzpicture}%
}
\newcommand{\playground}{%
  \begin{tikzpicture}[baseline=-0.5ex]
    \node[circle, fill={rgb:red,0.5020; green,0.0000; blue,0.5020}, text=white, inner sep=0pt, minimum size=1.2em, text width=1.3em, align=center] {\tiny {\textbf{P}}};
  \end{tikzpicture}%
}
\newcommand{\featured}{%
  \begin{tikzpicture}[baseline=-0.5ex]
    \node[circle, fill={rgb:red,0.0000; green,0.0000; blue,0.5020}, text=white, inner sep=0pt, minimum size=1.2em, text width=1.3em, align=center] {\tiny {\textbf{F}}};
  \end{tikzpicture}%
}
\newcommand{\community}{%
  \begin{tikzpicture}[baseline=-0.5ex]
    \node[circle, fill={rgb:red,0.1333; green,0.5451; blue,0.1333}, text=white, inner sep=0pt, minimum size=1.2em, text width=1.3em, align=center] {\tiny {\textbf{C}}};
  \end{tikzpicture}%
}
\newcommand{\research}{%
  \begin{tikzpicture}[baseline=-0.5ex]
    \node[circle, fill={rgb:red,0.0000; green,0.5451; blue,0.5451}, text=white, inner sep=0pt, minimum size=1.2em, text width=1.3em, align=center] {\tiny {\textbf{RS}}};
  \end{tikzpicture}%
}
\newcommand{\recruitment}{%
  \begin{tikzpicture}[baseline=-0.5ex]
    \node[circle, fill={rgb:red,0.0000; green,0.0000; blue,0.0000}, text=white, inner sep=0pt, minimum size=1.2em, text width=1.3em, align=center] {\tiny {\textbf{RC}}};
  \end{tikzpicture}%
}
\newcommand{\Regression}{%
  \begin{tikzpicture}[baseline=-0.5ex]
    \node[circle, fill={rgb:red,0.0000; green,0.5020; blue,0.5020}, text=white, inner sep=0pt, minimum size=1.2em, text width=1.3em, align=center] {\tiny {\textbf{RG}}};
  \end{tikzpicture}%
}
\newcommand{\MulticlassClassification}{%
  \begin{tikzpicture}[baseline=-0.5ex]
    \node[circle, fill={rgb:red,0.5412; green,0.1686; blue,0.8863}, text=white, inner sep=0pt, minimum size=1.2em, text width=1.3em, align=center] {\tiny {\textbf{MC}}};
  \end{tikzpicture}%
}
\newcommand{\MultitargetClassification}{%
  \begin{tikzpicture}[baseline=-0.5ex]
    \node[circle, fill={rgb:red,0.6980; green,0.1333; blue,0.1333}, text=white, inner sep=0pt, minimum size=1.2em, text width=1.3em, align=center] {\tiny {\textbf{MTC}}};
  \end{tikzpicture}%
}
\newcommand{\BinaryClassification}{%
  \begin{tikzpicture}[baseline=-0.5ex]
    \node[circle, fill={rgb:red,0.1804; green,0.5451; blue,0.3412}, text=white, inner sep=0pt, minimum size=1.2em, text width=1.3em, align=center] {\tiny {\textbf{BC}}};
  \end{tikzpicture}%
}
\newcommand{\MultitargetRegression}{%
  \begin{tikzpicture}[baseline=-0.5ex]
    \node[circle, fill={rgb:red,0.1176; green,0.5647; blue,1.0000}, text=white, inner sep=0pt, minimum size=1.2em, text width=1.3em, align=center] {\tiny {\textbf{MTR}}};
  \end{tikzpicture}%
}

\begin{table}[H]
\centering
\caption{\textbf{Symbols associated to competition types and competition categories on Kaggle.}}
\label{tab:symbol-legend}
\begin{tabular}{cc|cc}
\toprule
\textbf{Competition Type} & \textbf{Meaning} & \textbf{Kaggle Category} & \textbf{Meaning} \\
\midrule
\Regression & Regression & \gettingstarted & Getting Started \\
\BinaryClassification & Binary Classification & \playground & Playground \\
\MulticlassClassification & Multiclass Classification & \featured & Featured \\
\MultitargetRegression & Multitarget Regression & \community & Community \\
\MultitargetClassification & Multitarget Classification & \research & Research \\

\bottomrule
\end{tabular}
\end{table}

\begin{longtable}{p{5cm}
  c
  c
  c
  c
  c}
\caption{\textbf{List of the Kaggle competitions included in our benchmark.} Each row represents a Kaggle competition, showing its Kaggle ID (first column), its type, its category on Kaggle, the input modalities it involve,  the number of submissions that can be retained for the final scoring (\#Select Subs), and the number of  teams that participated in this competition. Note that the column \# Select Subs corresponds to $k_c$ in equation~\eqref{eq:calquantile}. See Table \ref{tab:symbol-legend} for the symbols appearing in the Competition Type and Kaggle Type columns.} \\
\hline\hline
\textbf{Competition} & \makecell{\textbf{Competition}\\\textbf{Type}} & \makecell{\textbf{Kaggle}\\\textbf{Category}} & \textbf{Modality} & \makecell{\textbf{\#Select}\\\textbf{Subs}} & \makecell{\textbf{\#Teams}} \\
\hline\hline
\endfirsthead
\hline\hline
\textbf{Competition} & \textbf{Task type} & \makecell{\textbf{Kaggle}\\\textbf{Type}} & \textbf{Modality} & \makecell{\textbf{\#Select}\\\textbf{Subs}} & \makecell{\textbf{\#Teams}} \\
\hline\hline
\endhead
nlp-getting-started & \BinaryClassification & \gettingstarted & Tab,Txt & $\geq 100$ & 856 \\
sentiment-analysis-on-movie-reviews & \MulticlassClassification & \playground & Txt & 2 & 860 \\
dogs-vs-cats-redux-kernels-edition & \BinaryClassification & \playground & Img & 2 & 1314 \\
dog-breed-identification & \MulticlassClassification & \playground & Img & 1 & 1280 \\
state-farm-distracted-driver-detection & \MulticlassClassification & \featured & Img & 2 & 1438 \\
histopathologic-cancer-detection & \BinaryClassification & \playground & Img & 2 & 1149 \\
whereami & \MulticlassClassification & \community & Img & 20 & 211 \\
5-flowers-image-classification & \MulticlassClassification & \community & Img & 20 & 69 \\
sign-language-image-classification & \MulticlassClassification & \community & Img & 2 & 41 \\
ml-olympiad-landscape-image-classification & \MulticlassClassification & \community & Img & 2 & 13 \\
noaa-right-whale-recognition & \MulticlassClassification & \research & Img & 2 & 364 \\
humpback-whale-identification & \MulticlassClassification & \featured & Img & 2 & 2120 \\
planttraits2024 & \MultitargetRegression & \research & Tab,Img & 1 & 398 \\
invasive-species-monitoring & \BinaryClassification & \playground & Img & 2 & 511 \\
tabular-playground-series-feb-2021 & \Regression & \playground & Tab & 2 & 1433 \\
playground-series-s3e11 & \Regression & \playground & Tab & 2 & 952 \\
playground-series-s3e14 & \Regression & \playground & Tab & 2 & 1875 \\
playground-series-s3e15 & \Regression & \playground & Tab & 2 & 693 \\
home-data-for-ml-course & \Regression & \gettingstarted & Tab & 2 & 106930 \\
bike-sharing-demand & \Regression & \playground & Tab & 2 & 3242 \\
tabular-playground-series-aug-2021 & \Regression & \playground & Tab & 2 & 1753 \\
forest-cover-type-prediction & \MulticlassClassification & \playground & Tab & 2 & 1692 \\
mercedes-benz-greener-manufacturing & \Regression & \featured & Tab & 2 & 3823 \\
goodreads-books-reviews-290312 & \MulticlassClassification & \community & Tab,Txt & 5 & 261 \\
sberbank-russian-housing-market & \Regression & \featured & Tab & 2 & 3264 \\
restaurant-revenue-prediction & \Regression & \featured & Tab & 2 & 2257 \\
scrabble-player-rating & \Regression & \playground & Tab & 2 & 301 \\
nlp1000-ml-challenge & \BinaryClassification & \community & Txt & 2 & 16 \\
nlpsci & \MulticlassClassification & \community & Txt & 20 & 44 \\
sf-crime & \MulticlassClassification & \playground & Tab & 2 & 2331 \\
otto-group-product-classification-challenge & \MulticlassClassification & \featured & Tab & 2 & 3507 \\
nitro-language-processing-3 & \BinaryClassification & \community & Txt & 3 & 45 \\
sbu-ai-finalproject & \MulticlassClassification & \community & Txt & 5 & 37 \\
avito-prohibited-content & \BinaryClassification & \featured & Tab,Txt & 2 & 284 \\
playground-series-s3e1 & \Regression & \playground & Tab & 2 & 689 \\
playground-series-s3e9 & \Regression & \playground & Tab & 2 & 765 \\
playground-series-s3e16 & \Regression & \playground & Tab & 2 & 1429 \\
playground-series-s3e25 & \Regression & \playground & Tab & 2 & 1632 \\
tabular-playground-series-jan-2021 & \Regression & \playground & Tab & 2 & 1728 \\
tabular-playground-series-aug-2022 & \BinaryClassification & \playground & Tab & 2 & 1888 \\
tabular-playground-series-apr-2021 & \BinaryClassification & \playground & Tab & 2 & 1244 \\
playground-series-s4e4 & \Regression & \playground & Tab & 2 & 2606 \\
ventilator-pressure-prediction & \Regression & \research & Tab & 2 & 2605 \\
playground-series-s4e5 & \Regression & \playground & Tab & 2 & 2788 \\
porto-seguro-safe-driver-prediction & \BinaryClassification & \featured & Tab & 2 & 5156 \\
playground-series-s3e3 & \BinaryClassification & \playground & Tab & 2 & 665 \\
santander-customer-satisfaction & \BinaryClassification & \featured & Tab & 2 & 5115 \\
playground-series-s3e5 & \Regression & \playground & Tab & 2 & 901 \\
cat-in-the-dat & \BinaryClassification & \playground & Tab & 2 & 1338 \\
playground-series-s4e1 & \BinaryClassification & \playground & Tab & 2 & 3632 \\
tabular-playground-series-sep-2022 & \Regression & \playground & Tab & 2 & 1381 \\
cat-in-the-dat-ii & \BinaryClassification & \playground & Tab & 2 & 1161 \\
playground-series-s3e22 & \MulticlassClassification & \playground & Tab & 2 & 1541 \\
playground-series-s3e26 & \MulticlassClassification & \playground & Tab & 2 & 1661 \\
playground-series-s3e7 & \BinaryClassification & \playground & Tab & 2 & 678 \\
jigsaw-toxic-comment-classification-challenge & \MultitargetClassification & \featured & Txt & 2 & 4539 \\
kaggle-pog-series-s01e03 & \MulticlassClassification & \community & Tab,Img & 2 & 91 \\
vehicle & \MulticlassClassification & \community & Img & 2 & 65 \\
plant-seedlings-classification & \MulticlassClassification & \playground & Img & 1 & 833 \\
sp-society-camera-model-identification & \MulticlassClassification & \featured & Img & 2 & 580 \\
plant-pathology-2020-fgvc7 & \MulticlassClassification & \research & Img & 2 & 1317 \\
world-championship-2023-embryo-classification & \BinaryClassification & \community & Img & 2 & 111 \\
leaf-classification & \MulticlassClassification & \playground & Tab,Img & 2 & 1595 \\
nitro-language-processing-2 & \MulticlassClassification & \community & Txt & 3 & 46 \\
playground-series-s3e24 & \BinaryClassification & \playground & Tab & 2 & 1908 \\
playground-series-s4e2 & \MulticlassClassification & \playground & Tab & 2 & 3587 \\
tmdb-box-office-prediction & \Regression & \playground & Tab,Txt & 2 & 1395 \\
playground-series-s3e19 & \Regression & \playground & Tab & 2 & 1172 \\
tabular-playground-series-may-2022 & \BinaryClassification & \playground & Tab & 2 & 1151 \\
tabular-playground-series-jan-2022 & \Regression & \playground & Tab & 2 & 1591 \\
dont-overfit-ii & \BinaryClassification & \playground & Tab & 2 & 2315 \\
statoil-iceberg-classifier-challenge & \BinaryClassification & \featured & Tab,Img & 2 & 3330 \\
diabetic-retinopathy-detection & \MulticlassClassification & \featured & Img & 2 & 660 \\
stumbleupon & \BinaryClassification & \featured & Tab,Txt & 2 & 624 \\
h2oai-predict-the-llm & \MulticlassClassification & \community & Txt & 2 & 92 \\
human-protein-atlas-image-classification & \MultitargetClassification & \featured & Img & 2 & 2160 \\
predict-who-is-more-influential-in-a-social-network & \Regression \BinaryClassification & \featured & Tab & 5 & 132 \\
bioresponse & \Regression & \featured & Tab & 5 & 698 \\
GiveMeSomeCredit & \Regression & \featured & Tab & 5 & 924 \\
galaxy-zoo-the-galaxy-challenge & \MultitargetClassification & \research & Img & 5 & 326 \\
playground-series-s5e5 & \Regression & \playground & Tab & 2 & 4316 \\
\label{tab:benchmark_competitions_list}
\end{longtable}
\vspace{0.5em}

\section{Additional results}

\begin{figure}[h!]
\centering
\includegraphics[width=.8\linewidth,trim={.1cm .5cm 2cm 1.5cm},clip]{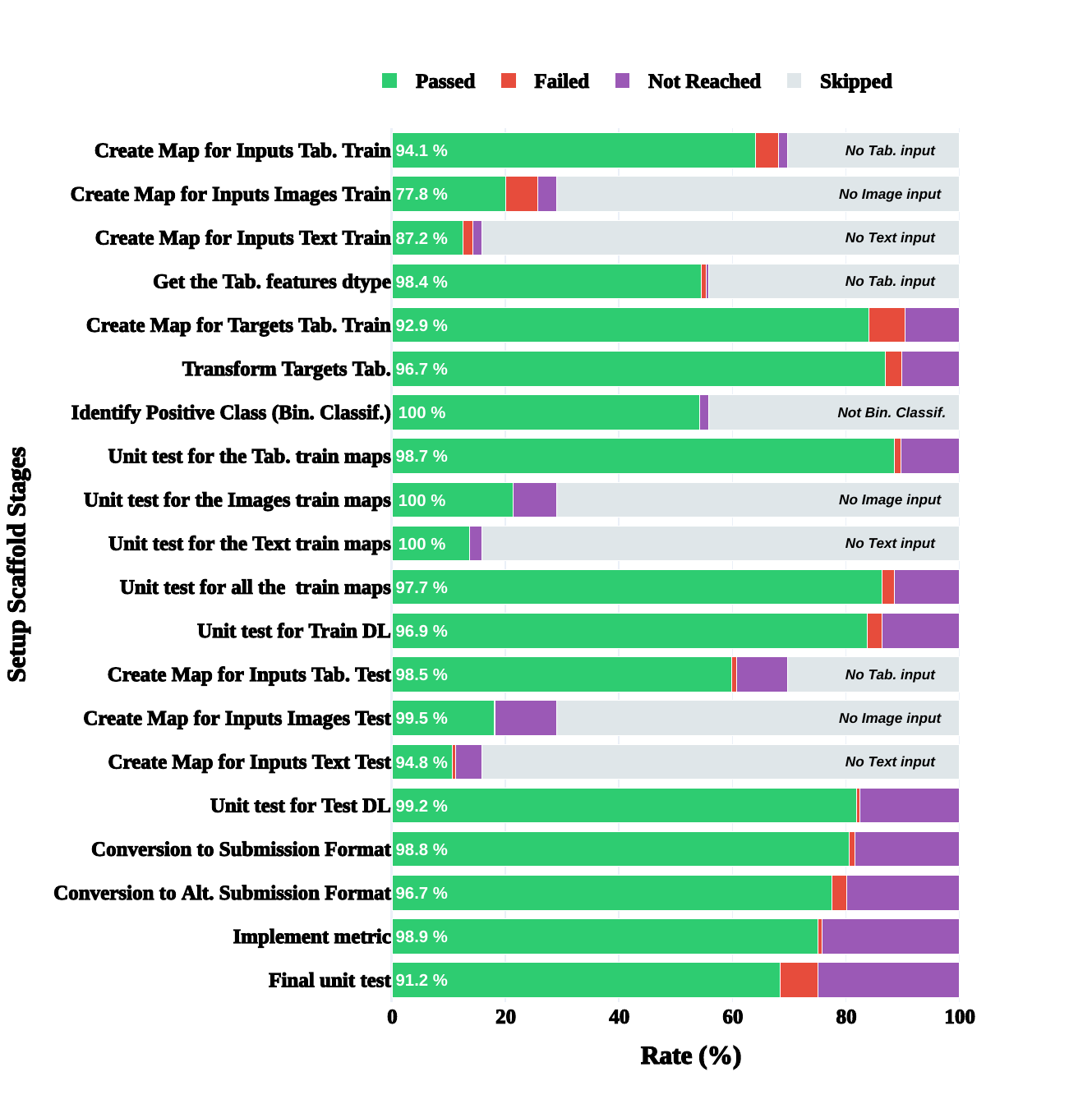}
\caption{\textbf{Success Rate per Setup Stage.} On this percentage stacked bar graph, each row represents a stage in the process of setting up a competition. Its corresponding bar represents the percentage of times this stage is a success, or is failed, or is not reached, or is skipped, averaged over all competitions.}
\label{fig:success-rate-per-setup-stage}
\end{figure}

We show in Figure~\ref{fig:success-rate-per-setup-stage} the success rate of each stage of the setup pipeline  for the different types of tasks.
For each stage, we track if it was a success (i.e. associated unit test passed) or a failure.
We also track if that stage was not reached, caused by the failure of a preceding stage.
Finally we also track if that stage is skipped as not all stages are required for each competition.
As described in Section~\ref{subsec:setup}, the stages to complete depends on the nature of the task and notably of the modalities of the input data.
Figure~\ref{fig:success-rate-per-setup-stage} shows the reason why a stage is skipped on the right-hand side of each bar.
Finally, the success rate (percentage in the green portion of the bar) shown for each stage is reflective of the number of times this particular stage is successfully done by the agent, if this stage is reached, averaged over all competitions.
This explains why, for example, there is a 77.8\% success rate for the stage "Create Map for Inputs Images Train" with a bar that is larger than the 87.2\% of the stage just below "Create Map for Inputs Text Train".

The takeaway from Figure~\ref{fig:success-rate-per-setup-stage} is that most stages have a high success rate.
There is a weaker point for creating the training image input map.
This can be explained by the often complicated of files structures of Kaggle competitions requiring image inputs.
The relatively smaller percentages for the stages such as "Create Map for Inputs Text Train" or "Unit Test for the Text train maps" can be explained by the fact that there are less competitions in our benchmark that require text inputs.
Nevertheless, across input modalities the overall stage success rate is high for \OurAgent{}.
In particular the overall setup success rate is \textbf{92.5\%}.

\captionsetup{justification=centering}

\section{Details on React-Agent baselines}\label{app:aide-modifications}

This section outlines the modifications made to the AIDE version designed for MLE-Bench, to align it with the specific requirements and conditions of our experiments. 
These changes were necessary to ensure that the ReAct Agent from AIDE could effectively operate within the experimental setup, particularly in relation to open-source models and task-specific execution constraints.

\paragraph{Open-Source Models and Structured Output Generation} A key modification involved adapting AIDE to support open-source models, such as Qwen2.5, that do not natively generate structured output.
Originally, AIDE was designed to work with models that inherently produced structured results, simplifying downstream processing. 
However, open-source models like Qwen2.5 often produce unstructured or loosely structured text.
In response to this, the input prompts given to the models were explicitly updated to request the generation of structured outputs, such as JSON format. 
Additionally, a post-processing step was introduced to parse and transform the raw LLM output into the required structured format. 
When issues arose, such as incomplete or misstructured output (e.g., missing expected keys), we sent a new request to the LLM, adding the formatting error to the prompt, in order to get a new valid output.
If after five retrials the output is still not in the expected format, we stop the runs and restart from scratch, which happened only a few times.

\paragraph{Prompt Update}
The prompt to the LLM was updated to include detailed information about the Kaggle competition, such as an overview of the competition and a description of the data. Additionally, the prompt was enhanced to provide system resource details, including the remaining total time, the total allowed number of iterations, and the number of iterations that had been completed. This update ensured that the LLM had access to comprehensive context for better performance and decision-making during the execution process.

\paragraph{Execution time and Best solutions} 

All computer vision and natural language processing competitions were allocated a runtime budget of four days, while tabular competitions were assigned a two-day budget.
This allocation was designed to provide the system with ample time to generate and refine high-quality solutions. 
For CV and NLP tasks, AIDE was configured with a maximum execution time of 18 hours per node, whereas tabular tasks were limited to 9 hours per node to reflect their generally lower computational requirements. 
During execution, AIDE was permitted to perform up to 5,000 improvement iterations to iteratively enhance its solutions. Additionally, the system was allowed a maximum debug depth of 3 from an initial node, with a 50\% probability of selecting a given node for debugging. 
For all tasks, AIDE retained the top four solutions based on the competition-specific evaluation metric.

\paragraph{Kaggle for Real-Time Evaluation} AIDE was integrated with Kaggle to facilitate real-time evaluation and validation of its generated submissions. 
Once the best four submissions were identified, they were submitted to Kaggle for public and private scoring. 
The resulting Kaggle scores were then used to assess the performance of each submission, providing an external benchmark for comparison and ensuring that the system’s outputs were evaluated against established competition standards.

\paragraph{RAG Database setup}
The database for the ReAct (Qwen) + RAG baseline was carefully constructed by collecting 24 Kaggle competitions that began in or after 2021 and were not included in our benchmark set. 
The selected competitions span a balanced mix of domains, including 7 computer vision (CV), 7 natural language processing (NLP), 7 tabular data challenges, and 3 multimodal tasks, ensuring diverse coverage of data science problem types. Table~\ref{tab:rag_competitions_list} lists all the selected competitions, along with the number of notebook summaries and technical reports used to build the RAG database, as well as the corresponding year of each competition.
\begin{table}
    \centering
    \resizebox{\textwidth}{!}{%
    \begin{tabular}{llccc}\toprule
          ID&Competition Name&  Year& \# Tech. Reports & \# Notebooks\\\midrule
          1& \href{https://www.kaggle.com/c/UBC-OCEAN} {UBC-OCEAN}&  2024&10&10\\
          2& \href{https://www.kaggle.com/c/womens-march-mania-2022}{womens-march-mania-2022}&  2022&7&10\\
          3& \href{https://www.kaggle.com/c/predict-student-performance-from-game-play}{predict-student-performance-from-game-play}&  2023&10&10\\
  4& \href{https://www.kaggle.com/c/novozymes-enzyme-stability-prediction}{novozymes-enzyme-stability-prediction}&2023&7&10\\
  5& \href{https://www.kaggle.com/c/czii-cryo-et-object-identification}{czii-cryo-et-object-identification}& 2025& 10&10\\
  6& \href{https://www.kaggle.com/c/jigsaw-toxic-severity-rating}{jigsaw-toxic-severity-rating}& 2022& 10&10\\
  7& \href{https://www.kaggle.com/c/feedback-prize-english-language-learning}{feedback-prize-english-language-learning}& 2022& 10&10\\
  8& \href{https://www.kaggle.com/c/icr-identify-age-related-conditions}{icr-identify-age-related-conditions}& 2023& 10&10\\
  9& \href{https://www.kaggle.com/c/happy-whale-and-dolphin}{happy-whale-and-dolphin}& 2022& 10&10\\
  10& \href{https://www.kaggle.com/c/commonlitreadabilityprize}{commonlitreadabilityprize}& 2021& 10&10\\
  11& \href{https://www.kaggle.com/c/amex-default-prediction}{amex-default-prediction}& 2022& 10&10\\
  12& \href{https://www.kaggle.com/c/rsna-miccai-brain-tumor-radiogenomic-classification}{rsna-miccai-brain-tumor-radiogenomic-classification}& 2021& 10&10\\
  13& \href{https://www.kaggle.com/c/rsna-2023-abdominal-trauma-detection}{rsna-2023-abdominal-trauma-detection}& 2023& 10&10\\
  14& \href{https://www.kaggle.com/c/commonlit-evaluate-student-summaries}{commonlit-evaluate-student-summaries}& 2023& 10&10\\
  15& \href{https://www.kaggle.com/c/feedback-prize-effectiveness} {feedback-prize-effectiveness}& 2022& 10&10\\
  16& \href{https://www.kaggle.com/c/predict-energy-behavior-of-prosumers}{predict-energy-behavior-of-prosumers}& 2024& 10&10\\
  17& \href{https://www.kaggle.com/c/mens-march-mania-2022}{mens-march-mania-2022}& 2022& 6&10\\
  18& \href{https://www.kaggle.com/c/petfinder-pawpularity-score}{petfinder-pawpularity-score}& 2022& 10&10\\
  19& \href{https://www.kaggle.com/c/cassava-leaf-disease-classification}{cassava-leaf-disease-classification}& 2021& 10&10\\
  20& \href{https://www.kaggle.com/c/llm-detect-ai-generated-text}{llm-detect-ai-generated-text}& 2024& 10&10\\
  21& \href{https://www.kaggle.com/c/home-credit-credit-risk-model-stability}{home-credit-credit-risk-model-stability}& 2024& 2&10\\
  22& \href{https://www.kaggle.com/c/rsna-breast-cancer-detection}{rsna-breast-cancer-detection}& 2023& 10&10\\
  23& \href{https://www.kaggle.com/c/learning-agency-lab-automated-essay-scoring-2}{learning-agency-lab-automated-essay-scoring-2}& 2024& 8&10\\
  24& \href{https://www.kaggle.com/c/hpa-single-cell-image-classification}{hpa-single-cell-image-classification}&  2021&10&10\\ \bottomrule
    \end{tabular}
    }
    \caption{Overview of Inputs for RAG Database: Selected competitions, Year of the competitions, Number of technical reports and Number of Notebook summaries}.
    \label{tab:rag_competitions_list}

\end{table}

\section{Algorithmic Overview of TabPFN Fine-Tuning}

Algorithms \ref{pseudocode:kmeans}, \ref{pseudocode:tabpfn_inference}, and \ref{pseudocode:tabpfn_finetuning} present the pseudocode for dataset sampling using KMeans clustering, prediction using TabPFN, and TabPFN fine-tuning, respectively. It is important to note that the input datasets for all TabPFN variants, including the baseline and fine-tuned models, are generated by the setup pipeline of \OurAgent{}. 
\begin{algorithm}[h!]
\caption{KMeans Clustering-Based Sampling}\label{pseudocode:kmeans}
\begin{algorithmic}[1]
\Require \texttt{ctx\_length}, \texttt{train\_data}, \texttt{test\_data}

\State $(\texttt{train\_data}, \texttt{test\_data}) \gets \texttt{Preprocess(train\_data, test\_data)}$
\State $\texttt{kmeans} \gets \texttt{ApplyKMeansClustering}(\texttt{test\_data})$
\State $(\texttt{test\_data}, \texttt{train\_data}) \gets \texttt{AssignClusterLabels}(\texttt{kmeans})$
\State $\texttt{train\_data\_subset} \gets \texttt{SampleTrainData}(\texttt{train\_data}, \texttt{ctx\_length})$
\State \Return \texttt{train\_data\_subset}
\end{algorithmic}
\end{algorithm}

\begin{algorithm}
\caption{TabPFN-Based Prediction}\label{pseudocode:tabpfn_inference}
\begin{algorithmic}[1]
\Require \texttt{ctx\_len}, \texttt{train}, \texttt{test}, \texttt{is\_regression}
\State \texttt{test\_batches} $\gets$ \texttt{CreateBatches(test, ctx\_len)}
\State \texttt{all\_predictions} $\gets$ [ ]
\ForAll{\texttt{test\_batch} in \texttt{test\_batches}}
    \If{\texttt{len(train)} $>$ \texttt{ctx\_len}}
        \State \texttt{sampled\_ctx} $\gets$ \texttt{KMeansSampling(ctx\_len, train, test\_batch)}
    \Else
        \State \texttt{sampled\_ctx} $\gets$ \texttt{train}
    \EndIf
    \State \texttt{x\_ctx} $\gets$ \texttt{getFeatures(sampled\_ctx)}
    \State \texttt{y\_ctx} $\gets$ \texttt{getLabels(sampled\_ctx)}
    \State \texttt{x\_target} $\gets$ \texttt{getFeatures(test\_batch)}
    \State \texttt{predictions} $\gets$ \texttt{TabPFN(x\_ctx, y\_ctx, x\_target, is\_regression)}
    \State \texttt{all\_predictions.append(predictions)}
\EndFor
\State \Return \texttt{all\_predictions}
\end{algorithmic}
\end{algorithm}

\begin{algorithm}
\caption{TabPFN Fine-Tuning}\label{pseudocode:tabpfn_finetuning}
\begin{algorithmic}[1]
\Require \texttt{ctx\_len}, \texttt{org\_train}, \texttt{org\_test}, \texttt{is\_regression}
\State \texttt{train\_data},\texttt{val\_data}, \texttt{test\_data} $\gets$ \texttt{SplitDataset(org\_train)}
\State \texttt{val\_data\_subsets} $\gets$ \texttt{CreateBatches(val\_data, ctx\_len)}
\State \texttt{train\_data\_subsets} $\gets$ [ ]
\ForAll{\texttt{v} in \texttt{val\_data\_subsets}}
    \State \texttt{sample} $\gets$ \texttt{KMeansSampling(ctx\_len, train\_data, v)}
    \State \texttt{train\_data\_subsets.append(sample)}
\EndFor
\For{$i = 1$ \textbf{to} \texttt{len(train\_data\_subsets)}}
    \State \texttt{train\_subset} $\gets$ \texttt{train\_data\_subsets[i]}
    \State \texttt{val\_subset} $\gets$ \texttt{val\_data\_subsets[i]}
    \State \texttt{model} $\gets$ \texttt{LoadModel(model\_path, is\_regression)}
    \State \texttt{initial\_loss} $\gets$ \texttt{EvaluateModel(model, train\_subset, test\_data)}
    \ForAll{\texttt{data} in \texttt{TrainDataLoader(train\_subset)}}
        \State \texttt{model} $\gets$ \texttt{FineTune(model, data, val\_subset)}
        \State \texttt{validation\_loss} $\gets$ \texttt{EvaluateModel(model, data, test\_data)}
    \EndFor
\EndFor
\State \texttt{all\_preds} $\gets$ \texttt{Pred(model\_path, org\_train, org\_test, is\_regression)}
\State \Return \texttt{all\_preds}
\end{algorithmic}
\end{algorithm}

\section{Examples of Prompts and Responses}
Below are some examples of prompts and answers from the LLM that our agent can query.
Even though they are presented hereafter in a chat format, note that the LLM does not keep a memory of the chat as generally thought of in recent chat-based systems. 
The prompts are long because at each interaction we add all the necessary information about the specific step the agent is performing, knowing that the queried LLM has no context or records of previous interactions. 
The prompts are therefore shortened for the purpose of display.
The paths in the code snippets are taken from our runs so we also anonymised them e.g. \texttt{/PATH/TO/RAW/DATA/...} or \texttt{/PATH/TO/WORKSPACE/...}

\paragraph{A Note on Prompts} The agent is performing a complex task broken up into a sequence of smaller subtasks. 
These include summarising information and also past responses. 
These summaries are the main bulk of the text present in the prompts. They include summarisation of the competition description, the raw data and all past steps the agent has done if relevant.
The other big part of the prompts are views. 
These are usually a raw text output from a terminal command such as ``\texttt{\$ls}''or python ``\texttt{print()}'' statement, e.g. the view of the raw data directory or a head view of some dataframes along with their column types, etc. Finally, some prompts can contain examples of similar situations seen in similar competitions that the agent can pick itself and reuse as examples through RAG. 
The rest is a set of general instructions manually written. 

\subsection{Setup}\label{Sec:ExtendedPrompts}
The example below is taken from the setup of the competition \texttt{leaf-classification} which has tabular and image inputs.
We show examples of prompts and responses from the LLM in the particular case where the agent creates a plan for coding and then fill a code template based on the plan. 
They contain all the necessary information about the task and instructions about the specific step the agent is currently doing for a successful answer. 
In the particular case below, the LLM is prompted to generate a plan and then the code to create the training map for the image modality, having already done the tabular modality.
\begin{userbox}

\# Data Description\\
Dataset Description\\
The dataset consists approximately 1,584 images of leaf specimens (16 samples each of 99 species) which have been converted to binary black leaves against white backgrounds. 
Three sets of features are also provided per image: a shape contiguous descriptor, an interior texture histogram, and a ﬁne-scale margin histogram. 
For each feature, a 64-attribute vector is given per leaf sample.
Note that of the original 100 species, we have eliminated one on account of incomplete associated data in the original dataset.\\

File descriptions\\
train.csv - the training set\\
test.csv - the test set\\
sample\_submission.csv - a sample submission file in the correct format\\
images - the image files (each image is named with its corresponding id)\\

Data fields\\
id - an anonymous id unique to an image\\
margin\_1, margin\_2, margin\_3, ..., margin\_64 - each of the 64 attribute vectors for the margin feature\\
shape\_1, shape\_2, shape\_3, ..., shape\_64 - each of the 64 attribute vectors for the shape feature\\
texture\_1, texture\_2, texture\_3, ..., texture\_64 - each of the 64 attribute vectors for the texture feature\\

\#\#\# View of the top-level directory, these are the data you can use to solve this task:\\
\begin{lstlisting}[style=bashstyle]
/PATH/TO/RAW/DATA
+-leaf-classification/
  |
  +-images/
  | +-226.jpg
  | +-342.jpg
  | +-1534.jpg
  | +-... <additional files>
  |
  +-train.csv
  |
  +-test.csv
\end{lstlisting}

\# Modalities Description
For the leaf classification task, the input modalities needed to solve the task are as follows:\\

\#\#\# Input Modalities\\
- **Tabular Input Map:**\\
  - **YES** because the dataset includes tabular data with pre-extracted features (margin, shape, and texture).\\
  - **Related Files:**\\
    - `train.csv`\\
    - `test.csv`\\

- **Image Input Map:**\\
  - **YES** as there are images of leaf specimens that can be used for additional feature extraction or as an alternative input.\\
  - **Related Files:**\\
    - `images/` (directory containing the leaf images)\\

- **Text Input Map:**\\
  - **NO** because there is no presence of text modality in this task.\\

\#\#\# Summary\\

- **Tabular Input Map:** YES -> Related Files: `train.csv`, `test.csv`\\
- **Image Input Map:** YES -> Related Files: `images/\\
- **Text Input Map:** NO\\

This structured approach ensures that all relevant input modalities are considered and utilized for the leaf classification task.\\

\#\#\# These are some of the tables you have asked information about:\\
\#\#\#\# View of table `/PATH/TO/RAW/DATA/leaf-classification/train.csv`:
\begin{lstlisting}[style=bashstyle]
/PATH/TO/RAW/DATA/leaf-classification/train.csv
...
<Skipped for prompt display purposes in paper appendix>
...
[2 rows x 194 columns]
\end{lstlisting}

\begin{lstlisting}[style=bashstyle]
<class 'pandas.core.frame.DataFrame'>
RangeIndex: 990 entries, 0 to 989
Columns: 194 entries, id to texture64
dtypes: float64(192), int64(1), object(1)
memory usage: 1.5+ MB

- column species contains strings with values in ['Acer_Opalus', 'Pterocarya_Stenoptera', 'Quercus_Hartwissiana', 'Tilia_Tomentosa', 'Quercus_Variabilis', 'Magnolia_Salicifolia', 'Quercus_Canariensis', 'Quercus_Rubra', 'Quercus_Brantii', 'Salix_Fragilis', 'Zelkova_Serrata', 'Betula_Austrosinensis', 'Quercus_Pontica', 'Quercus_Afares', 'Quercus_Coccifera', 'Fagus_Sylvatica', 'Phildelphus', 'Acer_Palmatum', 'Quercus_Pubescens', 'Populus_Adenopoda', 'Quercus_Trojana', 'Alnus_Sieboldiana', 'Quercus_Ilex', 'Arundinaria_Simonii', 'Acer_Platanoids', 'Quercus_Phillyraeoides', 'Cornus_Chinensis', 'Liriodendron_Tulipifera', 'Cytisus_Battandieri', 'Rhododendron_x_Russellianum', 'Alnus_Rubra', 'Eucalyptus_Glaucescens', 'Cercis_Siliquastrum', 'Cotinus_Coggygria', 'Celtis_Koraiensis', 'Quercus_Crassifolia', 'Quercus_Kewensis', 'Cornus_Controversa', 'Quercus_Pyrenaica', 'Callicarpa_Bodinieri', 'Quercus_Alnifolia', 'Acer_Saccharinum', 'Prunus_X_Shmittii', 'Prunus_Avium', 'Quercus_Greggii', 'Quercus_Suber', 'Quercus_Dolicholepis', 'Ilex_Cornuta', 'Tilia_Oliveri', 'Quercus_Semecarpifolia', 'Quercus_Texana', 'Ginkgo_Biloba', 'Liquidambar_Styraciflua', 'Quercus_Phellos', 'Quercus_Palustris', 'Alnus_Maximowiczii', 'Quercus_Agrifolia', 'Acer_Pictum', 'Acer_Rufinerve', 'Lithocarpus_Cleistocarpus', 'Viburnum_x_Rhytidophylloides', 'Ilex_Aquifolium', 'Acer_Circinatum', 'Quercus_Coccinea', 'Quercus_Cerris', 'Quercus_Chrysolepis', 'Eucalyptus_Neglecta', 'Tilia_Platyphyllos', 'Alnus_Cordata', 'Populus_Nigra', 'Acer_Capillipes', 'Magnolia_Heptapeta', 'Acer_Mono', 'Cornus_Macrophylla', 'Crataegus_Monogyna', 'Quercus_x_Turneri', 'Quercus_Castaneifolia', 'Lithocarpus_Edulis', 'Populus_Grandidentata', 'Acer_Rubrum', 'Quercus_Imbricaria', 'Eucalyptus_Urnigera', 'Quercus_Crassipes', 'Viburnum_Tinus', 'Morus_Nigra', 'Quercus_Vulcanica', 'Alnus_Viridis', 'Betula_Pendula', 'Olea_Europaea', 'Quercus_Ellipsoidalis', 'Quercus_x_Hispanica', 'Quercus_Shumardii', 'Quercus_Rhysophylla', 'Castanea_Sativa', 'Ulmus_Bergmanniana', 'Quercus_Nigra', 'Salix_Intergra', 'Quercus_Infectoria_sub', 'Sorbus_Aria']
\end{lstlisting}

Your goal is to create a map for the training image inputs.\\
The first column is the "id" containing the unique ID of each data point, you need to create id so it can be used later to match each data point to their target. 
For example, the filename name can be use as unique identifier. The subsequent columns are the paths to the training input images for `train\_img\_input\_map.csv`. Do not include target columns in this table.\\

Now, write a **short** (no more than 3 steps) plan of how you would do that. If there is an error in a previous attempt, explain in **one** sentence why this error happened and how to fix it. Do not write code yet, simply the plan.
\end{userbox}

\begin{flushright}
\begin{assistantbox}
\#\#\# Plan to Create the Training Image Input Map\\

1. **Extract Image IDs and Paths:**\\
   - Read the `train.csv` file to extract the `id` column, which contains the unique IDs for each image.\\
   - Construct the image file paths using the `id` values and the directory path to the images (e.g., `images/226.jpg`).\\

2. **Create the Image Input Map CSV:**\\
   - Create a new DataFrame with two columns: `id` and `image\_path`.\\
   - Populate the `image\_path` column with the constructed file paths.\\

3. **Save the Image Input Map CSV:**\\
   - Save the DataFrame to a CSV file named `train\_img\_input\_map.csv`.\\

\#\#\# Error Handling\\
- If the `id` values in `train.csv` do not match the filenames in the `images/` directory, ensure that the filenames are correctly formatted and match the `id` values exactly.
\end{assistantbox}
\end{flushright}

\begin{userbox}
\begin{lstlisting}[style=bashstyle]
<Skipped first part of prompt repeating the previous prompt for display purposes in paper appendix>
\end{lstlisting}

\#\#\# Create `train\_img\_input\_map.csv`\\
Your goal is to create a map for the training image inputs.
The first column of that map must be named "id" and must contain the unique ID of each data point.
If this column exists already in the data, make sure to rename it to "id", otherwise create it.
The subsequent columns must be the paths to the training input images.
Do not include target columns in the table you are creating, this will be done later in another table.\\

\#\#\# This is the code you helped write earlier to create the train tabular inputs:\\
\begin{lstlisting}[style=pythonstyle]
"""
This script creates the data map for the tabular training inputs, i.e. `train_tab_input_map.csv`.
This map is a `pd.DataFrame` object populated with respect to the data description and the data present in the raw data directory.
The first column of the created map must be the "id" of each test data point.
The subsequent columns of `train_tab_input_map.csv` are the tabular training input features, if any.

>>> # Example
>>> train_tab_input_map = pd.DataFrame(columns=["id", "tab_input_feature1", "tab_input_feature2", ...])

Create train_tab_input_map
"""
# <|fim_begin|>
import os
import pandas as pd

import os
import pandas as pd

# Define the absolute path to the train.csv file
train_csv_path = "/PATH/TO/RAW/DATA/leaf-classification/train.csv"

# Read the train.csv file
train_df = pd.read_csv(train_csv_path)

# Select the relevant columns: id and all feature columns, excluding the species column
train_tab_input_map = train_df.drop(columns=['species'])

# Save the new DataFrame to a CSV file
train_tab_input_map.to_csv("./train_tab_input_map.csv", index=False)
print("`train_tab_input_map.csv` created and saved.")


# <|fim_end|>
# save
train_tab_input_map.to_csv("./train_tab_input_map.csv", index=False)
print("`train_tab_input_map.csv` created and saved.")
\end{lstlisting}

The script reads the train.csv file, selects the id and feature columns (excluding species), and saves the result to `train\_tab\_input\_map.csv`.\\

\#\#\# Complete the code\\
Complete the following code template in order to create the data map `train\_img\_input\_map.csv`:
\begin{lstlisting}[style=pythonstyle]
"""
This script creates the data map for the image training inputs, i.e. `train_img_input_map.csv`.
This map is a `pd.DataFrame` object populated with respect to the data description and the data present in the raw data directory.
The first column of the created map must be the "id" of each test data point.
The subsequent columns of `train_img_input_map.csv` are the paths to the training input images, if any.

>>> # Example
>>> train_img_input_map = pd.DataFrame(columns=["id", "img_input1", "img_input2", ...])

Create train_img_input_map
"""
# <|fim_begin|>
import os
import pandas as pd
# <|fim_hole|>

# <|fim_end|>
# save
train_img_input_map.to_csv("./train_img_input_map.csv", index=False)
print("`train_img_input_map.csv` created and saved.")
\end{lstlisting}

Pay EXTREME attention to the path where you load from. You should always provide absolute path to avoid `FileNotFound` error.
Do not re-set index of the dataframe to create a new id column, instead use the columns present in the dataframe already.
Create a new index column only when absolutely necessary.
Finally, the step to save the final table is already written in the template - do not rewrite this part.
The map is saved in the current directory on purpose.\\

Tip: IF you are using the image file name as id, make sure they are converted to strings.

\#\#\# Plan:\\
\#\#\# Plan to Create the Training Image Input Map\\

1. **Extract Image IDs and Paths:**\\
   - Read the `train.csv` file to extract the `id` column, which contains the unique IDs for each image.\\
   - Construct the image file paths using the `id` values and the directory path to the images (e.g., `images/226.jpg`).\\

2. **Create the Image Input Map CSV:**\\
   - Create a new DataFrame with two columns: `id` and `image\_path`.\\
   - Populate the `image\_path` column with the constructed file paths.\\

3. **Save the Image Input Map CSV:**\\
   - Save the DataFrame to a CSV file named `train\_img\_input\_map.csv`.\\

\#\#\# Error Handling\\
- If the `id` values in `train.csv` do not match the filenames in the `images/` directory, ensure that the filenames are correctly formatted and match the `id` values exactly.
Your response should be the code to complete the above template, in the format:\\
```python\\
code\\
```\\
\end{userbox}

\begin{flushright}
\begin{assistantbox}
\begin{lstlisting}[style=pythonstyle]
    """
This script creates the data map for the image training inputs, i.e. `train_img_input_map.csv`.
This map is a `pd.DataFrame` object populated with respect to the data description and the data present in the raw data directory.
The first column of the created map must be the "id" of each test data point.
The subsequent columns of `train_img_input_map.csv` are the paths to the training input images, if any.

>>> # Example
>>> train_img_input_map = pd.DataFrame(columns=["id", "img_input1", "img_input2", ...])

Create train_img_input_map
"""
# <|fim_begin|>
import os
import pandas as pd

# Define the absolute path to the train.csv file
train_csv_path = "/PATH/TO/RAW/DATA/leaf-classification/train.csv"
images_dir = "/PATH/TO/RAW/DATA/leaf-classification/images/"

# Read the train.csv file
train_df = pd.read_csv(train_csv_path)

# Ensure the id column is a string
train_df['id'] = train_df['id'].astype(str)

# Construct the image file paths
train_df['image_path'] = train_df['id'].apply(lambda x: os.path.join(images_dir, f"{x}.jpg"))

# Create the image input map DataFrame
train_img_input_map = train_df[['id', 'image_path']]

# <|fim_end|>
# save
train_img_input_map.to_csv("./train_img_input_map.csv", index=False)
print("`train_img_input_map.csv` created and saved.")
\end{lstlisting}
\end{assistantbox}
\end{flushright}

\subsection{Setup with RAG Example}
In the example below we show how an example of a prompt containing an example retrieve looks like. 
This is during the setup of the competition \texttt{ml-olympiad-landscape-image-classification} and in particular when generating the code for the training map for the image modality.
The previous attempt had en error in the unit test which is shown in the prompt as the agent is asked to retry.
The RAG engine retrieves some similar example from a competition already setup. In that other competition, namely \texttt{kaggle-pog-series-s01e03}, the agent had a similar issue and managed to fix it.
These steps are retrieved and displayed in the current prompt so that the agent can take inspiration from that similar error being fixed and try to adapt a fix to the current issue it is facing.

\begin{userbox}
\#\#\# Data Description\\
Dataset Description\\
The data contains masked images with random patches. These random patches can cover anything between 20\% to 60\% of the image. Your goal is to correctly predict the classes of the images from the "test" directory. All images are 256x256 having 3 channels.\\

Files\\
train.csv - the training set\\
test.csv - the test set\\
sample.csv - a sample submission file in the correct format\\
Columns\\
image - Path of the image relative to directory. Eg. In train.csv, mountain/12345.jpg is relative to the "train" directory, whereas in test.csv, 12345.jpg is relative to the "test" directory.\\
label - True label of the image (only in train.csv) .\\
Submission files should have these column headers, but not the column numbers. Use df.to\_csv(filepath, index=False) to save without the column numbers.\\

\#\#\# View of the top-level directory, these are the data you can use to solve this task:\\
\begin{lstlisting}[style=bashstyle]
/PATH/TO/RAW/DATA
+-ml-olympiad-landscape-image-classification/
  |
  +-train/
  | |
  | +-forest/
  | | +-64935341.jpg
  | | +-89262487.jpg
  | | +-16354101.jpg
  | | +-... <additional files>
  | |
  | +-buildings/
  | | +-15220800.jpg
  | | +-30615322.jpg
  | | +-48061322.jpg
  | | +-... <additional files>
  | |
  | +-glacier/
  | | +-10983747.jpg
  | | +-72741892.jpg
  | | +-11747834.jpg
  | | +-... <additional files>
  | |
  | +-street/
  | | +-16518706.jpg
  | | +-88133361.jpg
  | | +-17503215.jpg
  | | +-... <additional files>
  | |
  | +-mountain/
  | | +-11476151.jpg
  | | +-14340600.jpg
  | | +-72470093.jpg
  | | +-... <additional files>
  | +-... <additional folders>
  |
  +-test/
  | +-62449337.jpg
  | +-18432742.jpg
  | +-84761017.jpg
  | +-... <additional files>
  |
  +-test.csv
  |
  +-train.csv
\end{lstlisting}

\#\#\# Modalities Description\\
\#\#\# Input Modalities for the Task\\

- **tabular input map**: NO because there is no presence of tabular data (other than the image paths and labels, which are not used as features in the model).\\
- **image input map**: YES as there are images in this task -> related files: \\
- `/PATH/TO/RAW/DATA/ml-olympiad-landscape-image-classification/train/` (training images)\\
- `/PATH/TO/RAW/DATA/ml-olympiad-landscape-image-classification/test/` (test images)\\
- **text input map**: NO because there is no presence of text modality in this task.\\

\#\#\# Summary\\

- **Tabular Input**: NO\\
- **Image Input**: YES (related files: `train/`, `test/`)\\
- **Text Input**: NO\\

\#\#\# These are some tips to avoid common mistakes in the code:\\

- To avoid the error related to non-unique filenames, append the class name to the filenames to ensure they are unique. For example, change '12345678.jpg' in the 'forest' directory to 'forest\_12345678.jpg'. This can be done by concatenating the class name with the filename, separated by an underscore or another delimiter.\\
- To avoid the 'File not found' error, ensure that all file paths listed in the 'train.csv' and 'test.csv' files are correct and that the corresponding files exist in the dataset. You can write a script to verify the existence of all files listed in these CSV files.\\

\#\#\# Create `train\_img\_input\_map.csv`\\
Your goal is to create a map for the training image inputs.
The first column of that map must be named "id" and must contain the unique ID of each data point.
If this column exists already in the data, make sure to rename it to "id", otherwise create it.
The subsequent columns must be the paths to the training input images.
Do not include target columns in the table you are creating, this will be done later in another table.\\

\#\#\# This was the resulting code after your last attempt:\\
\begin{lstlisting}[style=pythonstyle]
"""
This script creates the data map for the image training inputs, i.e. `train_img_input_map.csv`.
This map is a `pd.DataFrame` object populated with respect to the data description and the data present in the raw data directory.
The first column of the created map must be the "id" of each test data point.
The subsequent columns of `train_img_input_map.csv` are the paths to the training input images, if any.

>>> # Example
>>> train_img_input_map = pd.DataFrame(columns=["id", "img_input1", "img_input2", ...])

Create train_img_input_map
"""
# <|fim_begin|>
import os
import pandas as pd

"""
This script creates the data map for the image training inputs, i.e. `train_img_input_map.csv`.
This map is a `pd.DataFrame` object populated with respect to the data description and the data present in the raw data directory.
The first column of the created map must be the "id" of each test data point.
The subsequent columns of `train_img_input_map.csv` are the paths to the training input images, if any.

>>> # Example
>>> train_img_input_map = pd.DataFrame(columns=["id", "img_input1", "img_input2", ...])

Create train_img_input_map
"""
# <|fim_begin|>
import os
import pandas as pd

# Load the training data
train_df = pd.read_csv("/PATH/TO/RAW/DATA/ml-olympiad-landscape-image-classification/train.csv")

# Extract the image filenames and use them as unique IDs
train_df['id'] = train_df.apply(lambda row: f"{row['label']}_{row['image'].split('/')[-1]}", axis=1)

# Ensure that the filenames are unique
if train_df['id'].nunique() != len(train_df):
    raise ValueError("Filenames are not unique. Please append a unique identifier to duplicate filenames.")

# Create the input map DataFrame
train_img_input_map = train_df[['id', 'image']].rename(columns={'image': 'img_input1'})

# <|fim_end|>
# save
train_img_input_map.to_csv("./train_img_input_map.csv", index=False)
print("`train_img_input_map.csv` created and saved.")


# <|fim_end|>
# save
train_img_input_map.to_csv("./train_img_input_map.csv", index=False)
print("`train_img_input_map.csv` created and saved.")
\end{lstlisting}

\#\#\# This was the output:\\
\begin{lstlisting}[style=bashstyle]
`train_img_input_map.csv` created and saved.
`train_img_input_map.csv` created and saved.
\end{lstlisting}

\#\#\# The error in the unit test was:\\
\begin{lstlisting}[style=bashstyle]
multiprocessing.pool.RemoteTraceback: 
"""
Traceback (most recent call last):
  File "/home/jdoran/uv_python/cpython-3.10.15-linux-x86_64-gnu/lib/python3.10/multiprocessing/pool.py", line 125, in worker
    result = (True, func(*args, **kwds))
  File "/home/jdoran/uv_python/cpython-3.10.15-linux-x86_64-gnu/lib/python3.10/multiprocessing/pool.py", line 48, in mapstar
    return list(map(*args))
  File "/PATH/TO/WORKSPACE/ml-olympiad-landscape-image-classification/seed_0/unit_test_map_img_input_train.py", line 45, in check_file_exists
    raise FileNotFoundError(f"File not found: {file_path}")
FileNotFoundError: File not found: glacier/59804579.jpg
"""

The above exception was the direct cause of the following exception:

Traceback (most recent call last):
  File "/PATH/TO/WORKSPACE/ml-olympiad-landscape-image-classification/seed_0/unit_test_map_img_input_train.py", line 65, in <module>
    check_files(files_to_check)
  File "/PATH/TO/WORKSPACE/ml-olympiad-landscape-image-classification/seed_0/unit_test_map_img_input_train.py", line 58, in check_files
    results = pool.map(check_file_exists, file_list)
  File "/home/jdoran/uv_python/cpython-3.10.15-linux-x86_64-gnu/lib/python3.10/multiprocessing/pool.py", line 367, in map
    return self._map_async(func, iterable, mapstar, chunksize).get()
  File "/home/jdoran/uv_python/cpython-3.10.15-linux-x86_64-gnu/lib/python3.10/multiprocessing/pool.py", line 774, in get
    raise self._value
FileNotFoundError: File not found: glacier/59804579.jpg
\end{lstlisting}

You should fix this error next.\\

\#\#\# An example of a similar error being fixed:\\
Failing code:\\
\begin{lstlisting}[style=pythonstyle]
"""
This script creates the data map for the image training inputs, i.e. `train_img_input_map.csv`.
This map is a `pd.DataFrame` object populated with respect to the data description and the data present in the raw data directory.
The first column of the created map must be the "id" of each test data point.
The subsequent columns of `train_img_input_map.csv` are the paths to the training input images, if any.

>>> # Example
>>> train_img_input_map = pd.DataFrame(columns=["id", "img_input1", "img_input2", ...])

Create train_img_input_map
"""
# <|fim_begin|>
import os
import pandas as pd

"""
This script creates the data map for the image training inputs, i.e. `train_img_input_map.csv`.
This map is a `pd.DataFrame` object populated with respect to the data description and the data present in the raw data directory.
The first column of the created map must be the "id" of each test data point.
The subsequent columns of `train_img_input_map.csv` are the paths to the training input images, if any.

>>> # Example
>>> train_img_input_map = pd.DataFrame(columns=["id", "img_input1", "img_input2", ...])

Create train_img_input_map
"""
# <|fim_begin|>
import os
import pandas as pd

# Define the absolute path to the train.csv file
train_csv_path = "/PATH/TO/RAW/DATA/kaggle-pog-series-s01e03/train.csv"

# Load the train.csv file into a DataFrame
train_df = pd.read_csv(train_csv_path)

# Generate a unique ID by concatenating seed_id and view
train_df['id'] = train_df['seed_id'].astype(str) + '_' + train_df['view']

# Check for duplicates in the generated ID
if train_df['id'].duplicated().any():
    raise ValueError("There are duplicate IDs in the generated unique ID column.")

# Construct the full path to each image
base_image_dir = "/PATH/TO/RAW/DATA/kaggle-pog-series-s01e03/train/"
train_df['img_input1'] = base_image_dir + train_df['image']

# Select the relevant columns: 'id' and 'img_input1'
train_img_input_map = train_df[['id', 'img_input1']]

# <|fim_begin|>
# save
train_img_input_map.to_csv("./train_img_input_map.csv", index=False)
print("`train_img_input_map.csv` created and saved.")


# <|fim_end|>
# save
train_img_input_map.to_csv("./train_img_input_map.csv", index=False)
print("`train_img_input_map.csv` created and saved.")
\end{lstlisting}

Failing code error:\\
\begin{lstlisting}[style=bashstyle]
multiprocessing.pool.RemoteTraceback: 
"""
Traceback (most recent call last):
  File "/home/jdoran/uv_python/cpython-3.10.15-linux-x86_64-gnu/lib/python3.10/multiprocessing/pool.py", line 125, in worker
    result = (True, func(*args, **kwds))
  File "/home/jdoran/uv_python/cpython-3.10.15-linux-x86_64-gnu/lib/python3.10/multiprocessing/pool.py", line 48, in mapstar
    return list(map(*args))
  File "/PATH/TO/WORKSPACE/kaggle-pog-series-s01e03/seed_0/unit_test_map_img_input_train.py", line 45, in check_file_exists
    raise FileNotFoundError(f"File not found: {file_path}")
FileNotFoundError: File not found: /PATH/TO/RAW/DATA/kaggle-pog-series-s01e03/train/train/03511.png
"""

The above exception was the direct cause of the following exception:

Traceback (most recent call last):
  File "/PATH/TO/WORKSPACE/kaggle-pog-series-s01e03/seed_0/unit_test_map_img_input_train.py", line 65, in <module>
    check_files(files_to_check)
  File "/PATH/TO/WORKSPACE/kaggle-pog-series-s01e03/seed_0/unit_test_map_img_input_train.py", line 58, in check_files
    results = pool.map(check_file_exists, file_list)
  File "/home/jdoran/uv_python/cpython-3.10.15-linux-x86_64-gnu/lib/python3.10/multiprocessing/pool.py", line 367, in map
    return self._map_async(func, iterable, mapstar, chunksize).get()
  File "/home/jdoran/uv_python/cpython-3.10.15-linux-x86_64-gnu/lib/python3.10/multiprocessing/pool.py", line 774, in get
    raise self._value
FileNotFoundError: File not found: /PATH/TO/RAW/DATA/kaggle-pog-series-s01e03/train/train/03511.png

\end{lstlisting}

Fixed code:\\
\begin{lstlisting}[style=pythonstyle]
"""
This script creates the data map for the image training inputs, i.e. `train_img_input_map.csv`.
This map is a `pd.DataFrame` object populated with respect to the data description and the data present in the raw data directory.
The first column of the created map must be the "id" of each test data point.
The subsequent columns of `train_img_input_map.csv` are the paths to the training input images, if any.

>>> # Example
>>> train_img_input_map = pd.DataFrame(columns=["id", "img_input1", "img_input2", ...])

Create train_img_input_map
"""
# <|fim_begin|>
import os
import pandas as pd

"""
This script creates the data map for the image training inputs, i.e. `train_img_input_map.csv`.
This map is a `pd.DataFrame` object populated with respect to the data description and the data present in the raw data directory.
The first column of the created map must be the "id" of each test data point.
The subsequent columns of `train_img_input_map.csv` are the paths to the training input images, if any.

>>> # Example
>>> train_img_input_map = pd.DataFrame(columns=["id", "img_input1", "img_input2", ...])

Create train_img_input_map
"""
# <|fim_begin|>
import os
import pandas as pd

# Define the absolute path to the train.csv file
train_csv_path = "/PATH/TO/RAW/DATA/kaggle-pog-series-s01e03/train.csv"

# Load the train.csv file into a DataFrame
train_df = pd.read_csv(train_csv_path)

# Generate a unique ID by concatenating seed_id and view
train_df['id'] = train_df['seed_id'].astype(str) + '_' + train_df['view']

# Check for duplicates in the generated ID
if train_df['id'].duplicated().any():
    raise ValueError("There are duplicate IDs in the generated unique ID column.")

# Construct the full path to each image
base_image_dir = "/PATH/TO/RAW/DATA/kaggle-pog-series-s01e03/train/"
train_df['img_input1'] = base_image_dir + train_df['image'].str.split('/').str[-1]

# Select the relevant columns: 'id' and 'img_input1'
train_img_input_map = train_df[['id', 'img_input1']]

# <|fim_end|>
# save
train_img_input_map.to_csv("./train_img_input_map.csv", index=False)
print("`train_img_input_map.csv` created and saved.")


# <|fim_end|>
# save
train_img_input_map.to_csv("./train_img_input_map.csv", index=False)
print("`train_img_input_map.csv` created and saved.")
\end{lstlisting}

\#\#\# Complete the code\\
Complete the following code template in order to create the data map `train\_img\_input\_map.csv`:\\
\begin{lstlisting}[style=pythonstyle]
"""
This script creates the data map for the image training inputs, i.e. `train_img_input_map.csv`.
This map is a `pd.DataFrame` object populated with respect to the data description and the data present in the raw data directory.
The first column of the created map must be the "id" of each test data point.
The subsequent columns of `train_img_input_map.csv` are the paths to the training input images, if any.

>>> # Example
>>> train_img_input_map = pd.DataFrame(columns=["id", "img_input1", "img_input2", ...])

Create train_img_input_map
"""
# <|fim_begin|>
import os
import pandas as pd
# <|fim_hole|>

# <|fim_end|>
# save
train_img_input_map.to_csv("./train_img_input_map.csv", index=False)
print("`train_img_input_map.csv` created and saved.")
\end{lstlisting}

Pay EXTREME attention to the path where you load from. You should always provide absolute path to avoid `FileNotFound` error.
Do not re-set index of the dataframe to create a new id column, instead use the columns present in the dataframe already.
Create a new index column only when absolutely necessary.
Finally, the step to save the final table is already written in the template - do not rewrite this part.
The map is saved in the current directory on purpose.\\

Tip: IF you are using the image file name as id, make sure they are converted to strings.\\

\#\#\# Plan:\\
1. **Load the Training Data**: Read the `train.csv` file to get the image paths and labels.\\
2. **Create Unique IDs**: Generate a unique ID for each image by concatenating the label and the image filename.\\
3. **Construct the Input Map**: Create a DataFrame with the unique IDs and the corresponding image paths, and save it as `train\_img\_input\_map.csv`.\\

\#\#\# Error Explanation:\\
The error occurred because the file path in the `train.csv` file was incorrect or the file did not exist in the specified directory. To fix it, ensure that the file paths in `train.csv` are correct and that the files exist in the `train/` directory.\\
Your response should be the code to complete the above template, in the format:\\
```python\\
code\\
```\\

\end{userbox}

\begin{assistantbox}
\begin{lstlisting}[style=pythonstyle]
"""
This script creates the data map for the image training inputs, i.e. `train_img_input_map.csv`.
This map is a `pd.DataFrame` object populated with respect to the data description and the data present in the raw data directory.
The first column of the created map must be the "id" of each test data point.
The subsequent columns of `train_img_input_map.csv` are the paths to the training input images, if any.

>>> # Example
>>> train_img_input_map = pd.DataFrame(columns=["id", "img_input1", "img_input2", ...])

Create train_img_input_map
"""
# <|fim_begin|>
import os
import pandas as pd

# Define the absolute path to the train.csv file
train_csv_path = "/PATH/TO/RAW/DATA/ml-olympiad-landscape-image-classification/train.csv"

# Load the train.csv file into a DataFrame
train_df = pd.read_csv(train_csv_path)

# Generate a unique ID by concatenating the label and the image filename
train_df['id'] = train_df.apply(lambda row: f"{row['label']}_{row['image'].split('/')[-1]}", axis=1)

# Check for duplicates in the generated ID
if train_df['id'].duplicated().any():
    raise ValueError("There are duplicate IDs in the generated unique ID column.")

# Construct the full path to each image
base_image_dir = "/PATH/TO/RAW/DATA/ml-olympiad-landscape-image-classification/train/"
train_df['img_input1'] = base_image_dir + train_df['image']

# Select the relevant columns: 'id' and 'img_input1'
train_img_input_map = train_df[['id', 'img_input1']]

# <|fim_end|>
# save
train_img_input_map.to_csv("./train_img_input_map.csv", index=False)
print("`train_img_input_map.csv` created and saved.")
\end{lstlisting}
\end{assistantbox}

\subsection{Solution Generation}
The example below is taken from the solution generation pipeline on the comeptition \texttt{dogs-vs-cats-redux-kernels-edition}. In this particular example, the agent has already done one submission which appears in some prompts. 
The agent is at the start of a new submission, generating the plans and codes for preprocessing the images and the image embedding model.

\begin{userbox}
\#\# Task description\\
\#\#\# Summary of the Task Description\\

**Main Goal:**\\
The goal of this competition is to classify images as either dogs or cats. The competition revisits the original "Dogs vs. Cats" challenge from 2013, but with modern machine learning techniques and tools, such as Kernels (Kaggle's online coding environment) and TensorFlow.\\

**Task Type:**\\
This is a binary classification task where the model must predict the probability that a given image is a dog. The evaluation metric is log loss, which measures the performance of the model by penalizing in correct predictions more severely as the predicted probability diverges from the actual label.\\

**Key Points:**\\
- **Dataset:** The dataset consists of images of dogs and cats.\\
- **Evaluation Metric:** Log loss.\\
- **Submission Format:** A CSV file with the format `id,label`, where `id` is the image ID and `label` is the predicted probability that the image is a dog.\\

\#\# Data description\\
\#\#\# View of the top-level directory /PATH/TO/RAW/DATA/dogs-vs-cats-redux-kernels-edition:\\
after having downloaded and extracted the raw data from the source, these are the data you can use to solve this task:\\
\begin{lstlisting}[style=bashstyle]
/PATH/TO/RAW/DATA
+-dogs-vs-cats-redux-kernels-edition/
  |
  +-test/
  | +-8715.jpg
  | +-12271.jpg
  | +-156.jpg
  | +-... <additional files>
  |
  +-train/
    +-cat.8852.jpg
    +-dog.9444.jpg
    +-cat.12231.jpg
    +-... <additional files>
\end{lstlisting}

\#\#\# Summary of Data Organization\\

The dataset is organized as follows:\\

\#\#\#\# Directory Structure\\
\begin{lstlisting}[style=bashstyle]
/PATH/TO/RAW/DATA/dogs-vs-cats-redux-kernels-edition/
  |
  +-test/
  |  +-8715.jpg
  |  +-12271.jpg
  |  +-156.jpg
  |  +-... (additional files)
  |
  +-train/
  |  +-cat.8852.jpg
  |  +-dog.9444.jpg
  |  +-cat.12231.jpg
  |  +-... (additional files)
  |
  +-sample_submission.csv
\end{lstlisting}

\#\#\#\# File Descriptions\\
1.**/PATH/TO/RAW/DATA/dogs-vs-cats-redux-kernels-edition/train/**\\
   - Contains 25,000 images of dogs and cats.\\
   - Each image filename includes the label:\\
     - `cat.$<$number$>$.jpg` for cat images.\\
     - `dog.$<$number$>$.jpg` for dog images.\\

2.**/PATH/TO/RAW/DATA/dogs-vs-cats-redux-kernels-edition/test/**\\
   - Contains 12,500 images.\\
   - Each image is named according to a numeric ID (e.g., `8715.jpg`).\\

3.**/PATH/TO/RAW/DATA/dogs-vs-cats-redux-kernels-edition/sample\_submission.csv**\\
   - A sample submission file with the following columns:\\
     - `id`: The numeric ID of the test image.\\
     - `label`: The predicted probability that the image is a dog (1 = dog, 0 = cat).\\

\#\#\#\# Feature and Target Names\\
- **Features:**\\
  - `id`: The numeric ID of the test image (used for the test set).\\

- **Target:**\\
  - `label`: The predicted probability that the image is a dog (1 = dog, 0 = cat).\\

\#\#\# Note\\
- The data is saved under `/PATH/TO/RAW/DATA/dogs-vs-cats-redux-kernels-edition`.\\
- The directory and file names must remain unchanged.\\

\#\# Available solutions\\
This is the list of the solutions that have already been generated:\\

* 2024-10-16-07-22-08: The provided code implements a comprehensive pipeline for the Dogs vs. Cats classification task. It includes a set of torchvision transforms to preprocess images, a PyTorch class `ImageEmbedder` that uses a pre-trained ResNet50 model to generate latent representations of images, and a `TabularHead` model to decode these embeddings into classification logits. The pipeline also defines appropriate loss functions for both regression and classification tasks, ensuring a robust and flexible approach to the problem.\\

\#\# Current solution progress\\
You have started creating a new solution. Its status is as follows:\\
- [PENDING] Embedder for images\\
    - [PENDING] Transform functions for images\\
    - [TODO] Embedder for transformed images\\
- [TODO] Logits and regression targets decoder\\

\#\# What to do next\\
You have to generate a code to transform the images using a composition of torchvision transforms.
You need to write a pytorch torchvision.transforms.Compose sequence that transforms the images appropriately with respect to the task.
You need to make sure the images are also transformed to torch.Tensor at some point as the images will be fed to a vision neural network.
You can use any image transforms (Resize, RandomCrop, ToTensor, ...) or even write custom transforms with Lambda() ...\\

Output a plan presented with maximum 4 bullet points describing the different steps to construct a pytorch torchvision transforms composition.
The plan should be tailored to the current task as it will be used to guide the implementation of the image transforms.
\end{userbox}

\begin{flushright}
\begin{assistantbox}
\#\#\# Plan for Constructing PyTorch torchvision Transforms Composition\\

1. **Resize Images**: Resize all images to a standard size (e.g., 224x224 pixels) to ensure they can be fed into a pre-trained model like ResNet50, which expects input of this size.\\

2. **Apply Random Augmentations**: Use random augmentations such as `RandomHorizontalFlip`, `RandomRotation`, and `ColorJitter` to increase the diversity of the training data and improve the model's generalization capabilities.\\

3. **Convert to Tensor**: Use `ToTensor` to convert the images to PyTorch tensors, which is a required format for feeding the images into a neural network.\\

4. **Normalize Images**: Apply normalization using the mean and standard deviation of the ImageNet dataset, as pre-trained models like ResNet50 are typically trained on ImageNet. This ensures the input data is in the expected range and format for the pre-trained model.
\end{assistantbox}
\end{flushright}

\begin{userbox}
\begin{lstlisting}[style=bashstyle]
<Skipped first part of prompt repeating the previous prompt for display purposes in paper appendix>
\end{lstlisting}

\#\# Current solution progress\\
You have started creating a new solution. Its status is as follows:\\
- [PENDING] Embedder for images\\
    - [PENDING] Transform functions for images\\
    - [TODO] Embedder for transformed images\\
- [TODO] Logits and regression targets decoder\\

\#\# What to do next\\
You have to generate a code to transform the images using a composition of torchvision transforms.
You need to write a pytorch torchvision.transforms.Compose sequence that transforms the images appropriately with respect to the task.
You need to make sure the images are also transformed to torch.Tensor at some point as the images will be fed to a vision neural network.
You can use any image transforms (Resize, RandomCrop, ToTensor, ...) or even write custom transforms with Lambda() ...\\

For example a good default transform is:\\
\begin{lstlisting}[style=pythonstyle]
import numpy as np
import pandas as pd
import torch
from PIL import Image
from torch import Tensor
from torch.utils.data import Dataset
from torchvision import transforms as T

DefaultImageInputTransform: T.Compose = T.Compose(
    [
        T.Resize((512, 512)),  # Resize to a fixed size (should make sense with respect to the task, not always needed)
        T.Lambda(lambda x: x if isinstance(x, torch.Tensor) else T.ToTensor()(x)),  # Convert image to tensor if needed
        T.Lambda(
            lambda x: x.repeat([3 if (i - x.ndim == -3 and d == 1) else 1 for i, d in enumerate(x.shape)])
        ),  # grayscale to rgb
        T.Lambda(lambda x: x[:3] if x.shape[0] > 3 else x),  # take only first 3 channels
    ]
)
\end{lstlisting}

Note that the last two transforms are using a Lambda function.
The penultimate one is an important one to include as its goal is to duplicate the number of channels to 3 if there happen to be grayscale images in the data
mixed with colored images.
The ultimate one is taking on ly the first 3 channels in case the images in the dataset have an inconsistent number of channels.
These two last transforms hence force all images to have the same number of channels.
If you are not sure all images in the data will have the same number of channels, it is advised to add a similar transform.
If you find it necessary to normalize the images, do it after taking care fo the number of channels and make sure the vectors you use match the image dimensions.\\

Use your plans to help you implement the image transform:\\
\#\#\# Plan for Constructing PyTorch torchvision Transforms Composition\\

1. **Resize Images**: Resize all images to a standard size (e.g., 224x224 pixels) to ensure they can be fed into a pre-trained model like ResNet50, which expects input of this size.\\

2. **Apply Random Augmentations**: Use random augmentations such as `RandomHorizontalFlip`, `RandomRotation`, and `ColorJitter` to increase the diversity of the training data and improve the model's generalization capabilities.\\

3. **Convert to Tensor**: Use `ToTensor` to convert the images to PyTorch tensors, which is a required format for feeding the images into a neural network.\\

4. **Normalize Images**: Apply normalization using the mean and standard deviation of the ImageNet dataset, as pre-trained models like ResNet50 are typically
trained on ImageNet. This ensures the input data is in the expected range and format for the pre-trained model.\\

Write the code to create an image transforms composition.\\
1. You must fill all the required parameters of the transforms you are using in a sensible way with respect to the image shapes of the current task.\\
2. You must not forget to import the packages when needed.\\
3. The transforms should be transforming the images to tensor at the appropriate step.\\
4. Try to use different transforms than this last attempt: The provided code implements a comprehensive pipeline for the Dogs vs. Cats classification task. It
 includes a set of torchvision transforms to preprocess images, a PyTorch class `ImageEmbedder` that uses a pre-trained ResNet50 model to generate latent repr
esentations of images, and a `TabularHead` model to decode these embeddings into classification logits. The pipeline also defines appropriate loss functions f
or both regression and classification tasks, ensuring a robust and flexible approach to the problem.. Do not repeat all the transforms from the attempt.\\

Your task is to fill the blank and implement the image transform in this code:\\
\begin{lstlisting}[style=pythonstyle]
# <|fim_begin|>
import os
import pandas as pd
import torch
from torch import nn
from PIL import Image
import torchvision.transforms as T
from torch.utils.data import DataLoader
from tqdm import tqdm

from agent.tools.data_map.map_dataset import MapDataset, map_dataset_collate_function

# --- Create CustomImageInputTransform, a composition of torchvision.transforms using torchvision.transforms.Compose
# <|fim_hole|>
# --- [End]

# <|fim_end|>
# Test that the transform works
root_path = "/PATH/TO/RAW/DATA/dogs-vs-cats-redux-kernels-edition/qwen2.5-72b-vp.3.5_seed_2"
tab_target_map_path=os.path.join(root_path, "train_tab_target_map.csv")
img_target_map_path=os.path.join(root_path, "train_img_target_map.csv")
train_dataset = MapDataset(
    img_input_map_path=os.path.join(root_path, "train_img_input_map.csv"),
    tab_target_map_path=tab_target_map_path if os.path.exists(tab_target_map_path) else None,
    img_target_map_path=img_target_map_path if os.path.exists(img_target_map_path) else None,
    img_input_transform=CustomImageInputTransform,
)
train_dataloader = DataLoader(
    train_dataset, batch_size=256, shuffle=True, collate_fn=map_dataset_collate_function, num_workers=20
)
shapes = set()
for batch in tqdm(train_dataloader):
    (indices, (_, img_inputs_batch, _), (tab_targets_batch, _, _)) = batch
    if img_inputs_batch.shape not in shapes:
        shapes.add(img_inputs_batch.shape[1:])  # ignore batch size as last batch can have fewer elements
    if len(shapes) > 1:
        raise RuntimeError(
            f"Multiple shapes detected! After applying CusotmImageTransform, some images have shapes "
            f"{' and '.join([str(s) for s in shapes])} "
            f"but should have the same shape for all images after the transform. "
            f"(Note that the first dimension is the batch size {train_dataloader.batch_size}).\n"
            f"Hint: if needed add a lambda transform that ensures the number of channels is, e.g. 3 in this example:\n"
            f"`T.Lambda(lambda x: x.repeat([3 if (i - x.ndim == -3 and d == 1) else 1 for i, d in enumerate(x.shape)])),`"
            f"\nand\n`T.Lambda(lambda x: x[:3] if x.shape[0] > 3 else x)`"
        )

print(f"Transformed image tensor shape: {img_inputs_batch.shape}")
\end{lstlisting}
\end{userbox}

\begin{flushright}
\begin{assistantbox}
\begin{lstlisting}[style=pythonstyle]
CustomImageInputTransform = T.Compose(
    [
        T.Resize((256, 256)),  # Resize to a standard size
        T.RandomCrop(224),  # Randomly crop to the input size expected by the model
        T.RandomHorizontalFlip(p=0.5),  # Random horizontal flip with 50% probability
        T.ColorJitter(brightness=0.2, contrast=0.2, saturation=0.2, hue=0.1),  # Random color jitter
        T.ToTensor(),  # Convert image to tensor
        T.Normalize(mean=[0.485, 0.456, 0.406], std=[0.229, 0.224, 0.225]),  # Normalize using ImageNet mean and std
        T.Lambda(
            lambda x: x.repeat([3 if (i - x.ndim == -3 and d == 1) else 1 for i, d in enumerate(x.shape)])
        ),  # grayscale to rgb
        T.Lambda(lambda x: x[:3] if x.shape[0] > 3 else x),  # take only first 3 channels
    ]
)
\end{lstlisting}
\end{assistantbox}
\end{flushright}

\begin{userbox}
\begin{lstlisting}[style=bashstyle]
<Skipped first part of prompt repeating the previous prompt for display purposes in paper appendix>
\end{lstlisting}
\#\# Current solution progress\\
You have started creating a new solution. Its status is as follows:\\
- [TODO] Embedder for text fields\\
- [TODO] Logits and regression targets decoder\\

\#\# What to do next\\
You have to generate a code to embed the images using a pre-trained torchvision model.
You need to write a pytorch torchvision model, where forward() takes the tensor transformed images as input, and that outputs a latent representation as a ten
sor.
You can use any pre-trained model from this list: resnet18, resnet50, resnet101, AlexNet, EfficientNetV2, GoogLeNet, MaxVit, MNASNet, RegNet\\

Output a plan presented with maximum 4 bullet points describing the different steps to construct a pytorch torchvision model to embed images.
The plan should be tailored to the current task as it will be used to guide the implementation of the embedder.
\end{userbox}

\begin{flushright}
\begin{assistantbox}
\#\#\# Plan to Construct a PyTorch torchvision Model to Embed Images\\

1. **Select and Load Pre-trained Model:**\\
   - Choose a pre-trained model from the provided list (e.g., `resnet50`).\\
   - Load the pre-trained model and remove the final fully connected layer to use it for feature extraction.\\

2. **Define Image Preprocessing:**\\
   - Define a set of torchvision transforms to preprocess the images (e.g., resizing, normalization).\\
   - Ensure the transforms are consistent with the pre-processing used during the training of the pre-trained model.\\

3. **Create Image Embedder Class:**\\
   - Define a PyTorch class `ImageEmbedder` that takes the pre-processed image tensor as input.\\
   - Use the pre-trained model to extract the latent representation (features) from the images.\\

4. **Implement Forward Method:**\\
   - Implement the `forward` method in the `ImageEmbedder` class to pass the input image tensor through the pre-trained model.\\
   - Return the latent representation tensor as the output of the `forward` method.\\

This plan will guide the implementation of the embedder, ensuring it is tailored to the current task of classifying dogs and cats.
\end{assistantbox}
\end{flushright}

\begin{userbox}
\begin{lstlisting}[style=bashstyle]
<Skipped first part of prompt repeating the previous prompt for display purposes in paper appendix>
\end{lstlisting}
\#\# Current solution progress\\
You have started creating a new solution. Its status is as follows:\\
- [TODO] Embedder for text fields\\
- [TODO] Logits and regression targets decoder\\

\#\# What to do next\\
You have to generate a code to embed the images using a pre-trained torchvision model.
You need to write a pytorch torchvision model, where forward() takes the tensor transformed images as input, and that outputs a latent representation as a ten
sor.
You can use any pre-trained models like resnet18, resnet50, resnet101, AlexNet, EfficientNetV2, GoogLeNet, MaxVit, MNASNet, RegNet, vit etc...\\

Ensure to unfreeze last few layers of the model based on the model you choose.\\

An example of this model code is:\\
\begin{lstlisting}[style=pythonstyle]
import torch
import torchvision.models as models
from torch import nn

class ImageEmbedder(nn.Module):
    def __init__(self, pretrained=True):
        super().__init__()

        resnet50 = models.resnet50(pretrained=pretrained)
        self.features = nn.ModuleList(resnet50.children())[:-1]
        self.features = nn.Sequential(*self.features)
        # optionally freeze pre-trained model weights
        for param in self.features.parameters():
            param.requires_grad = False

    def forward(self, x):
        output = self.features(x)
        output = output.view(x.size(0), -1)
        return output

    def unfreeze(self, n_last_layers=2):
        """Unfreeze some layers for fine-tuning."""
        layers = list(self.children())[:-1]

        for layer in layers[-n_last_layers:]:
            for param in layer.parameters():
                param.requires_grad = True
\end{lstlisting}

Use your plans to help you implement the pytorch embedder model:\\
\#\#\# Plan to Construct a PyTorch torchvision Model to Embed Images\\

1. **Select and Load Pre-trained Model:**\\
   - Choose a pre-trained model from the provided list (e.g., `resnet50`).\\
   - Load the pre-trained model and remove the final fully connected layer to use it for feature extraction.\\

2. **Define Image Preprocessing:**\\
   - Define a set of torchvision transforms to preprocess the images (e.g., resizing, normalization).\\
   - Ensure the transforms are consistent with the pre-processing used during the training of the pre-trained model.\\

3. **Create Image Embedder Class:**\\
   - Define a PyTorch class `ImageEmbedder` that takes the pre-processed image tensor as input.\\
   - Use the pre-trained model to extract the latent representation (features) from the images.\\

4. **Implement Forward Method:**\\
   - Implement the `forward` method in the `ImageEmbedder` class to pass the input image tensor through the pre-trained model.\\
   - Return the latent representation tensor as the output of the `forward` method.\\

This plan will guide the implementation of the embedder, ensuring it is tailored to the current task of classifying dogs and cats.\\

Write the code to create a model taking tensors of preprocessed features as input, and outputting an embedding tensor.\\
1. You can assume that the input dimension is already computed in the variable `input\_dim`.\\
2. You must not forget to import the packages when needed.\\
3. The model should be able to take as input a torch tensor of size `(batch, input\_dim)` and output a tensor of shape `(batch, latent\_dim)`.\\
4. Try to use newer models than this last attempt: The provided code implements a comprehensive pipeline for the Dogs vs. Cats classification task. It include
s a set of torchvision transforms to preprocess images, a PyTorch class `ImageEmbedder` that uses a pre-trained ResNet50 model to generate latent representations of images, and a `TabularHead` model to decode these embeddings into classification logits. The pipeline also defines appropriate loss functions for both
regression and classification tasks, ensuring a robust and flexible approach to the problem.. Do not repeat these pre-trained models from the attempt.\\

Your task is to fill the blank and implement the image embedder in this code:\\
\begin{lstlisting}[style=pythonstyle]
# <|fim_begin|>
import os.path

import torch
import pandas as pd
import torchvision.models as models
import torchvision.transforms as T
from torch import nn
from PIL import Image

from submissions.new_submission.img_transform import CustomImageInputTransform

# Load image features
root_path = "/PATH/TO/SETUP"
train_data_path = os.path.join(root_path, "train_img_input_map.csv")
x = pd.read_csv(train_data_path, index_col="id")

# --- Design torch model using a pre-trained torchvision model
# Create class ImageEmbedder(nn.Module) and implementing the methods:
#   - __init__() defines the model architecture, using a pretrained model such as resnet50
#   - forward:(x: torch.Tensor) -> torch.Tensor, embeds a batch of inputs and loads the image embeddings.
# <|fim_hole|>
# --- [End]

# Test that the model works
path = x.iloc[0, 0]
image = Image.open(path)
tensor_image = CustomImageInputTransform(image)
model = ImageEmbedder()
X_embed = model(tensor_image.unsqueeze(0))


# <|fim_end|>
# @NO_MEMORY_START@

def test_unfreeze_functionality() -> None:
    model = ImageEmbedder()
    assert hasattr(model, 'unfreeze'), "The model does not have an 'unfreeze' method."

    try:
        model.unfreeze()
    except Exception as e:
        assert False, f"Unfreeze method raised an exception: {e}"

    print("Unfreeze method exists and runs successfully.")


test_unfreeze_functionality()

print(f"Embedding size: {X_embed.shape}")
# @NO_MEMORY_END@
\end{lstlisting}
Only output the code that should be filled in the blank with no extra text. Don't forget to import any additional packages that you will use in the code.\\
Your response should be in the format:\\
```python\\
$<$code$>$\\
```
\end{userbox}

\begin{flushright}
\begin{assistantbox}
\begin{lstlisting}[style=pythonstyle]
class ImageEmbedder(nn.Module):
    def __init__(self, pretrained=True, latent_dim=2048):
        super().__init__()

        # Choose a newer model like EfficientNetV2
        efficientnet = models.efficientnet_v2_m(pretrained=pretrained)
        self.features = nn.ModuleList(efficientnet.features.children())
        self.features = nn.Sequential(*self.features)
        # Optionally freeze pre-trained model weights
        for param in self.features.parameters():
            param.requires_grad = False

    def forward(self, x):
        output = self.features(x)
        output = output.mean(dim=[2, 3])  

    def unfreeze(self, n_last_layers=2):
        """Unfreeze some layers for fine-tuning."""
        layers = list(self.features.children())

        for layer in layers[-n_last_layers:]:
            for param in layer.parameters():
                param.requires_grad = True
\end{lstlisting}
\end{assistantbox}
\end{flushright}
\clearpage

\subsection{Scaffold with CoT}

The example below is taken from a natural language processing competition (\texttt{sentiment-analysis-on-movie-reviews}). The solution below was generated by \OurAgent{} as one of the draft nodes. \OurAgent{} is provided with a CoT from the scaffolding stage to guide its initial solution generations.

\begin{userbox}
\#\# Introduction\\
You are a Kaggle grandmaster attending a competition. In order to win this competition, you need to come up with an excellent and creative plan for a solution and then implement this solution in Python. We will now provide a description of the task.\\

\#\# BENCHMARK INSTRUCTIONS\\
First, please read the instructions below \\
- **Competition-specific instructions**: Competition specific descriptions will be provided.\\
- **Dataset**: The dataset files for the competition are available in `*/PATH/TO/RAW/DATA/sentiment-analysis-on-movie-reviews`.Test dataset does not contain any targets.\\
- **Submission**: You can find sample submission format inside `*/PATH/TO/RAW/DATA/sentiment-analysis-on-movie-reviews` as specified `Data Overview.Always use same 'id' available inside sample submission file. You are not allowed to change 'id'.Based on sample submission file you have to generate submission.csv for the test data.
- **Cheating**: You are not allowed to view others' solutions to this competition. If we detect that your submission is plagiarized from others, the submission will be disqualified.\\

Wherever there is a conflict between the points above and the competition-specific instructions, competition points always takes priority.\\

Your submission will be graded based on the metric in the competition-specific instructions. Your goal is to achieve the best score possible on the competition task.\\

\#\# ADDITIONAL NOTES\\
**Compute**: You have access to 1 Tesla V100-PCIE-32GB with the appropriate drivers installed.\\
**Total Runtime**: You have a maximum of 172800 seconds to attempt this task. You may iteratively develop your solution within this time frame, but bear in mind that program runtime also counts towards this limit.\\
**Total Steps**: You have a maximum of 5000 steps to submit your solution.\\
You will be informed of the remaining time and steps at each step
.**Use as much of the available time as possible to refine and optimize your submission**: Your goal is to achieve the best possible score, not just to submit your first working solution.\\

\#\# Response format\\
Your response should be a brief outline/sketch of your proposed solution in natural language (3-5 sentences), followed by a single markdown code block (wrapped in ```) which implements this solution and prints out the evaluation metric. There should be no additional headings or text in your response. Just natural language text followed by a newline and then the markdown code block.\\

\#\# Solution sketch guideline\\
- Carefully examine the past solutions. These solutions may not be identical to the task at hand, but they provide valuable insights on strategies and techniques.\\
- Look for common strategies that led to success, such as specific modeling techniques, evaluation metrics, or feature engineering methods.\\
- Incorporate the lessons learned from these solutions into your proposed solution, making sure to adapt them to the current task.\\
- The solutions should inform how to approach the current task, focusing on critical decisions made in those examples and any relevant best practices.\\
- Do not copy the solutions verbatim, but leverage them as a foundation for crafting a novel, efficient, and creative approach.\\
- You are encouraged to explore new angles while respecting the context and recommendations from the previous successful strategies.\\
- Take the Memory section into consideration when proposing the design\\
- don't propose the same modelling solution but keep the evaluation the same.\\
- The solution sketch should be 3-5 sentences.\\
- Propose an evaluation metric that is reasonable for this task.\\
- Don't suggest to do EDA.\\

\#\# Installed Packages\\
Your solution can use any relevant machine learning packages such as: `torchvision`, `xgboost`, `timm`, `pandas`, `torch-geometric`, `scikit-learn`, `statsmodels`, `numpy`, `torch`, `bayesian-optimization`, `lightGBM`. Feel free to use any other packages too (all packages are already installed!). For neural networks we suggest using PyTorch rather than TensorFlow. \\

\#\# Implementation guideline \\
- \texttt{<TOTAL\_TIME\_REMAINING: 41.0hrs 23.0mins 32secs>} \\
- \texttt{<TOTAL\_STEPS\_REMAINING: 4987>} \\
- The code should **implement the proposed solution**, **print the value of the evaluation metric computed on a hold-out validation set**,\\
- **AND MOST IMPORTANTLY SAVE PREDICTIONS ON THE PROVIDED UNLABELED TEST DATA IN A `submission.csv` FILE IN THE ./submission/ DIRECTORY.**\\
- The code should be a single-file python program that is self-contained and can be executed as-is.\\
- No parts of the code should be skipped, don't terminate the before finishing the script.\\
- Your response should only contain a single code block.\\
- Be aware of the running time of the code, it should complete within 18 hours.\\
- Do not copy data to workspace, read data from the source specified.\\
- You can use the "./working" directory to store any temporary files that your code needs to create.\\
- The evaluation should be based on 5-fold cross-validation but only if that's an appropriate evaluation for the task at hand.\\

\#\# Data Overview\\

\begin{lstlisting}[style=bashstyle]
/PATH/TO/RAW/DATA
  |
  +-sampleSubmission.csv (66293 lines)
  |
  +-.. and 2 other files
\end{lstlisting}

\#\# Task description\\

Dataset Description \\
The dataset is comprised of tab-separated files with phrases from the Rotten Tomatoes dataset. The train/test split has been preserved for the purposes of benchmarking, but the sentences have been shuffled from their original order. Each Sentence has been parsed into many phrases by the Stanford parser. Each phrase has a PhraseId. Each sentence has a SentenceId. Phrases that are repeated (such as short/common words) are only included once in the data.\\
train.tsv contains the phrases and their associated sentiment labels. We have additionally provided a SentenceId so that you can track which phrases belong to a single sentence.\\
test.tsv contains just phrases. You must assign a sentiment label to each phrase.\\
The sentiment labels are:\\
0 - negative\\
1 - somewhat negative\\
2 - neutral\\
3 - somewhat positive\\
4 - positive\\

Evaluation\\
Submissions are evaluated on classification accuracy (the percent of labels that are predicted correctly) for every parsed phrase. The sentiment labels are:\\
0 - negative\\
1 - somewhat negative\\
2 - neutral\\
3 - somewhat positive\\
4 - positive\\
Submission Format\\
For each phrase in the test set, predict a label for the sentiment. Your submission should have a header and look like the following:\\
PhraseId,Sentiment\\
156061,2\\
156062,2\\
156063,2\\
...\\

\#\# Summary of Past Submissions\\

\#\#\# Submission 0 summary and metric value on a validation set:
The project involves a comprehensive pipeline for sentiment analysis. A TextEmbedder class utilizes a pre-trained DistilBERT model to convert text inputs into embeddings. To address class imbalances, appropriate techniques are applied. A PyTorch model with a TabularHead class is implemented to transform the text embeddings into sentiment logits, and both regression and classification loss functions are defined to train the model effectively.\\
{"submission.csv": 65.57734204793029, "submission\_alt.csv": 65.57734204793029}\\
-----\\
\#\#\# Submission 1 summary and metric value on a validation set:\\
The code implements a PyTorch model that uses a pre-trained RoBERTa model to embed text fields, tokenizes the input text, and returns the pooler output as the text embedding. It handles class imbalances and includes a TabularHead class to transform the RoBERTa embeddings into sentiment logits using fully connected layers with dropout and batch normalization. The code also implements regression and classification loss functions using Mean Squared Error (MSE) and Cross-Entropy loss, respectively.\\
{"submission.csv": 66.00794566192491, "submission\_alt.csv": 66.00794566192491}\\
-----\\
\#\#\# Submission 2 summary and metric value on a validation set:\\
The code implements a PyTorch model that uses a pre-trained BERT model to embed text fields. It tokenizes the input text and returns the pooler output as the text embedding. To handle class imbalances, appropriate techniques are applied. The model also includes a TabularHead, which is a series of fully connected layers that transform the BERT embeddings into sentiment logits. For the classification task, Cross-Entropy loss is used, and for the regression task, Mean Squared Error (MSE) loss is employed.\\
{"submission.csv": 66.42573369216967, "submission\_alt.csv": 66.42573369216967}\\

\end{userbox}

\begin{flushright}
\begin{assistantbox}
\#\#\# To tackle this sentiment analysis task, I will use a pre-trained BERT model to extract features from the text and then train a simple classifier on top of these features. The model will be trained using a cross-entropy loss function, and class imbalance will be handled using weighted loss. The performance will be evaluated using classification accuracy on a hold-out validation set..\\

\begin{lstlisting}[style=pythonstyle]
import pandas as pd
import numpy as np
import torch
from torch.utils.data import Dataset, DataLoader
from transformers import BertTokenizer, BertModel
from sklearn.model_selection import train_test_split
from sklearn.metrics import accuracy_score
import torch.nn as nn
import torch.optim as optim


# Define the dataset
class SentimentDataset(Dataset):
    def __init__(self, df, tokenizer, max_len=128):
        self.tokenizer = tokenizer
        self.max_len = max_len
        self.df = df

    def __len__(self):
        return len(self.df)

    def __getitem__(self, idx):
        text = self.df.iloc[idx]["Phrase"]
        if pd.isna(text):
            text = ""  # Handle NaN values
        label = (
            self.df.iloc[idx]["Sentiment"] if "Sentiment" in self.df.columns else None
        )

        encoding = self.tokenizer.encode_plus(
            text,
            add_special_tokens=True,
            max_length=self.max_len,
            padding="max_length",
            truncation=True,
            return_attention_mask=True,
            return_tensors="pt",
        )

        input_ids = encoding["input_ids"].flatten()
        attention_mask = encoding["attention_mask"].flatten()

        if label is not None:
            return {
                "input_ids": input_ids,
                "attention_mask": attention_mask,
                "label": torch.tensor(label, dtype=torch.long),
            }
        else:
            return {"input_ids": input_ids, "attention_mask": attention_mask}


# Define the model
class SentimentClassifier(nn.Module):
    def __init__(self, bert_model, num_classes=5):
        super(SentimentClassifier, self).__init__()
        self.bert = bert_model
        self.classifier = nn.Linear(bert_model.config.hidden_size, num_classes)

    def forward(self, input_ids, attention_mask):
        outputs = self.bert(input_ids=input_ids, attention_mask=attention_mask)
        pooled_output = outputs.pooler_output
        logits = self.classifier(pooled_output)
        return logits


# Load the data
train_df = pd.read_csv(
    "/PATH/TO/RAW/DATA/sentiment-analysis-on-movie-reviews/train.tsv",
    sep="\t",
)
test_df = pd.read_csv(
    "/PATH/TO/RAW/DATA/sentiment-analysis-on-movie-reviews/test.tsv",
    sep="\t",
)

# Split the data into training and validation sets
train_df, val_df = train_test_split(train_df, test_size=0.2, random_state=42)

# Tokenizer and model
tokenizer = BertTokenizer.from_pretrained("bert-base-uncased")
bert_model = BertModel.from_pretrained("bert-base-uncased")

# Create datasets and dataloaders
train_dataset = SentimentDataset(train_df, tokenizer)
val_dataset = SentimentDataset(val_df, tokenizer)
test_dataset = SentimentDataset(test_df, tokenizer)

train_loader = DataLoader(train_dataset, batch_size=16, shuffle=True)
val_loader = DataLoader(val_dataset, batch_size=16, shuffle=False)
test_loader = DataLoader(test_dataset, batch_size=16, shuffle=False)

# Model, loss, and optimizer
model = SentimentClassifier(bert_model)
model = model.to("cuda")

criterion = nn.CrossEntropyLoss(
    weight=torch.tensor([1.0, 1.0, 1.0, 1.0, 1.0]).to("cuda")
)
optimizer = optim.Adam(model.parameters(), lr=2e-5)

# Training loop
num_epochs = 3
for epoch in range(num_epochs):
    model.train()
    for batch in train_loader:
        input_ids = batch["input_ids"].to("cuda")
        attention_mask = batch["attention_mask"].to("cuda")
        labels = batch["label"].to("cuda")

        optimizer.zero_grad()
        outputs = model(input_ids, attention_mask)
        loss = criterion(outputs, labels)
        loss.backward()
        optimizer.step()

    # Validation
    model.eval()
    val_predictions = []
    val_labels = []
    with torch.no_grad():
        for batch in val_loader:
            input_ids = batch["input_ids"].to("cuda")
            attention_mask = batch["attention_mask"].to("cuda")
            labels = batch["label"].to("cuda")

            outputs = model(input_ids, attention_mask)
            _, preds = torch.max(outputs, dim=1)

            val_predictions.extend(preds.cpu().numpy())
            val_labels.extend(labels.cpu().numpy())

    val_accuracy = accuracy_score(val_labels, val_predictions)
    print(f"Epoch {epoch + 1}/{num_epochs}, Validation Accuracy: {val_accuracy:.4f}")

# Generate predictions for the test set
model.eval()
test_predictions = []
with torch.no_grad():
    for batch in test_loader:
        input_ids = batch["input_ids"].to("cuda")
        attention_mask = batch["attention_mask"].to("cuda")

        outputs = model(input_ids, attention_mask)
        _, preds = torch.max(outputs, dim=1)

        test_predictions.extend(preds.cpu().numpy())

# Save predictions to submission.csv
submission = pd.DataFrame(
    {"PhraseId": test_df["PhraseId"], "Sentiment": test_predictions}
)
submission.to_csv("./submission/submission.csv", index=False)
\end{lstlisting}
\end{assistantbox}
\end{flushright}

The example below is taken from a computer vision competition \texttt{5-flowers-image-classification}.The solution was generated by \OurAgent{} as one of the draft nodes. \OurAgent{} is provided with CoT from the scaffolding stage to guide its initial solution generation.

\begin{userbox}

\begin{lstlisting}[style=bashstyle]
...
<Skipped the common instructions for prompt display purposes in paper appendix, only changes are added below>
- <TOTAL_TIME_REMAINING: 44.0hrs 28.0mins 57secs> 
- <TOTAL_STEPS_REMAINING: 4996>
...
\end{lstlisting}

\#\# Data Overview\\
\begin{lstlisting}[style=bashstyle]
/PATH/TO/RAW/DATA
  |
  +-Sample_submission.csv (401 lines)
  |
  +-test/
  | +-1.jpg (31.8 kB)
  | +-10.jpg (4.5 kB)
  | +-100.jpg (16.9 kB)
  | +-... and 397 other files
  |
  +-train/
    |    +daisy/
    |    |   +-100080576_f52e8ee070_n.jpg (26.8 kB)
    |    |   +-10140303196_b88d3d6cec.jpg (117.2 kB)
    |    |   +-10172379554_b296050f82_n.jpg (36.4 kB)
    |    |   +-... and 688 other files
    |    +dandelion/
    |    |   +-10043234166_e6dd915111_n.jpg (26.7 kB)
    |    |   +-10200780773_c6051a7d71_n.jpg (23.3 kB)
    |    |   +-10294487385_92a0676c7d_m.jpg (21.7 kB)
    |    |   +-... and 948 other files
    |    +rose/
    |    |   +-10090824183_d02c613f10_m.jpg (20.2 kB)
    |    |   +-102501987_3cdb8e5394_n.jpg (29.8 kB)
    |    |   +-10503217854_e66a804309.jpg (208.6 kB)
    |    |   +-... and 691 other files
    |    +sunflower/
    |    |   +-1008566138_6927679c8a.jpg (143.2 kB)
    |    |   +-1022552002_2b93faf9e7_n.jpg (42.7 kB)
    |    |   +-1022552036_67d33d5bd8_n.jpg (60.8 kB)
    |    |   +-... and 656 other files
    |    +tulip/
    |    |   +-100930342_92e8746431_n.jpg (26.2 kB)
    |    |   +-10094729603_eeca3f2cb6.jpg (214.8 kB)
    |    |   +-10094731133_94a942463c.jpg (225.0 kB)
    |    |   +-... and 867 other files

\end{lstlisting}

\#\# Task description\\

Sample\_submission.csv has 400 rows and 2 columns. \\
The columns are: id, label \\

Dataset Description \\
Image classification: 5 kinds of flowers\\
There are pictures with 5 kinds of flowers.\\
labels: daisy/dandelion/rose/sunflower/tulip -> 0/1/2/3/4\\
The labels in submission file should be integer (0/1/2/3/4).\\
train data: 5 folders of train data\\
test data: 1 folder of test data\\

Evaluation \\
The evaluation metric for this competition is Categorization Accuracy. Labels in your submission file should be integer (or string) type, i.e., 0/1/2/3/4. \\rgin5, margin6, margin7, margin8, margin9, margin10, margin11, margin12, margin13... and 179 more columns\\

\#\# Summary of Past Submissions\\

\#\#\# Submission 0 summary and metric value on a validation set:\\
The project includes a comprehensive pipeline for processing and extracting features from images using a pre-trained EfficientNetV2-S model. Custom transformation pipelines are defined for both training and testing images, incorporating various augmentations and normalization techniques. The ImageEmbedder model is designed to generate latent representations of images, with the option to fine-tune the last few layers. Additionally, the project addresses class imbalances and implements a TabularHead model for handling both classification and regression tasks, complete with appropriate loss functions.\\
{"submission.csv": 0.9627714581178903}\\
-----\\
\#\#\# Submission 1 summary and metric value on a validation set:\\
The project involves creating a comprehensive pipeline for image processing and feature extraction using EfficientNetB0. For training images, a custom transformation pipeline is defined, including resizing, center cropping, random horizontal flips, random rotations, color jitter, conversion to tensor, and normalization. For test images, a simpler transformation pipeline is used, involving resizing, center cropping, conversion to tensor, and normalization. An ImageEmbedder class is implemented to extract latent representations from the transformed images, with the option to unfreeze the last few layers for fine-tuning. Additionally, class imbalances are handled, and a TabularHead model is implemented to decode the image embeddings and predict both regression and classification targets, along with corresponding loss functions.\\
{"submission.csv": 0.9524301964839711}\\
-----\\
\#\#\# Submission 2 summary and metric value on a validation set:\\
The project involves creating an end-to-end image processing pipeline for a multi-task learning problem. It includes defining custom image transformation pipelines for both training and testing, using a pre-trained ResNet50 model to extract image embeddings, handling class imbalances, and implementing a custom PyTorch model (TabularHead) to process these embeddings and produce both classification and regression outputs. Loss functions for both tasks are defined to optimize the model's performance.\\
{"submission.csv": 0.9503619441571872}\\

\end{userbox}

\begin{flushright}
\begin{assistantbox}
\#\#\# To tackle this image classification task, I will use a pre-trained ResNet50 model from `torchvision.models` to extract features from the images. The model will be fine-tuned on the training data, and predictions will be made on the test set. The evaluation metric will be the categorization accuracy on a hold-out validation set. \\

\begin{lstlisting}[style=pythonstyle]
import os
import pandas as pd
import numpy as np
import torch
import torch.nn as nn
import torch.optim as optim
from torchvision import models, transforms
from torch.utils.data import DataLoader, Dataset
from PIL import Image
from sklearn.model_selection import train_test_split
from sklearn.metrics import accuracy_score

# Define the dataset
class FlowerDataset(Dataset):
    def __init__(self, root_dir, transform=None, mode='train'):
        self.root_dir = root_dir
        self.transform = transform
        self.mode = mode
        self.image_paths = []
        self.labels = []

        if mode == 'train':
            for label, flower in enumerate(['daisy', 'dandelion', 'rose', 'sunflower', 'tulip']):
                folder_path = os.path.join(root_dir, flower)
                for image_name in os.listdir(folder_path):
                    self.image_paths.append(os.path.join(folder_path, image_name))
                    self.labels.append(label)
        else:
            for image_name in os.listdir(root_dir):
                self.image_paths.append(os.path.join(root_dir, image_name))

    def __len__(self):
        return len(self.image_paths)

    def __getitem__(self, idx):
        image_path = self.image_paths[idx]
        image = Image.open(image_path).convert('RGB')
        if self.transform:
            image = self.transform(image)
        
        if self.mode == 'train':
            label = self.labels[idx]
            return image, label
        else:
            return image, image_path

# Define transformations
train_transform = transforms.Compose([
    transforms.Resize((224, 224)),
    transforms.RandomHorizontalFlip(),
    transforms.ToTensor(),
    transforms.Normalize(mean=[0.485, 0.456, 0.406], std=[0.229, 0.224, 0.225])
])

test_transform = transforms.Compose([
    transforms.Resize((224, 224)),
    transforms.ToTensor(),
    transforms.Normalize(mean=[0.485, 0.456, 0.406], std=[0.229, 0.224, 0.225])
])

# Load datasets
train_dataset = FlowerDataset('/PATH/TO/RAW/DATA/5-flowers-image-classification/train', transform=train_transform, mode='train')
train_data, val_data, train_labels, val_labels = train_test_split(train_dataset, train_dataset.labels, test_size=0.2, stratify=train_dataset.labels, random_state=42)

train_loader = DataLoader(train_data, batch_size=32, shuffle=True)
val_loader = DataLoader(val_data, batch_size=32, shuffle=False)

# Define the model
model = models.resnet50(pretrained=True)
num_features = model.fc.in_features
model.fc = nn.Linear(num_features, 5)
model = model.to('cuda')

# Define loss function and optimizer
criterion = nn.CrossEntropyLoss()
optimizer = optim.Adam(model.parameters(), lr=0.001)

# Training loop
num_epochs = 10
for epoch in range(num_epochs):
    model.train()
    running_loss = 0.0
    for inputs, labels in train_loader:
        inputs, labels = inputs.to('cuda'), labels.to('cuda')
        optimizer.zero_grad()
        outputs = model(inputs)
        loss = criterion(outputs, labels)
        loss.backward()
        optimizer.step()
        running_loss += loss.item() * inputs.size(0)
    
    # Validation
    model.eval()
    val_loss = 0.0
    correct = 0
    total = 0
    with torch.no_grad():
        for inputs, labels in val_loader:
            inputs, labels = inputs.to('cuda'), labels.to('cuda')
            outputs = model(inputs)
            loss = criterion(outputs, labels)
            val_loss += loss.item() * inputs.size(0)
            _, predicted = torch.max(outputs, 1)
            total += labels.size(0)
            correct += (predicted == labels).sum().item()
    
    val_accuracy = correct / total
    print(f'Epoch {epoch+1}/{num_epochs}, Train Loss: {running_loss/len(train_data):.4f}, Val Loss: {val_loss/len(val_data):.4f}, Val Accuracy: {val_accuracy:.4f}')

# Test set predictions
test_dataset = FlowerDataset('/PATH/TO/RAW/DATA/5-flowers-image-classification/test', transform=test_transform, mode='test')
test_loader = DataLoader(test_dataset, batch_size=32, shuffle=False)

model.eval()
predictions = []
image_ids = []
with torch.no_grad():
    for inputs, image_paths in test_loader:
        inputs = inputs.to('cuda')
        outputs = model(inputs)
        _, predicted = torch.max(outputs, 1)
        predictions.extend(predicted.cpu().numpy())
        image_ids.extend([os.path.basename(path) for path in image_paths])

# Save predictions to submission.csv
submission = pd.DataFrame({'id': image_ids, 'label': predictions})
submission['id'] = submission['id'].str.replace('.jpg', '')
submission.to_csv('./submission/submission.csv', index=False)
\end{lstlisting}
\end{assistantbox}
\end{flushright}

The example below is taken from a tabular competition \texttt{bike-sharing-demand}.The solution was generated by \OurAgent{} as one of the draft nodes. \OurAgent{} is provided with CoT from the scaffolding stage to guide its initial solution generation.

\begin{userbox}

\begin{lstlisting}[style=bashstyle]
...
<Skipped the common instructions for prompt display purposes in paper appendix,  only changes are added below>
<**Total Runtime**: You have a maximum of 86400 seconds to attempt this task. You may iteratively develop your solution within this time frame, but bear in mind that program runtime also counts towards this limit.
**Total Steps**: You have a maximum of 5000 steps to submit your solution. 
- <TOTAL_TIME_REMAINING: 22.0hrs 39.0mins 51secs>
- <TOTAL_STEPS_REMAINING: 4982>
...
\end{lstlisting}

\#\# Data Overview\\

\begin{lstlisting}[style=bashstyle]
/PATH/TO/RAW/DATA
  |
  +-sampleSubmission.csv (6494 lines)
  |
  +-.. and 2 other files
\end{lstlisting}

\#\# Task description\\

Dataset Description \\

See, fork, and run a random forest benchmark model through Kaggle Scripts\\
You are provided hourly rental data spanning two years. For this competition, the training set is comprised of the first 19 days of each month, while the test set is the 20th to the end of the month. You must predict the total count of bikes rented during each hour covered by the test set, using only information available prior to the rental period.
Data Fields \\
datetime - hourly date + timestamp  \\
season -  1 = spring, 2 = summer, 3 = fall, 4 = winter\\ 
holiday - whether the day is considered a holiday\\
workingday - whether the day is neither a weekend nor holiday\\
weather - 1: Clear, Few clouds, Partly cloudy, Partly cloudy\\
2: Mist + Cloudy, Mist + Broken clouds, Mist + Few clouds, Mist\\
3: Light Snow, Light Rain + Thunderstorm + Scattered clouds, Light Rain + Scattered clouds\\
4: Heavy Rain + Ice Pallets + Thunderstorm + Mist, Snow + Fog \\
temp - temperature in Celsius\\
atemp - "feels like" temperature in Celsius\\
humidity - relative humidity\\\
windspeed - wind speed\\
casual - number of non-registered user rentals initiated\\
registered - number of registered user rentals initiated\\
count - number of total rentals\\

sampleSubmission.csv has 6493 rows and 2 columns.\\
The columns are: datetime, count\\

test.csv has 6493 rows and 9 columns.\\
The columns are: datetime, season, holiday, workingday, weather, temp, atemp, humidity, windspeed\\

train.csv has 10886 rows and 12 columns.\\
The columns are: datetime, season, holiday, workingday, weather, temp, atemp, humidity, windspeed, casual, registered, count\\

\#\# Summary of Past Submissions\\
We test multiple ensemble models and combined the best-performing ones to make final predictions. Each model has a different contribution based on its performance. Below are some models and their contributions in the final prediction, along with their hyperparameters. Use this information to come up with a solution for the competition. There may be better models and sets of hyperparameter that can be used so you are free to explore and come up with a better solution

\#\#\# catboost

\noindent
\colorbox{gray!10}{%
  \parbox{\dimexpr\textwidth-2\fboxsep\relax}{%
    \begin{adjustbox}{max width=\textwidth}
    \begin{tabular}{|c|c|c|c|c|c|c|c|c|c|}
    \hline
    \textbf{Learning Rate} & \textbf{Max Depth} & \textbf{Estimators} & \textbf{Bootstrap Type} & \textbf{Borders} & \textbf{Grow Policy} & \textbf{L2 Leaf Reg} & \textbf{Min Data Leaf} & \textbf{Random Strength} & \textbf{Contrib. (\%)} \\
    \hline
    0.05 & 12 & 100 & MVS & 64 & Depthwise & 0.3 & 5 & 1 & 22 \\
    0.05 & 16 & 300 & MVS & 510 & SymmetricTree & 0.3 & 10 & 10 & 12 \\
    0.15 & 13 & 70 & MVS & 32 & Depthwise & 0.1 & 5 & 1 & 11 \\
    0.05 & 16 & 300 & MVS & 510 & SymmetricTree & 0.2 & 1 & 100 & 8.9 \\
    0.05 & 16 & 300 & MVS & 1000 & SymmetricTree & 0.3 & 20 & 100 & 8.2 \\
    0.05 & 16 & 400 & MVS & 510 & SymmetricTree & 0.3 & 10 & 100 & 7.8 \\
    0.25 & 16 & 40 & Bayesian\_0 & 32 & Depthwise & 5 & 5 & 0 & 6.6 \\
    0.1 & 15 & 200 & Bayesian\_1 & 250 & Depthwise & 0.4 & 10 & 50 & 6.4 \\
    0.05 & 16 & 500 & Bayesian\_0 & 1000 & SymmetricTree & 0.5 & 100 & 100 & 3.2 \\
    0.25 & 15 & 40 & Bayesian\_5 & 32 & Depthwise & 2 & 5 & 0 & 3.1 \\
    \hline
    \end{tabular}

    \end{adjustbox}
  }%
}
\end{userbox}
\begin{flushright}

\begin{assistantbox}
\#\#\# I propose using a CatBoost model, which is known for its effectiveness in handling categorical features and reducing overfitting. CatBoost can capture complex patterns in the data, potentially leading to better performance. The evaluation metric will still be the Root Mean Squared Logarithmic Error (RMSLE).\\

\begin{lstlisting}[style=pythonstyle]
import pandas as pd
import numpy as np
from catboost import CatBoostRegressor
from sklearn.model_selection import train_test_split
from sklearn.metrics import mean_squared_log_error

# Load the data
train_df = pd.read_csv('/PATH/TO/RAW/DATA/bike-sharing-demand/train.csv')
test_df = pd.read_csv('/PATH/TO/RAW/DATA/bike-sharing-demand/test.csv')

# Preprocess the data
train_df['datetime'] = pd.to_datetime(train_df['datetime'])
test_df['datetime'] = pd.to_datetime(test_df['datetime'])

# Extract features and target
X_train = train_df.drop(['datetime', 'casual', 'registered', 'count'], axis=1)
y_train = np.log1p(train_df['count'])

X_test = test_df.drop(['datetime'], axis=1)

# Split the data into training and validation sets
X_train, X_val, y_train, y_val = train_test_split(X_train, y_train, test_size=0.2, random_state=42)

# Initialize the CatBoost model
model = CatBoostRegressor(
    learning_rate=0.05,
    max_depth=16,
    n_estimators=300,
    bootstrap_type='MVS',
    border_count=510,
    grow_policy='SymmetricTree',
    l2_leaf_reg=0.3,
    min_data_in_leaf=10,
    random_strength=100,
    silent=True
)

# Train the model
model.fit(X_train, y_train)

# Make predictions on the validation set
y_pred_val = model.predict(X_val)

# Calculate the RMSLE on the validation set
rmsle = np.sqrt(mean_squared_log_error(np.expm1(y_val), np.expm1(y_pred_val)))
print(f'Validation RMSLE: {rmsle}')

# Make predictions on the test set
y_pred_test = model.predict(X_test)

# Prepare the submission file
submission = pd.DataFrame({
    'datetime': test_df['datetime'],
    'count': np.expm1(y_pred_test)
})

# Save the submission file
submission.to_csv('./submission/submission.csv', index=False)
\end{lstlisting}
\end{assistantbox}
\end{flushright}

\subsection{AIDE with RAG}
The example below is taken from the solution generation process of one of the nodes in the \texttt{invasive-species-monitoring} competition. In this case, AIDE is provided with one of the most similar examples from the RAG database to serve as a guide to its solution generation.

\begin{userbox}

\begin{lstlisting}[style=bashstyle]
...
<Skipped the common instructions(same as \OurAgent{} CoT cases) for prompt display purposes in paper appendix, only changes are added below>
- <TOTAL_TIME_REMAINING: 94.0hrs 33.0mins 7secs>
- <TOTAL_STEPS_REMAINING: 4998>
...
\end{lstlisting}

\#\# Here is a few past experience cases written by an human expert for a relevant (but not the same) task\\

The overall design of the code is centered around a machine learning pipeline for the Kaggle Cassava Leaf Disease Classification competition. The pipeline includes data loading, transformation, model creation, and inference. It utilizes several pre-trained models for image classification, which are fine-tuned using cross-validation. The main script is designed to handle multiple models, different transformations, and test-time augmentation (TTA) to improve the robustness and accuracy of the predictions.\\

The model architecture is composed of multiple pre-trained models from the PyTorch Image Models (timm) library and Facebook Research's DeiT. The models used include `tf\_efficientnet\_b4\_ns`, `vit\_base\_patch16\_384`, and `seresnext50\_32x4d`. Each model is modified to fit the specific classification task by replacing the final fully connected layer to output the number of classes (5 for the Cassava leaf disease dataset). Here is a detailed breakdown of each model architecture:\\
1. **tf\_efficientnet\_b4\_ns**: EfficientNet-B4 variant that is optimized for performance and efficiency. The model's classifier layer is replaced with a new linear layer of size 5.\\
2. **vit\_base\_patch16\_384**: A Vision Transformer (ViT) base model with patch size 16 and input image size 384x384. The head layer is replaced with a linear layer of size 5.\\
3. **seresnext50\_32x4d**: A variant of the ResNeXt50 model with SE (Squeeze-and-Excitation) blocks. The final fully connected layer is replaced with a linear layer of size 5.\\

All models are loaded with or without pretraining and are adapted to the task by adjusting the output layer to match the number of classes in the dataset.\\

The important hyper-parameters in this code are set within the `CFG` class. Key parameters include:\\
- `models`: A list of model names to be used in the ensemble.\\
- `size`: Dictionary mapping each model to its input image size.\\
- `batch\_size`: Number of samples per batch during training and inference.\\
- `target\_size`: Number of classes in the dataset (5).\\
- `n\_fold`: Number of folds for cross-validation.\\
- `trn\_fold`: A dictionary specifying the best and final folds for each model that should be used during training and inference.\\
- `data\_parallel`: Boolean dictionary indicating whether to use data parallelism for each model.\\
- `transform`: Transformation types applied to the data during inference.\\
- `weight`: Weights assigned to each model during the final prediction aggregation.\\
- `tta`: Number of test-time augmentations to perform.\\
- `no\_tta\_weight`: Weight assigned to the non-augmented inference results.\\

The optimization objective is not explicitly defined in the provided code snippet, as the code focuses on the inference process. However, based on typical practices in such competitions, the model is likely trained to minimize a loss function, such as cross-entropy loss, using the Adam optimizer or another suitable optimizer. The model is evaluated using metrics like accuracy, and the goal is to maximize the accuracy score on the validation set during training.\\

The advanced machine learning techniques used in this code include:\\
1. **Ensemble Learning**: Multiple pre-trained models are combined to form an ensemble, which helps in improving the generalization and robustness of the predictions.\\
2. **Cross-Validation**: The dataset is split into multiple folds, and the models are trained and validated using these folds to ensure better performance and avoid overfitting.\\
3. **Data Augmentation**: Various data augmentation techniques are applied during inference to improve the model's ability to handle variations in the input data.\\
4. **Test-Time Augmentation (TTA)**: Multiple predictions are generated for each input image using different transformations, and the results are averaged to produce the final prediction.\\

Other important tricks that contribute to the high performance of this code include:\\
- **Model Weighting**: Each model in the ensemble is assigned a weight, which is used to aggregate the predictions. This helps in leveraging the strengths of different models.\\
- **Careful Selection of Pre-trained Models**: The choice of pre-trained models (e.g., EfficientNet, ViT, and ResNeXt) ensures that the ensemble covers a diverse range of architectures, which can capture different aspects of the data.\\
- **Custom Transformations**: The use of custom transformations tailored to each model helps in optimizing the performance of each individual model.\\
- **Logging and Monitoring**: The use of logging allows for detailed tracking of the training and inference processes, which is crucial for debugging and performance optimization.\\

\#\# Data Overview\\
Your solution can use any relevant machine learning packages such as: `torchvision`, `xgboost`, `timm`, `pandas`, `torch-geometric`, `scikit-learn`, `statsmodels`, `numpy`, `torch`, `bayesian-optimization`, `lightGBM`. Feel free to use any other packages too (all packages are already installed!). For neural networks we suggest using PyTorch rather than TensorFlow.\\
\begin{lstlisting}[style=bashstyle]
/PATH/TO/RAW/DATA
  |
  +-sample_submission.csv (1532 lines)
  |
  +-test/
  | +-1.jpg (31.8 kB)
  | +-10.jpg (4.5 kB)
  | +-100.jpg (16.9 kB)
  | +-... and 397 other files
  |
  +-test/
    |   +-1.jpg (648.4 kB)
    |   +-10.jpg (595.3 kB)
    |   +-100.jpg (637.2 kB)
    |   +-... and 1528 other files
  +-train/
    |   +-1.jpg (974.4 kB)
    |   +-10.jpg (1.1 MB)
    |   +-100.jpg (1.1 MB)
    |   +-... and 2291 other files

\end{lstlisting}

\#\# Task description\\

Dataset Description\\
The data set contains pictures taken in a Brazilian national forest. In some of the pictures there is Hydrangea, a beautiful invasive species original of Asia. Based on the training pictures and the labels provided, the participant should predict the presence of the invasive species in the testing set of pictures.\\
File descriptions\\
train.7z - the training set (contains 2295 images).\\
train\_labels.csv - the correct labels for the training set.\\
test.7z - the testing set (contains 1531 images), ready to be labeled by your algorithm.\\
sample\_submission.csv - a sample submission file in the correct format.\\
Data fields\\
name - name of the sample picture file (numbers)\\
invasive - probability of the picture containing an invasive species. A probability of 1 means the species is present.\\

Evaluation\\

Submissions are evaluated on area under the ROC curve between the predicted probability and the observed target.
Submission File\\
For each image in the test set, you must predict a probability for the target variable on whether the image contains invasive species or not. The file should contain a header and have the following format:\\
name,invasive\\
2,0.5\\
5,0\\
6,0.2\\
etc.

\end{userbox}

\begin{flushright}
\begin{assistantbox}
\#\#\# The solution will involve building a machine learning pipeline for image classification using a combination of pre-trained models. We will use a blend of models from the PyTorch Image Models (timm) library, including `tf\_efficientnet\_b4\_ns` and `vit\_base\_patch16\_384`. The models will be fine-tuned on the training data using cross-validation, and test-time augmentation (TTA) will be applied to improve the robustness of the predictions. The final predictions will be aggregated using weighted averaging, and the results will be saved in the required submission format.\\

\begin{lstlisting}[style=pythonstyle]
import os
import pandas as pd
import numpy as np
import torch
from torch.utils.data import DataLoader, Dataset
import torchvision.transforms as transforms
from timm import create_model
from sklearn.model_selection import StratifiedKFold
from sklearn.metrics import roc_auc_score
from PIL import Image

# Configuration
class CFG:
    models = ['tf_efficientnet_b4_ns', 'vit_base_patch16_384']
    size = {'tf_efficientnet_b4_ns': 384, 'vit_base_patch16_384': 384}
    batch_size = 16
    target_size = 1
    n_fold = 5
    trn_fold = {model: [0, 1, 2, 3, 4] for model in models}
    transform = {
        'train': transforms.Compose([
            transforms.Resize((384, 384)),
            transforms.RandomHorizontalFlip(),
            transforms.ToTensor(),
            transforms.Normalize([0.485, 0.456, 0.406], [0.229, 0.224, 0.225])
        ]),
        'test': transforms.Compose([
            transforms.Resize((384, 384)),
            transforms.ToTensor(),
            transforms.Normalize([0.485, 0.456, 0.406], [0.229, 0.224, 0.225])
        ])
    }
    tta = 5
    no_tta_weight = 0.5

# Dataset
class InvasiveDataset(Dataset):
    def __init__(self, df, root_dir, transform=None, mode='train'):
        self.df = df
        self.root_dir = root_dir
        self.transform = transform
        self.mode = mode

    def __len__(self):
        return len(self.df)

    def __getitem__(self, idx):
        img_name = os.path.join(self.root_dir, f"{self.df['name'].iloc[idx]}.jpg")
        image = Image.open(img_name).convert('RGB')
        if self.transform:
            image = self.transform(image)
        if self.mode == 'train':
            label = self.df['invasive'].iloc[idx]
            return image, label
        else:
            return image

# Model
def get_model(model_name, pretrained=True):
    model = create_model(model_name, pretrained=pretrained, num_classes=1)
    return model

# Training and Evaluation
def train_model(model, dataloader, criterion, optimizer, device):
    model.train()
    running_loss = 0.0
    for inputs, labels in dataloader:
        inputs, labels = inputs.to(device), labels.to(device).float().unsqueeze(1)
        optimizer.zero_grad()
        outputs = model(inputs)
        loss = criterion(outputs, labels)
        loss.backward()
        optimizer.step()
        running_loss += loss.item() * inputs.size(0)
    return running_loss / len(dataloader.dataset)

def evaluate_model(model, dataloader, device):
    model.eval()
    all_preds = []
    all_labels = []
    with torch.no_grad():
        for inputs, labels in dataloader:
            inputs, labels = inputs.to(device), labels.to(device).float().unsqueeze(1)
            outputs = model(inputs)
            preds = torch.sigmoid(outputs).cpu().numpy()
            all_preds.extend(preds)
            all_labels.extend(labels.cpu().numpy())
    return np.array(all_preds), np.array(all_labels)

# Main
def main():
    train_df = pd.read_csv('/PATH/TO/RAW/DATA/invasive-species-monitoring/train_labels.csv')
    test_df = pd.read_csv('/PATH/TO/RAW/DATA/invasive-species-monitoring/sample_submission.csv')
    skf = StratifiedKFold(n_splits=CFG.n_fold, shuffle=True, random_state=42)
    
    for model_name in CFG.models:
        model = get_model(model_name)
        model.to('cuda')
        criterion = torch.nn.BCEWithLogitsLoss()
        optimizer = torch.optim.Adam(model.parameters(), lr=1e-4)
        
        for fold, (train_idx, val_idx) in enumerate(skf.split(train_df, train_df['invasive'])):
            if fold not in CFG.trn_fold[model_name]:
                continue
            
            train_data = train_df.iloc[train_idx]
            val_data = train_df.iloc[val_idx]
            
            train_dataset = InvasiveDataset(train_data, '/PATH/TO/RAW/DATA/invasive-species-monitoring/train', CFG.transform['train'], 'train')
            val_dataset = InvasiveDataset(val_data, '/PATH/TO/RAW/DATA/invasive-species-monitoring/train', CFG.transform['test'], 'train')
            
            train_loader = DataLoader(train_dataset, batch_size=CFG.batch_size, shuffle=True, num_workers=4)
            val_loader = DataLoader(val_dataset, batch_size=CFG.batch_size, shuffle=False, num_workers=4)
            
            for epoch in range(5):  # 5 epochs for simplicity
                train_loss = train_model(model, train_loader, criterion, optimizer, 'cuda')
                val_preds, val_labels = evaluate_model(model, val_loader, 'cuda')
                val_auc = roc_auc_score(val_labels, val_preds)
                print(f"Fold {fold}, Epoch {epoch}, Train Loss: {train_loss:.4f}, Val AUC: {val_auc:.4f}")
        
        test_dataset = InvasiveDataset(test_df, '/PATH/TO/RAW/DATA/invasive-species-monitoring/test', CFG.transform['test'], 'test')
        test_loader = DataLoader(test_dataset, batch_size=CFG.batch_size, shuffle=False, num_workers=4)
        
        test_preds = []
        for _ in range(CFG.tta):
            preds = evaluate_model(model, test_loader, 'cuda')[0]
            test_preds.append(preds)
        test_preds = np.mean(test_preds, axis=0)
        
        test_df['invasive'] = test_preds
        test_df[['name', 'invasive']].to_csv(f'./submission/{model_name}_submission.csv', index=False)

if __name__ == "__main__":
    main()
\end{lstlisting}
\end{assistantbox}
\end{flushright}

\printbibliography
\end{refsection}

\end{document}